\documentclass[graybox,envcountchap,sectrefs]{svmono}

\usepackage{hyperref}
\usepackage{mathptmx}
\usepackage{helvet}
\usepackage{courier}
\usepackage{type1cm}         

\usepackage{makeidx}         
\usepackage{graphicx}        
\usepackage{multicol}        
\usepackage[bottom]{footmisc}

\usepackage{graphicx}
\usepackage{latexsym}
\usepackage{mathrsfs}
\usepackage{mathtools}
\usepackage{amssymb}
\usepackage{amsmath}
\usepackage{manfnt}
\usepackage{algorithm,algpseudocode}
\usepackage{lettrine}
\usepackage{esint}
\usepackage{booktabs}

\usepackage[geometry]{ifsym}

\def\R{\mbox{$I\!\!R$}}
\def\C{\mbox{\rm C}}

\def\path{\leadsto}
 
\def\ONE{\mbox{\boldmath $1$}}

\DeclareMathOperator*{\argmin}{arg\,min}

\def\circop{\mathop{\vcenter{\hbox{\includegraphics{./circ-0.mps}}}}}

\def\shape#1{\rm }

\def\eps{\varepsilon}

\def\I{\mathscr{I}}

\def\HS{\mathrm{HS}}
\def\HSs{\overline{\mathrm{HS}}}

\def\Id{\mbox{Id}}

\def\argmax{\mathop{\operator@font argmax}}

\def\B0{\mbox{\boldmath $0$}}

\def\Real{\mbox{$\mathbb{R}$}}

\newcommand{\bbR}{{\mathbb{R}}}
\newcommand{\bbX}{{\mathbb{X}}}

\newcommand{\Dgrad}{\mathrm{D}}
\newcommand{\tgrad}{D}
\newcommand{\sgrad}{\nabla}
\newcommand{\allphi}{\vec\varphi}
\newcommand{\allv}{\vec v}

\DeclareMathOperator{\spn}{span}
\DeclareMathOperator{\supp}{supp}
\DeclareMathOperator{\diam}{diam}
\DeclareMathOperator{\rnk}{rank}

\DeclareMathOperator{\Leb}{\mathcal{L}}

{\begin{list}{}{\setlength{\itemindent}{-\leftmargin}\setlength{\itemsep}{0pt}\setlength{\parsep}{0pt}}}%
{\end{list}}

\newcommand{\comment}[1]{}

\usepackage{newtxtext}       %
\usepackage{newtxmath}       
\usepackage[fulladjust]{marginnote} 

\usepackage{atbegshi,picture}

\reversemarginpar 
\let \marginpar=\marginnote 

\def\g#1\g{\marginpar{\vbox{\hsize=3pc
\def\|#1|{\includegraphics{./figs/wwf-#1.mps}}
\baselineskip=10pt
\lineskip=0pt \lineskiplimit=0pt\parindent=0pt
\mathsurround=1pt \tolerance=2000
\hyphenpenalty=300                                                            
\exhyphenpenalty=300                                                          
\doublehyphendemerits=100000                                                  
\finalhyphendemerits=\doublehyphendemerits   
\footnotesize\textsl{#1}}}}

\newif\ifdraft
\draftfalse

\newcommand{\draftA}[1]{\ifdraft{\color{orange}#1}\fi}
\ifdraft\usepackage[notref,notcite]{showkeys}\fi

\font\manfnt=manfnt 
\def\becomes{\ifmmode\ \hbox\fi{\manfnt y}\ } 

\def\boxformat{
}

\def\mixthree{\begingroup
  \def\\{\noalign{\penalty-200}}
    \halign\bgroup         
          \hfil$##$\hfil\quad&         
          \vtop{\noindent\hsize=26.9em ##\medskip}\cr}
          
\def\endmix{\egroup\endgroup}

\makeindex             

\begin{document}

\title{DEEP LEARNING TO SEE \\ 
\Large Towards New Foundations of Computer Vision}
\author{\Large
Alessandro Betti, Marco Gori and Stefano Melacci
}
\institute{Marco Gori \at DIISM, University of Siena, 
 \email{marco.gori@unisi.it}\\
}
%
%

\AtBeginShipoutNext{\AtBeginShipoutUpperLeft{%
  \put(1in + \hoffset + \oddsidemargin,-5pc){\makebox[0pt][l]{
\boxformat 
  \framebox{\vbox{
\hbox{Published by Springer, Cham---Cite this book:}
\smallskip
\hbox{\url{https://doi.org/10.1007/978-3-030-90987-1}}}
  }}}%
}}

\maketitle
\begin{dedication}
To All People Who Love To Ask Questions 
\end{dedication}
\tableofcontents

\preface
%
%
%
\vskip -5pc
\lettrine[lraise=0.1, nindent=0em, slope=-.5em]{D}{eep}  learning has revolutionized computer\index{deep learning}
vision and visual perception. Amongst others, the great representational power of convolutional neural networks and the elegance and efficiency of Backpropagation have played a crucial role.
By and large, there is a strong scientific recognition of their popularity, which is very 
well deserved. However, as yet, most significant results are still based on the 
truly artificial supervised learning communication protocol, 
which sets in fact a battlefield for computers, but it is far from being natural.
In this book we argue that, when relying on supervised learning, we
have been working on a problem that is - from a computational point of view - 
remarkably different and like more difficult 
with respect to the one offered by Nature, where motion is
in fact in charge for generating visual information. 
Couldn't be the case that
motion is fact nearly all what we need for learning to see? 
Otherwise, how could eagles acquire such a spectacular visual skills? What else could they grasp from a video to 
extract precious information for learning? For sure, just like other animals, they do not undergo 
a massive supervision, but only a  reinforcement signal due to their natural interactions with the environment.
Current deep learning\index{deep learning}
approaches  based on supervised images mostly neglect the crucial role of 
temporal coherence. It looks like Nature did a nice job by using time to sew all the video frames. 
When computer scientists began to cultivate the idea of interpreting natural
video, in order to simplify the problem they remove time, the connecting wire between
frames. As a consequence, video turned into huge collections of images, where 
temporal coherence was lost, which means that we are neglecting a fundamental clue to interpret visual information, 
and that we have ended up into problems where the extraction of the visual concepts 
can only be based on  spatial  regularities.

Based on the underlying representational capabilities of deep architectures 
and learning algorithms that are still related to Backpropagation,
in this book we propose that the massive image supervision can in fact be replaced
with the natural communication protocol arising from living in a visual environment,
just like animals do. This leads to formulate learning regardless of the accumulation
of labelled visual databases, but simply by allowing visual agents to live in their
own visual environments. We claim that feature learning arises mostly from motion invariance 
principles that turns out to be fundamental for detecting the object identity as well as 
supporting object affordance. 

This book introduces two fundamental principles of visual perception.  
The {\em first principle} involves consistency issues, namely  the preservation of material points identity  
during motion. Depending on the pose, some of those points are projected onto the retina.
Basically, the material points of an object are subject to  {\em motion invariance of the 
corresponding pixels on the retina}. A moving object clearly doesn't  change its identity and, therefore,
imposing an invariance leads to a natural formulation of object recognition.
Interestingly, more than the recognition of an object category, this leads to the discovering of 
its identity. 

Motion information doesn't only confer object identity, but also its affordance,
which corresponds with its function in real life.
Affordance makes sense for a species of animal, where specific actions take place.
A chair, for example, has the affordance of seating a human being, but it can have 
other potential uses. The {\em second principle of visual perception} is about its 
affordance as transmitted by coupled objects - typically humans.  The principle states that
the affordance is invariant under the coupled object movement. Hence, a chair gains the 
seating affordance independently of the movement of the person who's seating (coupled object).

The theory of deep learning\index{deep learning}
to see that is herein proposed is independent of the body 
of the visual agent since it is only based on information-based principles.
In particular, we introduce a vision field theory for expressing those motion invariance
principles. The theory enlightens the indissoluble pair of
visual features and their conjugated velocities, thus extending the
classic brightness invariance principle for the optical flow estimation.
The emergence of visual features in the natural framework  of visual environments is given
a systematic foundation by establishing information-based laws that naturally enable 
deep learning processes.

%
%
The ideas herein presented have been stimulated by a number of questions that we regard of fundamental
importance for the construction of a theory of vision. 
How can animals conquer visual skills without requiring the ``intensive supervision'' we impose to machines?
What is the role of time? More specifically, what is the interplay between the time of the agent  and the time of the environment?
Can animals see in a world of shuffled frames like computers do?
How can we perform semantic pixel labelling by receiving only a few supervisions?
Why has the visual cortex evolved towards a hierarchical organization and why did it split into
two functionally separated  mainstreams?
Why top level visual skills are achieved in nature by animals with foveated eyes thanks to focus of attention? 
What drives eye movements?
Why does it take 8-12 months for newborns to achieve adult visual acuity?
How can we develop ``linguistic focusing mechanisms''   that can drive the process of object recognition?

This book is a humble attempt at addressing these questions and it's far away from
providing a definite answer. However, 
the proposed theory gives foundations and insights to stimulate future investigations
and specific applications to computer vision. 
Moreover, the field theory herein proposed might  also open the doors to
disclose interesting problems in visual perception and  capture experimental evidence in neuroscience.
\vspace{\baselineskip}
\begin{flushright}\noindent
Siena,\hfill {\it Alessandro Betti}\\
August 2021\hfill {\it Marco Gori}\\
\hfill {\it Stefano Melacci}\\
\end{flushright}

\extrachap{Acknowledgements}
%
%
It's hard not to forget people who have contributed in different ways to shape the 
ideas that are proposed in this book. 

%
%
First, the research team in SAILab (Siena Artificial intelligence Lab) has played 
the most important support for maturing the ideas elaborated in this book. 
Matteo Tiezzi, Dario Zanca, Enrico Meloni, Lapo Faggi, and Simone Marullo, who are some of the members of the 
SAILab research team in computer vision have contributed with their 
experimental results to shape our ideas and make the theory simpler and more general.
There are in fact significant traces of early studies in SAILab from the collaboration with Marco Lippi
and Marco Maggini, who contributed to develop the first approaches to learning to see
by using motion invariance.  In particular, discussions with Marco Maggini, who early discovered 
a number of slippery issues in the incorporation of motion invariance and clearly 
identified the major difficulties in carrying out learning in the temporal domain, 
 have been extremely inspiring.

%
%
The road that has led to this book certainly crosses some inspiring meeting with Marcello 
Pelillo and, later on, with Fabio Roli. Marcello ignited my latent passion 
for unifying and that of looking for invariants in vision. Together with Fabio, years
ago, we cultivated the dream of a truly new way of facing computer vision challenges within the ``en plein air'' 
framework, which reminds us of painting outdoor - machines which learn directly in their
own environment. This is in fact addressed at the end of the book. Most of the comments definitely come from
early discussions which began with the Workshop GIRPR 2014. 

%
%
Oswald Lanz has been for us the main reference and source of inspirations for studies in
optical flow, which has  subsequently given rise to the two principles of perceptual vision. 
In particular, the conception of the idea of specific velocities associated with visual features has been 
originated by his clear presentation of the state of the art in optical flow and in tracking. 

%
%
The studies by Tomaso Poggio  on developing visual features under invariance stimulated very
fruitful discussion in connection with the Workshop on Workshop on ``Biologically Plausible Learning'' at
LOD 2020 and have fueled significantly the development of the theory presented in this book.

%
%
During the PhD studies of Alessandro Betti we benefited from a number of constructive criticisms
from Stefano Soatto and Michael Bronstein. Amongst others, they stimulated the importance of 
setting up an experimental framework adequate to assess the performance. In the same 
direction, a discussion with Bastian Leibe on the current state of the art in computer vision has
 contributed to shape and reinforce the idea of ``en plein air'' discussed in the Epilogue of 
 the book, thus promoting the fundamental principle of replacing the accumulation of visual databases
 with virtual visual environments.
The studies by Ulisse Stefanelli on the reformulation of the principle of least action in Physics
were a fundamental source of inspiration for the development of the on-line formulation of learning
reported in this book. The same idea used in mechanics gives rise to on-line gradient-based learning which
is nicely related to classic stochastic gradient descent. 

%
%
We have been inspired by a a number of studies in neuroscience, particularly on the mechanisms 
behind eye movements. Leonardo Chelazzi's visit to SAILab was very influential concerning the subsequent
development of computational models of focus of attention. We strongly benefited from discussions
on the different kinds of eye movements and, particularly, on the supposed inhibition of video
transmission during saccadic movements. The collaboration with Alessandra Rufa offered the primary
support on the formulation of theory of gravitational attraction of attention, which is also at the
basis of the local  spatiotemporal model reported in the book.
Giuseppe Boccignone provided a rich bibliographic support and stimulated many discussions 
mostly on the joint role of action and perception and on the vision blurring process in newborns and 
in chicks.


\chapter{Motion is the protagonist of vision
}

\AtBeginShipoutNext{\AtBeginShipoutUpperLeft{%
  \put(1in + \hoffset + \oddsidemargin,-5pc){\makebox[0pt][l]{
\boxformat 
  \framebox{\vbox{
\hbox{Published by Springer, Cham---Cite this chapter:}
\smallskip
\hbox{\url{https://doi.org/10.1007/978-3-030-90987-1_1}}}
  }}}%
}}

\vspace{-4cm}
\begin{quote}
There's not a morning I begin
without a thousand questions running through my mind
... The reason why a bird was given wings
If not to fly, and praise the sky ...\\
 {\em  From Yentl, "Where is it Written?" - I.B. Singer, The Yeshiva Boy}
\end{quote}

\section{Introduction}
A good
way to approach to fundamental problems is to pose the right questions.
Posing the right questions does require a big picture in mind and the need to identify 
reasonable intermediate steps. When, years ago, we began together 
the contamination with computer vision, 
a lot questions run through our mind that we think make this field really fascinating!
First of all, we were wondering whether can we regard computer vision as an application of 
machine learning\index{machine learning}. We early came up with the answer that while learning does play a primary
role in vision, the current methodologies don't really address its essence yet.
\marginpar{\small The challenge of posing the right questions}
The impressive quality of visual skills  ordinarily exhibited by humans and animals led us to pose 
a number of specific questions:  
How can animals conquer visual skills without requiring the ``intensive supervision'' we impose on machines?
What is the role of time? More specifically, what is the interplay between the time of the agent  and the time of the environment?
Can animals see in a world of shuffled frames like computers do?
How can we perform semantic pixel labelling without receiving any specific single supervision on that task?
Why has the visual cortex evolved towards a hierarchical organization and why did it split into
two functionally separated  mainstreams?
Why top level visual skills are achieved in nature by animals with foveated eyes thanks to focus of attention? 
What drives eye movements?
Why does it take 8-12 months for newborns to achieve adult visual acuity?
How can we develop ``linguistic focusing mechanisms''   that can drive the process of object recognition?
In the last few years, these questions have driven our curiosity in the science of vision. 
In this book we address those questions while 
disclosing information-based laws on the emergence of visual features that drive 
the computational processes of visual perception in case of natural communication protocols. 
\marginpar{\small Couldn't be the case that in computer vision we have been working on a problem that is
significantly more difficult than the one offered by Nature?}
Couldn't be the case that in computer vision we 
have been working on a problem that is significantly more difficult than the one offered by Nature.
This is mostly what we ask in this chapter, where we stimulate the unifying principle
of extracting visual information from motion only. 
This chapter is organized as follows. In the next section, we begin giving a big picture
of what this book is about, while in Section~\ref{SupLeaSec} we begin addressing
the motivations by stressing the artificial nature of the supervised learning protocol
and the involved complexity issues. In Section~\ref{Unbilicalsec} we emphasize
the importance of cutting the umbilical link with pattern recognition, while
in Section~\ref{VideoEnvSec} we stress the importance of working with 
video instead of collections of labelled images. Finally, in Section~\ref{10Q-sec}
we formulate {\em ten questions} for a theory of vision that drives most of the
analyses reported in this book.

\section{The big picture}
The remarkable progress in computer vision on object recognition in the last few years 
achieved by deep convolutional neural networks~\cite{lecun2015deeplearning} is strongly connected with
the availability of huge labelled data paired with strong and suitable computational
resources. Clearly, the corresponding supervised communication protocol between machines and
the visual environments is far from being natural. This protocol sets in fact a battlefield for computers. 
We should not be too surprised that very good results are already obtained, just like 
the layman is not surprised that computers are quick to do multiplication!
Still, one question is in order: Isn't the case that  while facing 
nowadays object recognition problems we have been working on a problem that is
significantly more difficult than the one offered by Nature?
Current deep learning\index{deep learning} 
approaches  based on supervised images mostly neglect the crucial role of 
temporal coherence. It looks like Nature did a nice job by using time to sew all the video frames. 
When computer scientists began to cultivate the idea of interpreting natural
video, in order to simplify the problem they remove time, the connecting wire between
frames. As a consequence, video turned into huge collections of images, where 
temporal coherence was lost. 
At a first glance, this is reasonable, especially if we consider that, traditionally, video were
heavy data sources. However, a closer look reveals that 
we are in fact mostly neglecting a fundamental clue to interpret visual information, 
and that we have ended up into problems where the extraction of the visual concepts 
can only be based on  spatial  regularities.

%
%
As we decide to frame visual learning processes in their own natural video environment
we early realize that perceptual visual skills cannot emerge from massive supervision on different object categories.
Moreover, linguistic skills arise in children when vision has already achieved a remarkable
degree of development and, most importantly, there are animals with excellent visual skills.
How can an eagle distinguish a prey at distances of kilometers? 
%
%
The underlying hypothesis that gives rise to the natural laws of visual perception proposed in this book 
is that visual information comes from motion. 
Foveated animals 
move their eyes, which means that even still images are perceived as patterns that change
over time. 
This book arises from the belief that we can understand the underlying 
information-based laws of visual perception regardless of the body of the agent.
No matter whether we are studying chicks, newborns, or computers, the information 
that arises from the surroundings through the light that enters their ``eyes'' is the same. 
As they open their eyes, they need to tackle problems 
that involve functional issues, that are only related to the nature of the source.
Since information is interwound with motion, we propose to explore the consequences of
stressing the assumption that the focus on motion is in fact {\em nearly all that we need}. 
When trusting this viewpoint one early realizes that, since we involve the optical flow,
the underlying computational model must refer to single pixels at a certain time.
In Nature, best visual perception is in fact achieved by animals which focus attention,
a feature that turns out to play a crucial role in terms of computational complexity.  
We show that this pixel-based assumption leads to state two invariant principles that drive the construction of 
computational models of visual perception, whenever  the agent is supposed
to establish natural communication protocols. 

\marginpar{\small Motion is nearly all what you need: The two principles of motion invariance}
A major claim here is that motion is nearly all you need for extracting information from a visual source.
We introduce  {\em two principles of visual perception} which express the invariance with
respect to the object and to the coupled-object motion, respectively. 
Motion is what offers us an object in all its poses.
Classic translation, scale, and rotation invariances can clearly be gained by appropriate movements of a given object.
However, the experimentation of visual interaction due to motion goes well beyond 
the need of these invariances
and it includes the object deformation, as well as its obstruction. 
Only small portions can be enough for the detection, even in 
presence of environmental noise.
The {\em first principle of visual perception} involves consistency issues, namely  the preservation of material points identity  
during motion. Depending on the pose, some of those points are projected onto the retina.
Basically, the material points of an object are subject to  {\em motion invariance of the 
corresponding pixels on the retina}. A moving object clearly doesn't  change its identity and, therefore,
imposing an invariance leads to a natural formulation of object recognition.
Interestingly, more than the recognition of an object category, this leads to the discovering of 
its identity. 
Motion information doesn't only confer object identity, but also its affordance\index{affordance},
which corresponds with its function in real life.
Affordance makes sense for a species of animal, where specific actions take place.
A chair, for example, has the affordance of seating a human being, but it can have 
other potential uses. The {\em second principle of visual perception} is about its 
affordance as transmitted by coupled objects - typically humans.  The principle states that
the affordance is invariant under the coupled object movement. Hence, a chair gains the 
seating affordance independently of the movement of the person who's seating (coupled object).

\marginpar{\small Motion invariance principles and information-based laws of feature development}
These two principles drive an information-based approach to visual perception. 
It's based on a field theory which predicts the visual features, including the 
object codes, and their associated velocities in all pixels of the retina for any frame.
Motion invariance leads to formulate constraint-based computational model where both the
velocities and the features are unknown variables, which are determined under the
regularization principle which reminds us of predicting coding, where we
enforce the reproduction of the visual information from motion invariant features.

The view of visual perception proposed in this book might have an impact on 
computer vision, where one could open the challenge of abandoning the protocol
of massive supervised learning based on huge labelled databases towards 
the more natural and simple framework in which machines, just like animals, are 
expected to learn to see in their own visual environment.\\
~\\
\section{Supervised learning is an artificial learning protocol}
\label{SupLeaSec}
The professional construction of huge supervised visual 
data bases has significantly contributed to the spectacular 
performance of deep learning\index{deep learning}. However, the extreme exploitation 
of the truly artificial communication protocol of supervised learning 
has been experimenting its drawbacks, including the vulnerability
to adversarial attacks, which might also be tightly connected to the
negligible role typically played by  the temporal structure of video signals.

\marginpar{\small Aren't we missing something?  
It looks like Nature did a great job by using time to ``sew all the video frames'', whereas it  goes unnoticed to our eyes!} 
At the dawn of pattern recognition, when scientists began to cultivate the
idea of interpreting video signals, in order to simplify the problem 
of dealing with a huge amount of information they removed time, the connecting thread between frames, and began playing with 
the pattern regularities that emerge at the spatial level. As a consequence, many
tasks of computer vision were turned into problems
formulated on  collections of images, where 
the crucial role of temporal coherence was neglected. 
Interestingly, when considering the general problem of object recognition
and scene interpretation, the joint role of the computational resources and the access to huge visual databases of supervised data has 
contributed to erect nowadays
``reign of computer vision''.  
At a first glance this is reasonable, especially if you consider that videos were
traditionally heavy data sources  to be played with. 
However, a closer look reveals that 
we are in fact neglecting a fundamental clue to interpret visual information, 
and that we have ended up into problems where the extraction of the visual concepts is mostly based on  spatial  regularities.
On the other hand, reigns have typically consolidated rules from which it's hard to escape.  This is common in novels and real life.
``The Three Princes of Serendip'' is the English version of ``Peregrinaggio di tre giovani figliuoli del re di Serendippo,'' published by Michele Tramezzino in Venice on 1557. These princes journeyed widely, and as they traveled they continually made discoveries, by accident and sagacity, of things they were not seeking. A couple of centuries later, in a letter of January 28, 1754 to a British envoy in Florence, the English politician and writer Horace Walpole coined a new term: serendipity, which is succinctly characterized as the art of finding something when searching for something else. Couldn't similar travels open new
scenario in computer vision? Couldn't the visit to 
well-established scientific domains, where time is dominating the scene, 
open new doors to an in-depth understanding of vision? 
We need to stitch the frames to recompose the video using time as a thread, the same thread we had extracted to work on the images at the birth of the discipline.

\marginpar{\small Beyond a peaceful interlude: 
learning theories based on video more than on images!}
This book is a travel towards the frontiers of the science of vision with special emphasis
on object perception. We drive the discussion by a number
of  questions - see Section~\ref{10Q-sec} - that mostly arise
as one tries to interpret and disclose natural vision processes 
in a truly computational framework. 
Regardless of the success in addressing those question,  this book comes from the 
awareness that posing right questions on the discipline might themselves stimulate 
its progress.  The driving principle of abandoning supervised images
and let the agent live in its own visual environment by exploiting 
the variations induced by motion leads to discover 
general information-based principles that can shed light also
on the neurobiological structure of visual processes. Hopefully, this 
can also opens the doors to an in-depth re-thinking of computer vision.


\section{Cutting the umbilical cord with pattern recognition}
\label{Unbilicalsec}
In the eighties, Satosi Watanabe wrote a seminal book~\cite{10.5555/4394}
in which he early pointed out the different facets of pattern recognition that
can be regarded as  perception, categorization,  induction, pattern recognition,
as well  as statistical decision making.
Most of the modern work on computer vision for object perception fits with 
Watanabe's view of pattern recognition as statistical decision making, and pattern recognition as categorization.
Based on optimization schemes with billions of variables and universal  
approximation capabilities, the spectacular results of deep learning have elevated 
this view of pattern recognition to a position where it is hardly debatable. 
While the emphasis on a general theory of vision was already 
the main objective at the dawn of the discipline~\cite{Marr82}, 
its evolution has been mostly marked to significant experimental achievements. 
Most successful approaches  seem to be the natural outcome of a very well-established tradition in pattern recognition methods working on images, which have given rise to nowadays emphasis on collecting big labelled image databases (e.g.~\cite{imagenet_cvpr09}).

%
%
In spite of these successful results, this could be the time of an in-depth rethinking of what we have 
been doing, especially by considering the remarkable traditions of the 
overall field of vision.  In his seminal book,
Freeman J. Dyson~\cite{dyson2004infinite},  discusses two distinct styles of scientific thinking: unifying and diversifying. He claimed that most sciences are dominated by one or the other in various periods of their history. Unifiers are people whose driving passion is to find general principles, whereas 
Diversifiers are people with the passion of details. 
Historical progress of science is often a blend of unification and diversification. 
Couldn't it be the right time to exploit the impressive literature in the field of vision
to conquer a more unified view of object perception? 

%
%
\marginpar{\small From pattern recognition to computer vision}
In the last few years, a number of studies in psychology and cognitive science have 
been pushing truly novel
approaches to vision. In~\cite{Kingstone2010}, it is pointed out that 
a critical problem that continues to bedevil the study of human cognition is related to 
the  remarkable successes gained in experimental psychology, where 
one is typically involved in simplifying
the experimental context  with the purpose to discover causal relationships. 
In so doing we minimize the complexity of the environment and maximize the experimental control,
which is typically done also in computer vision when we face object recognition. 
However, one might ask whether such a simplification is really adequate and, most
importantly, if it is indeed a simplification. Are we sure that treating vision as a collection
of unrelated frames leads to simplification of learning visual tasks? 
In this book we argue that this not the case, since 
cognitive processes vary substantially with changes in context. 
\marginpar{\small Cognitive Ethology and learning in the wild}
When promoting the actual environmental interaction, in~\cite{Kingstone2010}
a novel research approach, called ``Cognitive Ethology'', is introduced
where one opts to explore first how people behave in a truly naturally situation. 
Once we have collected experience and evidence in the actual environment then 
we can move into the laboratory to test hypotheses.  This strongly suggests
that also machines should learn in the wild!

%
%

Other fundamental lessons come from the school of robotics
for whatever involves the control of object-directed actions. 
In~\cite{Ballard2011}, it is pointed out that 
``the purpose of vision is very different when looking at a static scene with respect to when engaging in real-world behavior.''  
The interplay between  extracting visual information and coordinating the motor actions 
is a crucial issue to face for gaining an in-depth understanding of vision.  
One early realizes that manipulation of objects is not something
that we learn from a picture; it looks like we definitely need to act ourself if we want to 
gain such a skill. Likewise, the perception of the objects that we manipulate 
can nicely get a reinforcement from such a mechanical feedback.
The mentioned interplay between perception and action finds an
intriguing convergence in the natural processes of gaze control and, overall, on 
the focus of attention~\cite{DBLP:journals/ai/Ballard91}.
It looks like  {\em animate vision} goes beyond passive information extraction
and plays an important role in better posing most vision tasks.

%
%
\marginpar{\small Predictive coding and mental visualization processes}
The studies in computer vision might benefit significantly also from the exploration of
the links with predictive coding~\cite{rao:natneuro99} that have had a remarkable impact in neuroscience.
In that framework, one is willing to study theories of brain in which it constantly generates and updates a ``mental model'' of the environment. Overall, the model is supposed  to generate its own predictions of sensory input and 
compare them to the actual sensory input. 
The prediction error is expected to be used to update and revise the mental model.
While most of the studies in deep learning\index{deep learning} have been focused on the direct learning of object categories, 
there are a few contributions also in the direction of performing a sort of predictive coding by 
means of auto-encoding architectures~\cite{xia2017w}.
In this book we go one step beyond by enforcing the extraction of motion independent features.
The corresponding invariance is the secrete for identifying the objects as well
as the affordance. We shall see that the extracted codes have an inherently symbolic nature
which makes it possible to think of images and, consequently to carry out the 
mental visualization processes. 

\section{Dealing with video instead of images}
\label{VideoEnvSec}
While the emphasis on methods rooted on still images is common on  most of nowadays 
state of the art object recognition approaches, in this book we argue
 that there are strong arguments to start exploring the more natural visual interaction 
 that animals experiment in their own environment. 
The idea of shifting to video is very much 
related to the growing interest of {\em learning in the wild} that has been explored in the
last few years\footnote{See. e.g.~\url{https://sites.google.com/site/wildml2017icml/}.}.
The learning processes that take place in this kind of environments have a different
nature with respect to those that are typically considered in machine learning. 
While ImageNet~\cite{imagenet_cvpr09} is a collection of unrelated images, a video supports information only when motion is involved. In presence of still images that last for awhile, the
 corresponding stream of equal frames only conveys the information of a single image
 - apart from the duration of the interval in which the video has been kept constant. 
As a consequence,  visual environments mostly diffuse information only when
motion is involved. As time goes by, the information is only carried out by motion,
which modifies one frame to the next one according to the optical flow. Once we 
deeply capture this fundamental feature of vision,
we realize that a different theory of machine learning\index{machine learning} is needed that must be 
capable of naturally processing streams instead of a collection of independent images.   

The scientific communities of pattern recognition and computer vision
have a longstanding tradition of strong intersections. 
As the notion of pattern began to circulate at the end of the 
Fifties, the discipline of pattern recognition grew up quickly
around it. No matter what kind of representation a
pattern is given, its name  conveys the  underlining assumption that,
regardless of the way they are obtained,
the recognition process involves single entities. 
It is in fact their accumulation in appropriate collections which 
gives rise to statistical machine learning. 
\marginpar{\small The dominant foundations are still those promoted by the pioneers
of machine learning and pattern recognition}
While computer vision can hardly be delimited in terms   of both
methodologies and applications, the dominant foundations are
still those promoted by the pioneers of pattern recognition.  
Surprisingly enough, there is no much attention on the crucial
role of the position (pixel) on which the decision is carried out
and, even more, the role of time in the recognition processes
doesn't seem to play a central role. 
It looks like we are mostly ignoring  that we are in front of
spatiotemporal information, whose reduction to isolated patterns
might not be a natural solution. 


In the last decades, the massive production of electronic documents,
along with their printed version, 
has given rise to specialized software tools to extract textual information
from optical data. Most optical documents, like 
tax forms or invoices, are characterized by a certain layout which
dramatically simplifies the process of information extraction. 
Basically, as one recognizes the class of a document, its layout
offers a significant prior on what we can expect to find in its
different areas. For those documents, the segmentation process 
can often be given a somewhat formal description, so as most of the problems are
reduced to deal with the presence of noise. As a matter of 
fact, in most real-world problems, the noise doesn't compromise
significantly the segmentation, that is very well driven
by the expectations provided in each pixel of the documents.
These guidelines have been fueling the field of documental analysis
and recognition (DAR), whose growth in the last few years has led to
impressive results~\cite{DBLP:journals/pami/MarinaiGS05}. 
Unfortunately, in most real-world problems, as we move to natural images and vision,
the methodology used in DAR is not really effective in most challenging problems.
The reason is that there is no longer a reliable anchor
to which one can cling for segmenting the objects 
of a scene. While we can provide a clear description of chars and 
lines in optical documents, the same doesn't hold for the picture
of a car which is mostly hidden by a truck during the overtaking. 
Humans exhibit a spectacular detection ability 
by simply relying on  small glimpses at different scale and 
rotations. In no way those  cognitive processes are reducible to
the well-posed segmentation problems of chars and lines in optical documents.
As we realize that there is a car, we can in fact provide its segmentation.
Likewise, if an oracle gives us the segmented portion of a car,
we can easily classify it. Interestingly, we don't really know 
which of the two processes is given a priority - if any. 
\marginpar{\small The chicken-egg dilemma}
We are trapped into the {\em chicken-egg dilemma} on whether 
classification of objects must take place first of segmentation or vice versa.
Amongst others, this issue has been massively investigated
in~\cite{DBLP:conf/eccv/BorensteinU02}
and early pointed out in~\cite{UllmanShimon1979Tiov}.
This intriguing dilemma might be connected with the 
absence of  focus of attention,
which necessarily leads to holistic mechanisms of information extraction.
Unfortunately, while holistic mechanisms are required at a certain level of abstraction, 
the segmentation is a truly local process that involves also
low level features.

The bottom line is that most problems
of computer vision are posed according to the historical evolution of
the applications more than according to an in-depth analysis of the underlying computational
processes. While this choice has been proven to be successful in many real-world
cases, stressing this research guideline might led, on the long run, 
to sterile directions. 
Somewhat outside the mainstream of massive exploration of supervised 
learning, Poggio and Anselmi~\cite{Poggio:2016:VCD} pointed out the 
crucial role of incorporating appropriate visual invariance
into deep nets to go beyond the simple translation equivariance 
that is currently characterizing convolutional networks. They propose an elegant mathematical framework 
on visual invariance and enlighten some intriguing neurobiological connections. 
Couldn't it be the case that the development of appropriate invariances is exactly what 
we need to go one step beyond and also to improve significantly the performnce
on object recognition tasks?

\section{Ten questions for a theory of vision}
\label{10Q-sec}
A good way to attack important problems is to pose the right question.
John Tukey, who made outstanding contributions on the Fourier series, 
credited for the invention of the  term ``bit'',  during his research activity underlined the importance of posing
appropriate questions for an actual scientific development. In his own words:
\begin{quote}{\em
Far better an approximate answer to the right question, which is often vague, than the exact answer to the wrong question, which can always be made precise.
}\end{quote}
\marginpar{\small The ten driving questions}
Overall, posing appropriate questions can open a debate and solicit answers. 
However, the right questions cannot be easily posed since, while they need  
a big pictures in mind, they also need the identification of reasonable intermediate steps. 
It is often the case that while addressing little problems, inconsistencies arise that
suggest the formulation of better questions. Here we formulate ten 
questions on the emergence of visual skills in nature that might also contribute
to the development of  new approaches to computer vision that is based
in processing of video instead of huge collections of images. 
While this book is far from purporting to provide definitive answers to those questions, 
it contains some insights that might stimulate an in-depth re-thinking mostly of 
object perception. Moreover it might also suggest different research directions in the 
control of object-directed action.

\begin{enumerate}
\item [{\bf Q1}]
How can animals conquer visual skills without requiring ``intensive supervision''?
\begin{quote}
	In the last few years, the field of computer vision has been 
	at the center of a revolution. 
	Millions of supervised images have driven a substantial change in the
	methodology behind object recognition. No animal in nature can compete
	with machines on the game of supervised learning. Yet, animals 
	don't rely on linguistic  supervision. Similarly, humans conquer visual
	skills without such a boring communication protocol! 
	There must be some important reasons behind this fundamental difference. 
\end{quote}
\item [{\bf Q2}]
How can animals gradually conquer visual skills in their own environments?
\begin{quote}
	Based on their genetic heritage, animals learn to see while living. 
	Machines share some analogies:  
	First, the transfer of visual features learned in different experimental domains
	somewhat reminds us the importance of genetic heritage in nature.
	Second, the artificial learning processes, which are
	based on the weight updating, also remind us the gradual process 
	of learning that we observe in nature. However, this analogy
	is only apparent, since nowadays deep learning carries out an 
	optimization process whose iteration steps don't correspond 
	with the environmental time that animals experiment during their life.  
	It looks like, machine learning is neglecting the notion of time 
	in a process which is inherently driven by time!
\end{quote}

\item [{\bf Q3}]
Could  children really acquire visual skills in 
such an artificial world, which is the one we are presenting to machines?
Don't  shuffled visual frames increase the complexity of learning to see?
\begin{quote}
	Visual image databases can be thought of the outcome of  shuffling the frames
	of a video. We should pay attention to this issue, which likely
	surprises the layman, whereas it doesn't seem to attract significant attention
	in the scientific community. Couldn't be the case that we have been
	facing a problem more difficult than the one offered by Nature?   
	ImageNet~\cite{imagenet_cvpr09} and related visual databases somehow correspond with
	the permutation of visual sources.
\end{quote}

\item [{\bf Q4}] 
	How can humans exhibit such an impressive skill of properly labelling
	single pixels without having received explicit pixel-wise supervisions?
	It is not the case that such a skill is a sort of ``visual primitive''
	that cannot be ignored for efficiently conquering additional skills
	on object recognition and scene interpretation?
\begin{quote}
	When the emphasis is on applications involving object recognition, we don't necessarily need to 
	process information at pixel level. 
	However, there are cases, like medical image segmentation, where
	pixel-based computation is required. Interestingly, humans 
	exhibit great performance at semantic pixel labelling. 
	One might wonder whether such a human skill is in fact
	a fundamental primitive needed for an in-depth understanding 
	of vision. Moreover, as it will become clear in the rest of the
	book, the construction of a theory on perceptual vision very much benefits
	from the local reference to single pixels of both optical flow\index{optical flow} and visual features.
	The focus on pixel-based features and velocities somewhat leads to interpret
	vision in terms of a field theory. 
\end{quote}

\item [{\bf Q5}]
 Why are the visual mainstreams in the brain of primates 
 organized according to a hierarchical architecture 
with receptive fields? Is there any reason why this solution has been
developed in biology?

\begin{quote}
	This seems to be one of the secretes of the success of deep convolutional
	nets, that are currently learning under the supervised protocol. 
	Interestingly, it looks like the hierarchical structure, along with
	the concept of receptive field are inherently playing a crucial role
	in biology and in machines. However, one might wonder whether 
	the typical weight sharing, that is used for capturing the 
	translation equivariance is really needed. 
\end{quote}

\item [{\bf Q6}]
Why are there two different mainstreams in the primates' brain? What are the reasons for such 
a different neural evolution?

\begin{quote}
	The answer to this question is not only of interest in neuroscience.
	It seems reasonable to invoke functional specialization of the brain
	with distinct areas dedicated to perception and action, respectively. 
	There should be some fundamental difference for those neurons to be
	allocated separately. Capturing such a difference would likely help
	the development of computer vision systems.
\end{quote}

\item [{\bf Q7}]
	Why do primates and other animals focus attention, whereas 
	others, like the frog, do not? Why are  diurnal visual skills best
	achieved by foveated eyes? Why has not evolution led to develop human
	retina as a uniform structure composed of cones only---the higher resolution cells---instead
	of the variable resolution structure that is mostly based on rods?
	Apart from anatomical reasons, why has a non-uniform resolution been 
	developed?

\begin{quote}
	More than sixty years ago, the visual behavior of the frog was posing an interesting
	puzzle~\cite{4065609} which is mostly still on the table. 	
	The frog  will starve to death surrounded by food if it is not moving!
	His choice of food is determined only by size and movement, but he cannot 
	perceive information in still images. Interestingly, the frog doesn't focus
	attention like humans and most primates, who can very well perceive 
	also information in still images. Again, it could be the case that
	this different visual skills is rooted in fundamental information-based
	principles and that, in this case, the act of focus of attention plays
	a crucial role. 
\end{quote}

\item [{\bf Q8}]
What are the mechanisms that drive eye movements? 

\begin{quote}
	The mechanisms behind eye movements are likely optimizing the 
	acquisition of visual information. This is an exciting challenge, especially
	if you consider that the process of eye movement  is interwound with learning. While in the early
	stage of life, the information is mostly in details and moving objects, as 
	time goes by, humans focus on specific meaningful objects. 
	While focus of attention\index{focus of attention} helps object  perception, it's quite obvious that
	 also the opposite holds true, so as eye movements are likely driven
	 by an intriguing computational loop.
\end{quote}

\item [{\bf Q9}]
	Why does it take 8-12 months for newborns to achieve adult visual acuity?
	Is the development of adult visual acuity a biological issue or does it come 
	from higher level computational laws of vision?
	
\begin{quote}
	The  conquering of appropriate strategies for driving
	eye movements seems to expose a facet of a learning behavior that has been
	observed in newborns. Couldn't be the case that such a filtering 
	is motivated by the need of protection against information overloading?
	As newborns open their eyes they experiment a 
	visual stream where information extraction does require to capture
	very complex spatiotemporal regularities. It looks like such a 
	process is based on a sort of equilibrium so as children react 
	by appropriate visual skills to properly smoothed visual streams.
	Such an equilibrium can hardly be associated with nowadays 
	dominating gradient-based computations.
\end{quote}

\item [{\bf Q10}]
	How can we develop ``linguistic focusing mechanisms'' 
	that can drive the process of object recognition?

\begin{quote}
	The computational processes of vision that take place in nature
	can hardly be understood until we decouple them from language.
	The massive adoption of supervised learning seems to go exactly 
	in the opposite direction! Objects are perceived by linguistic 
	supervision, whereas vision in nature takes place mostly by
	unsupervised learning. Linguistic interactions can definitely 
	support the development of specific human visual competence,
	but they likely help also at developmental stages in which 
	most important and fundamental visual skills have been already
	achieved.	 
\end{quote}
\end{enumerate}

The rest of the book is mostly a travel driven by the need to address 
these questions. The pixel-wise computational processes stimulated in {\bf Q4}
along with {\bf Q3} on the need to introduce the natural notion of time 
suggest splitting the analysis in two parts dedicated to focus of attention
mechanisms and the investigation of motion invariance principle\index{motion invariance}. 
The corresponding conception of information-based laws of perception
discussed on Chapter~\ref{driving-principles}, along with the study of 
focus of attention and motion invariance lead to address all the above questions.

\begin{table}
\caption{Reference to  the section where we address the ten questions throughout the book.}
\centering
\begin{tabular}{c@{\qquad}c@{\quad}c@{\quad}c@{\quad}
c@{\quad}c@{\quad}c@{\quad}c@{\quad}c@{\quad}c@{\quad}c}
 \toprule
\textit{Question} & {\bf Q1}&  {\bf Q2}& {\bf Q3}& {\bf Q4}& {\bf Q5}& {\bf Q6}& {\bf Q7}& {\bf Q8}& {\bf Q0}& {\bf Q10}   \\
\midrule
 & 1.2 &  1.4 &  3.1 &  2.1 &  2.2 & 3.1 & 2.2   & 2.4 &  5.2 & 6 \\
 \textit{Sections}& 1.3 &  2.2 &   &   &  4.1 & 3.4 & 2.3  &  &  5.5 &  \\
 &  &   &   &  &  4.3 &  & 2.4 &  &  &  \\
 &  &   &   &   &  &  &  2.5 &  &   &  \\
\bottomrule
\end{tabular}
\label{wwq}
\end{table}

\chapter{Focus of attention}
\AtBeginShipoutNext{\AtBeginShipoutUpperLeft{%
  \put(1in + \hoffset + \oddsidemargin,-5pc){\makebox[0pt][l]{
\boxformat 
  \framebox{\vbox{
\hbox{Published by Springer, Cham---Cite this chapter:}
\smallskip
\hbox{\url{https://doi.org/10.1007/978-3-030-90987-1_2}}}
  }}}%
}}

\vspace{-4cm}
\begin{quote}
The frog does not seem to see or, at any rate, is not concerned with the detail of 
stationary parts of the world around him. He will starve to death surrounded by food if it is not moving. 
His choice of food is determined only by size and movement.\\
{\em J. Y. Lettvin, H. R. Maturana, W. S. McCulloch, and W. H. Pitts, 1959}
\end{quote}

\section{Introduction}
Why does it happen? This is definitely
a curious behavior, a somewhat surprising deficit for us to understand.
The frog can successfully catch flying insects, but when it's served with 
the food in an appropriate bowl starves to death! Hence, she doesn't see 
still images, which suggests that their interpretation is a more complex visual task than in case of moving objects. 
In order to address this puzzle, we start addressing Question no. 4 
on human capability of performing pixel semantic labeling. 
This propagates the interest to the exploration of the visual skills of
foveated animals and to the mechanisms of focus of attention\index{focus of attention}, that are
at the basis of the vision field theory herein discussed.\index{vision fields}
\marginpar{\small The velocity of the point of focus of attention is the first field of the theory}
The velocity of the point where the agent focuses its attention is in fact the 
first fundamental field that is involved in the proposed information-based laws of
perceptual vision. We begin the study of this field simply because it enables 
a full motion-based interaction in foveate-based animals, that is conjectured to be 
of fundamental importance also in machines. As we assume that there is an 
underlying process of eye movements we recognize the importance of continuously 
interacting with motion fields, even in the case of still images. As it will be pointed out 
in this chapter, this is in fact a remarkable difference with respect with the visual 
process taking place in animals, that like frogs, are based on significantly less 
effective visual skills.  In presence of eye movements with 
focus of attention
everything is moving. The external motion, which comes either in case of moving 
objects or in case of moving agent, is integrated with the internal motion 
of the agent (i.e. eye and/or head movements). As it will be seen in the rest 
of the book, this is of crucial importance in order to gain high level perceptual 
skills.
The chapter is organized as follow. In the next section we discuss how can 
humans perform  pixel semantic labeling, which directly leads us to consider 
the capability of focussing attention. In Section~\ref{Focus-section}, we give some
insights of the computational processes associated with focus of attention that
are based on the evolution of the animal visual system. In Section~\ref{WhyDoWeFAO}
we address the issue of the reasons why we focus attention and, finally, in
Section~\ref{WDeyeM} we study the driving mechanisms beyond eye movements.

\section{How can humans perform  pixel semantic labeling?}
\label{PixelwiseCompS}
Many computer vision tasks still rely on the pattern model, that is
based on an
opportune pre-processing of video information represented by a vector. 
Surprisingly enough, state of the art  approaches to object recognition already
offer quite accurate scene descriptions in specific real-world contexts, without  
necessarily relying on the semantic labelling of each
pixel. A global computational scheme emerges that is typically 
made more and more effective when the environment in which the machine is supposed to 
work is quite limited, and it is known in advance. In object recognition
tasks, the number of the classes that one expects to recognize in the environment
can dramatically affect the performance. 
\marginpar{\small Semantic pixel labelling as a fundamental visual primitive.}
Interestingly, very high accuracy can be
achieved without necessarily being able to perform the object segmentation and, therefore, 
without needing to perform pixel semantic labeling. 
However, for an agent to conquer visual capabilities in a broad context,
it seems to be very useful to rely on appropriate primitives. 
We humans can easily describe
a scene by locating the objects in specific positions and we can describe their eventual
movement. 
This requires a deep integration of visual and linguistic skills, that are
required to come up with compact, yet effective video descriptions. 
Humans semantic pixel labelling is driven by the focus of attention,
that is at the core of all important computational processes of vision.
While pixel-based decisions are 
inherently interwound  with a certain degree of ambiguity, they are remarkably
effective. The linguistic attributes that we can extract are related to 
the context of the pixel that is taken into account for label attachment, while
the ambiguity is mostly a linguistic more than a visual issue. 
In a sense, this primitive  is fundamental for conquering higher abstraction levels. How can this be done?
The focus on single pixels allows us to go beyond object segmentation 
based on sliding windows.
Instead of dealing with object proposals~\cite{DBLP:conf/eccv/ZitnickD14},
a more primitive task is that of attaching symbols to single pixels in the retina.
The bottom line is that human-like linguistic descriptions of visual scenes 
is gained on top of pixel-based features that, as a byproduct,
must also allow us to perform semantic labeling.  The task of semantic pixel labelling
leads to process the retina by focussing attention on the given pixel, while
considering the information in its neighborhood. This clearly opens the doors
to an in-depth re-thinking of computational processes of vision. It is not only the 
frame content, but also where we focus attention in the retina  that does matter. 

Human ability of exhibiting semantic labeling at pixel level is really challenging.
The visual developmental processes conquer this ability nearly without  
pixel-based supervisions. It seems that such a skill is mostly  the outcome of the 
acquisition of the capability to perform object segmentation. This is obtained by constructing 
the appropriate memberships of the pixels that define the segmented regions. 
When thinking of the classic human communication protocols, one early realizes that
even though it is rare to provide pixel-based supervision, 
the information that is linguistically conveyed to describe visual scenes makes implicit
reference to the focus of attention. This holds regardless of the scale of the visual entity 
being described. Hence, the emergence of the capability of performing pixel semantic
label seems to be deeply related to the emergence of focus of attention mechanisms. 
The most striking question, however, is how can humans construct such a 
spectacular segmentation without a specific pixel-based supervision! Interestingly,
we can focus on a pixel and attach meaningful labels, without having been 
instructed for that task.
\marginpar{\small {\bf Q4}: Why semantic pixel labelling?}
\begin{svgraybox}
The primitive of pixel semantic
labelling is likely crucial for the construction of human-like visual skills. 
There should be a hidden supervisor in nature 
that, so far, has nearly been neglected.
We conjecture that it is the optical flow\index{optical flow}
which plays the central role 
for object recognition. The decision on its recognition must be invariant under
motion, a property that does require a formulation in the temporal direction.
The capability of focussing on $(x,t)$ seems to break the circularity of the 
mentioned chicken-egg dilemma. As it will be seen in the following, 
the local reference to pixels enables the statement of visual constraints
on motion invariance that must be jointly satisfied.
\end{svgraybox}
%

\section{Insights from evolution of the animal visual system}
\label{Focus-section}
It is well-known that the presence of the fovea in the retina leads to focus attention on details in the scene.
\marginpar{\small Mammals and haplorhine primates}
Such a specialization of the visual system is widespread 
among vertebrates, it is present in some snakes and fishes, 
but among mammals is restricted to haplorhine primates.
In some nocturnal primates, like the owl monkey and in the tarsier, the fovea is morphologically distinct 
and appears to be degenerate. Owl monkey's  visual system is somewhat different from other monkeys and apes. 
As it retina develops, its dearth of cones and its surplus of rods mean
that this focal point never forms. 
Basically, a fovea is most often found in diurnal animals, thus supporting the idea that it is supposed to play an
important role for capturing details of the scene~\cite{Ross2004}.
But why haven't many mammals developed such a rich vision system
based on foveated retinas? 
Early mammals, which emerged in the shadow of the dinosaurs, were likely forced to 
nocturnal lives, so as to avoid to become their prey~\cite{PMID:32269365}.
\marginpar{\small The long nocturnal evolution of mammals' eyes and the 
bottleneck hypothesis}
In his seminal monograph,  Gordon Lynn Walls~\cite{Walls1942}
conjectured that there has been a long nocturnal evolution of mammals' eyes, which
is the reason of the remarkable differences with respect to those of other vertebrates. 
The idea became known as the ``nocturnal bottleneck'' hypothesis~\cite{Gerkema_2013}.
Mammals' eyes tend to resemble those of nocturnal birds and lizards, 
but this 
does not currently hold for humans and closely related monkeys and apes. 
It looks they re-evolved features useful
for diurnal living after they abandoned a nocturnal lifestyle upon dinosaur extinction. 
It is worth mentioning that  haplorhine primates are not the only mammals which focus attention 
in the visual environment. Most mammals have quite a well-developed visual system for
dealing with details. For example, it has been shown that dogs possess quite a good visual
system that share many features with those of haplorhine primates~\cite{10.1371/journal.pone.0090390}. 
A retinal region with a primate fovea-like cone photoreceptor density has been identified
but without the excavation of the inner retina. Similar anatomical structure observed in rare human subjects has been named fovea-plana. 
Basically, the results found in~\cite{10.1371/journal.pone.0090390} challenge the
 dogma that within the phylogenetic tree of mammals, haplorhine primates with a fovea are the sole lineage in which the retina has a central bouquet of cones. 
\marginpar{\small Fovea versus area centralis}
In non-primate mammals, there is a central region of
specialization, called {\em area centralis}, which is often located close to the optic axis and demonstrates a local increase in
photoreceptor and retinal ganglion cell density
that plays a somehow dual role with respect to the fovea.
Like in  haplorhine primates, in those non-primate mammals we experiment 
focus of attention mechanisms that are definitely important from a functional viewpoint. 

%
%
This discussion suggests that the evolution of animals' visual system has followed many different paths
that, however, are related to focus of attention mechanisms, that are typically more effective for 
diurnal animals. There is, however, an evolution path which is definitely set apart, in which
the frog is most classic representer. 
\marginpar{\small The frog dilemma}
More than sixty years ago, the visual behavior of the frog was posing an interesting
puzzle~\cite{4065609} which is mostly still on the table. In the words of the authors:
\begin{quote}{\em
The frog does not seem to see or, at any rate, is not concerned with the detail of 
stationary parts of the world around him. He will starve to death surrounded by food if it is not moving. 
His choice of food is determined only by size and movement.
}\end{quote}
No mammal experiment shows such a surprising behavior! However, the frog is not expected
to eat like mammals. When tadpoles hatch and get free, they attach themselves to plants 
in the water such as grass weeds and cattails. 
They stay there for a few days and eat tiny bits of algae. 
Then the tadpoles release themselves from the plants and begin to swim freely, 
searching out algae, plants and insects to feed upon. At that time their visual system 
is ready. Their food requirements are definitely different from what mammals need
and their visual system has evolved accordingly for catching flying insects. 
Interestingly, unlike mammals, the studies in~\cite{4065609}
already pointed out that the frogs' retina is characterized by
uniformly distributed receptors with neither fovea nor area centralis.
Interestingly, this means that the frog doesn't focus attention by eye movements.
 
 \marginpar{\small The frog has got uniformly distributed receptors with neither fovea nor
 area centralis}
 One can easily argue that any action that animals carry out 
needs to prioritize the frontal view.
On the other hand, this leads to the detriment of the peripheral vision, that 
is also very important. In addition, this could apply for the dorsal system whose
neurons are expected to provide information that is useful to support movements
and actions. Apparently, the ventral mainstream, with neurons 
involved in the ``what'' function, does not seem to benefit from foveated eyes.
Nowadays most successful computer vision models for object recognition, just like frogs,
use a uniformly distributed retina and do not 
focus attention. 
However, unlike frogs, machines seem to conquer human-like recognition capabilities 
on still images.
\marginpar{\small Unlike frogs, convolutional deep nets recognize the food served in the bowl,
but only thanks to the truly artificial supervised learning protocol!}
 Interestingly, unlike frogs, nowadays machines recognize quite well food 
properly served in a bowl.   These capabilities might be
due to the current strong supervised communication protocol. Machines benefit from
tons of supervised input/output pairs, a process which, as already pointed out, cannot
be sustained in nature. 
On the other hand, as already pointed out, in order to attack the task of understanding 
what is located in a certain position, it is natural to think of  eyes based on fovea or on 
area centralis. The eye movements with the corresponding trajectory
of the focus of attention (FOA) is also clearly interwound with the temporal structure of 
video sources. 
In particular, humans experiment eye movements when looking at fixed objects, 
which means that they continually experiment motion. Hence, also in case of fixed images, 
conjugate, vergence, saccadic, smooth pursuit, and vestibulo-ocular movements 
lead to acquire visual information from relative motion.  We claim that the 
production of such a continuous visual stream naturally drives
feature extraction,
since the corresponding convolutional filters, charged of representing features for
object recognition, are expected to provide consistent information during motion. 
The enforcement of this consistency condition creates a mine of visual data
during animal life! Interestingly, the same can happen for machines. 
Of course, we need to compute the optical flow at pixel level so as to enforce
the consistency of all the extracted features. Early studies 
on this problem~(see e.g.~\cite{HornAI1981}), along with recent related 
improvements (see e.g.~\cite{Baker:2011})
suggests to determine the velocity field by enforcing brightness invariance. 
\marginpar{\small Couldn't focus of attention involve motion processes capable of
generating a mine of supervised information?}
As the optical flow is gained, it can be used to enforce motion consistency on the
visual features.
These features can be 
conveniently combined and used to recognize objects.
Early studies driven by these ideas are  reported in~\cite{DBLP:journals/cviu/GoriLMM16},
where the authors propose the extraction of visual features as a
constraint satisfaction problem, mostly based on information-based principles
and  early ideas on motion invariance. 

\section{Why focus of attention?}
\label{WhyDoWeFAO}
\marginpar{\small Focus of attention as an information-based process}
In this section we mostly try to address Question no. 7 on the reason why some animals 
focus attention. As already pointed out, first of all, it looks like focus of 
attention creates a uniform framework in which visual information comes
from motion. 
There are a number of surprising convergent
issues that strongly support the need for focus of attention mechanisms for conquering 
top level visual skills that is typical of diurnal animals.
Basically, it looks like we are faced with functional issues which mostly obey
information-based principles that hold regardless of the body of the agent.
\begin{enumerate}
\marginpar{\small Eight reasons for FOA}
\item {\em The FOA drives the definition of  visual primitives at pixel level}\\
	The already mentioned visual skill that humans possess to perform pixel
	semantic labeling clearly indicates their capability of focusing on specific 
	points in the retina with high resolution.  Hence, FOA is needed if we want to 
	perform such a task. Another side of the coin does reflect the underlying
	assumption of understanding perceptual vision on the basis of a 
	field theory, where features and velocities are indissolubly paired. \\

\item{\em Variable resolution retina}\\ 
\label{VariableRR}
	Once we focus on a certain pixel, the computation of the associated
	visual fields, which might be semantically interpreted,  benefits from 
	a variable spatial resolution that is decreasing as we move far away from the
	point of the focus. This seems to be quite a controversial issue. 
	One can easily argue that given the same overall resolution,
	while we see better close to the focus, we have lower peripheral 
	visual skills. However, this claim makes sense if we consider only
	the process which takes place on single frames and we disregard
	the temporal dimension. Basically,  visual processes in Nature 
	come with a certain velocity, an issue which  makes it possible 
	to fool human eyes in the video production. Hence, if we restrict 
	the frame rate to classic cinema ratio ($24$ frames/sec) 
	then a more dense presence of cones, where the eye is focussing
	attention, makes it possible to use acquisitions at high resolution
	in different regions of the retina as a consequence of saccadic movements.
	It looks like we need a trade off between the velocity of the scan path
	in focus of attention and the spatial distribution of the resolution. 
	Clearly, as we increase the degree of different resolution in the retina
	we need to increase consequently the velocity of the scan paths.
	While this discussion offers a convincing description of the 
	appropriate usage of the different resolution, it doesn't 
	fully address {\bf Q7} on why hasn't evolution increased 
	the resolution also at the periphery in foveated animals. 
	This will issue be addressed in Sec.~\ref{obj-rec-sec}.

\item	{\em Eye movements and FOA help estimating the probability distribution on the retina}\\
	At any time a visual agent clearly needs to possess a good estimation of the
	probability distribution over the pixels of the retina. This is important whenever 
	we consider visual tasks for which the position does matter. 
	This involves both the {\em where} and {\em what} neurons of the dorsal and ventral
	mainstream~\cite{GoodaleMilner92}. In both cases
	it is quite obvious that any functional risk associated with the given task 
	should avoid reporting errors in regions of the retina where there is a uniform
	color. The probability distribution is of fundamental importance and it is definitely
	related to saliency maps that can be gained by FOA.\\
	
\item {\em Eye movements and FOA: There's always motion!}\\
	The interplay between the FOA and motion invariance\index{motion invariance} properties is the 
	key for understanding human vision and general principles that drive 
	object recognition and scene interpretation. In order to understand the 
	nice circle that is established during the processes of learning in vision, 
	let us start exploring the very nature of eye movements in humans. 
	Basically, they produce visual sequences that are separated by saccadic
	movement, during which no information is acquired. 
	In the case of micro-saccades 
	the corresponding micro-movements explore regions with a remarkable
	amount of details that are somehow characterized  by certain   features.
	The same holds true for smooth pursuit. No matter what kind of movement is
	involved, apart from saccadic movements, the visual information is always
	paired with motion, which leads to impose invariances on the 
	extracted feature\footnote{An in-depth coverage of this issue is in Ch.~\ref{MInv-sec}.}. 
	This indicates a remarkable difference between different animals;
	the frog doesn't produce such a movement in presence of still images!
	It is in fact the presence of eye movements with FOA which somehow unifies the
	visual interactions, since we are always in presence of motion.
	Clearly, the development of such a motion invariance in this case does exploit
	much more information and can originate skills that are absent in some animals.
	Basically, the focus of attention mechanisms originate an impressive amount of information
	that, in this case, Nature offers for free! 
	 As it will be pointed out in Ch.~3, this is very important 
	for the information that arises from motion invariance principles.
	\\

\item {\em Saccades, visual segments, and ``concept drift''}\\ 
	The saccadic movements contribute to perform ``temporally segmented computations''
	over the retina on the different sequences produced by micro-saccadic movements.
	When discussing motion invariance we will address the problems connected 
	with  {\em concept drift}, where an object is slightly transformed  into another one. 
	Clearly, this  could dramatically affect the practical implementation of the 
	motion invariance. For example, the child's Teddy Bear could be slowly transformed into a nice dog.	
	Preserving motion consistency would lead to confuse bears with dogs!
	 However, amongst different types of FOA trajectories,
	the saccadic movements play the fundamental role of resetting the 
	process, which clearly faces directly problems of concept drift.	\\

\item {\em FOA helps disambiguation at learning time}\\  
\label{DisambiguatingLT}
	A puzzle is offered at learning time when two or more instances of the 
	same object are present in the same frame, maybe with different poses
	and scales. The FOA in this case helps disambiguating 
	the enforcement of  motion invariance. While the enforcement of weight sharing
	is ideal for directly implementing translation equivariance, such a constraint
	doesn't facilitate other more complex invariances that can better be achieved
	by its removal. As it will be more clear in Ch.~\ref{driving-principles},
	the removal of the weight constraint adds one more tensorial dimension
	with significant additional space requirements. However, we shall see 
	appropriate mechanisms for efficiently facing this. 
	Basically the constraint of motion invariance under FOA movements
	naturally disambiguates the processing on different instances of the same 
	object since the computation takes place, at any time, on a single instance.\\	

\item {\em FOA drives the temporal interpretation of scene understanding}\\
    Animals which focus attention receive information from the environment which 
    is in fact driven and defined by the FOA. In particular, humans acquires information
    on the object(s) on which they are focusing attention. This is a simple, yet powerful
    mechanism for receiving supervision. Without such pointing mechanism, information on 
    the current frame cannot refer to a specific position, which makes learning more difficult.
	The importance of FOA involves also the interpretation of visual scenes. It is in fact the
	way FOA is driven which sequentially selects the information 
	for the scene interpretation. Depending on the purpose of the
	agent and on its level of scene understanding the FOA is
	consequently driven.
	This process clearly shows the fundamental role on the selection of the 
	points where to focus attention, an issue which is described in the following
	section. Once again, we are back to the issue of the limited number of frames/sec
	which characterizes natural video. The scene interpretation has its own 
	dynamics which very well fits the corresponding mechanisms of focus of 
	attention.\\
	 
\item {\em FOA helps disambiguating illusions}\\
	Depending on where an agent with foveated eyes focuses attention 
	concepts that, strictly speaking, don't exist can emerge, thus originating
	an illusion. A noticeable example is  the 
	Kanizsa's triangle, but it looks like other illusions arise for 
	related reasons.  
	You can easily experiment that as you approach any detail, 
	it is perfectly perceived without any ambiguity. 
	A completion mechanism arises that leads us to perceive the
	 triangle  as soon as we move away from the figure and the mechanism 
	is favored by focussing attention on the barycenter. 
	Interestingly, the different views coming from different points
	where  an agent with foveated eyes focuses attention 
	likely helps disambiguating illusions, a topic that
	has been recently studied in classic convolutional 
	networks~\cite{DBLP:journals/corr/abs-1903-01069,
	DBLP:conf/cogsci/BakerEKL18}.
	Kanizsa's triangle is not a special case of illusion. There's a 
	huge literature on related cases and, even the Barber's pole~\cite{Fisher2001}
	that is discussed in the following is an intriguing example of 
	the role of focus of attention in the different interpretations that
	can be offered of the same object.\\
\end{enumerate}

\marginpar{\small Question no. 7 on motivations for focus of attention}
The analysis on foveated-based neural computation helps understanding also the reason why 
humans cannot see video with a number of frames per second that exceeds the classic
sampling threshold. It turns out that this number is clearly connected with the velocity of the
scan paths of the focus of attention. Of course, this is a computational issue which
goes beyond biology and clearly affects machines as well. 

	\begin{svgraybox}
	The above items provide strong evidence on the 
	reasons why  foveated eyes turn out to be very effective for 
	scene understanding.
	Interestingly, we can export the information-based  principle of focussing attention
	to computer retinas by simulating eye movements. There is more: 
	machines could provide  multiple focuses  of attention which could 
	increase their visual skills significantly.
	\end{svgraybox}

\section{What drives eye movements?}
Foveated animals need to move their eyes to properly focus attention. 
Human eyes make jerky saccadic movements during ordinary visual 
acquisition.  One reason for these movements is that the fovea
provides high-resolution in portions of about $1,2$ degrees.
\label{WDeyeM}
\marginpar{\small A quick tour in the literature}
Because of such a small high resolution portions, the overall sensing of a scene
does require intensive movements of the fovea. Hence, the fovea movements
do represent a good alternative to eyes with uniformly high resolution 
retina. The information-based principles  discussed so far lead us to conclude that foveated
retinas with saccadic movements is in fact a solution that is 
computationally sustainable and very effective.
Fast reactions to changes in the surrounding visual environment require efficient attention mechanisms to reallocate computational resources to most relevant locations in the visual field. 

Visual attention plays a central role in our daily activities. While we are
playing, teaching a class or driving a vehicle, the amount of
information our eyes collect is way greater than what  we are able
to process~\cite{allport1989visual,koch2006much}. To work properly, we need a
mechanism that only locates the most relevant objects, thus optimizing the computational resources~\cite{ungerleider2000mechanisms}. Human visual attention performs this task so efficiently that, at conscious level, it goes unnoticed.

Attention mechanisms have been the subject of massive investigation also in machines, 
especially whenever they are asked to solve tasks related with human perception such as  video compression, 
where loss of quality is not perceivable by viewers~\cite{itti2004automatic,hadizadeh2013saliency}, 
or caption generation~\cite{liu2017attention,chen2018boosted}. Following the seminal works by Treisman et al.~\cite{treisman1980feature,treisman1969strategies} and Koch and Ullman~\cite{koch1987shifts}, as well as
the first computational implementations~\cite{itti1998model}, over the last three decades scientists have presented numerous attempts to model FOA~\cite{borji2012state}. 
The notion of \textit{saliency map} has been introduced, which consists
of a spatial map that indicates the probability of focusing on each pixel.   However, attention is essentially a dynamical process and neglecting the temporal dimension may critically lead to a poor description of the phenomenon~\cite{boccignone2019problems}. 
%
Under the \textit{centralized saliency map hypothesis}, shifts in visual attention can be 
generated through a winner-take-all mechanism~\cite{koch1987shifts}, selecting the most relevant location in space at each time step. 
Still, the temporal dynamics of the human visual selection process is not considered. Some authors have tried to fill this gap with different approaches, for example by preserving the centrality of the saliency map introducing a
hand-crafted human bias to choose subsequent fixations~\cite{le2015saccadic}.
Similarly,~\cite{jiang2016learning} tries to formalize the idea that during visual exploration top-level cues continue to increase their importance, to the disadvantage of more perceptive low-level
information. In~\cite{khosla2007bio} the authors
propose a bio-inspired visual attention model based on the pragmatic
choice of certain proto-objects and learning the order in which these are attended. 
All of these approaches still assume the centrality of a saliency map to perform a long stack of global computations
over the entire visual field before establishing the next fixation
point. This is hardly compatible with what is done by humans where, most likely, attention modulates visual signals before they even reach cortex~\cite{briggs2007fast, mcalonan2008guarding} and restricts computation to a small portion 
of the available visual information~\cite{treisman1980feature, schlingensiepen1986importance}.
In~\cite{boccignone2004modelling}, the gaze shift is modeled as the realization of a stochastic process. A saliency map is used to represent a landscape upon which non-local transition probabilities generate a constrained random walk. 

\marginpar{\small Virtual masses and gravitational models}
More recently, Zanca et al. proposed an approach that is inspired by physics to model the process of visual attention as a continuous dynamic phenomenon \cite{zanca2017variational,zanca2019gravitational}. 
The focus of attention is modeled as a point-like particle gravitationally
attracted by 
virtual masses originated from details and motion in the visual scene. Masses due to details are determined by the magnitude of the gradient of the brightness, while masses due to motion are proportional to the norm of the optical flow. This framework can be applied to both images and videos, as long as one considers a static image as a video whose frames are repeated at each time step. Moreover, the model proposed in~\cite{zanca2019gravitational} also implements the inhibition of return mechanism, by  decreasing the saliency  of a given area of the retina that has already been explored in the previous moments. Unlike the other described approaches, the prediction of the focus does not rely on a centralized saliency map, but it acts directly on early representations of basic features organized in spatial maps. 
Besides the advantage in real-time applications, these models make it possible to characterize patterns 
of eye movements (such as \textit{fixations}, \textit{saccades} and \textit{smooth pursuit}) and, despite their simplicity, 
they reach the state of the art in scanpath prediction and proved to predict shifts in visual attention better than the classic winner-take-all~\cite{zanca2020gravitational}. 
\marginpar{\small Gravitational models are non-local in space and, therefore, are not biologically
plausible}
However, when looking at these gravitational models
from the biological and computational perspective, one promptly realizes that finding the focus of attention
at a certain time does require the access to all the visual information of the retina to sum up the attraction arising from any virtual mass. Basically, those models are 
not local in space.


While current computational models keep improving their predictive ability thanks to the increasing availability of data, 
they are still far away from the effectiveness and efficiency exhibited by foveated animals. 
An in-depth investigation on biologically-plausible computational models of focus of attention 
that exhibit spatiotemporal locality is very important also for computer vision, where one
relies on parallel and distributed implementations. 
The ideas given in~\cite{DBLP:journals/corr/abs-2006-11035}, properly re-elaborated, are
herein proposed mostly because of their effectiveness and biological plausibility.
\marginpar{\small Visual stimuli corresponding to spatial details and motion turns out be act as 
sources for the focus of attention field and originate wave propagation which, unlike 
for the gravitational field, has a finite propagation velocity}
A computational model is proposed where
attention emerges as a wave propagation process originated by visual stimuli corresponding to details and motion information. 
The resulting field obeys the principle of {\em inhibition of return},
so as not to get stuck in potential holes, and 
extend previous studies in~\cite{zanca2019gravitational} with the main objective of providing 
spatiotemporal locality. In particular,  the idea of modeling the focus of attention by a gravitational process 
finds its evolution in the corresponding local model based on Poisson equation on the corresponding potential. 
Interestingly, Newtonian gravity yields an instantaneous propagation of signals, so as a sudden change in the mass density of a given pixel immediately affects the focus of attention, regardless of its location on the retina. 
These studies are driven by the principle that there are in fact sources which drive attention (e.g. masses in 
a gravitational field). 

\begin{figure}[H]
	\centering
	\includegraphics{./figs/fig-3.mps}
\caption{Gravitational model of  focus of attention. The cognitive process of focus of attention 
follows the model of mass attraction, which turn out to be the translation of spatiotemporal visual details.
Here we have  displayed a
general density distribution and
the corresponding 2D potential.}
\label{NeuronModel-Fig}
\end{figure}


Here we discuss a paradigm-shift in the computation of the attraction  model
proposed in~\cite{zanca2019gravitational}, which is inspired by the 
classic link between global gravitational or electrostatic forces and the associated
Poisson equation on the corresponding potential, that can be regarded as a 
spatially local computational model. 
While the link is intriguing, modeling the focus of attention by the force emerging from 
the static nature of the Poisson potential does not give rise to a truly local 
computational process, since one needs solving the Poisson equation for
each frame. 
This means that such a static model is still missing the temporal propagation that takes place
in peripheral vision mechanisms. We show that the temporal dynamics which arises from
diffusion and wave-based mechanisms are effective to naturally implement local 
computation in both time and space. The intuition is that attention is also driven by 
virtual masses that are far away from the current focus by means of wave-based and/or diffusion
propagation. We discuss the two different mechanisms of propagation and prove their 
reduction to gravitational forces as the velocity of the propagation goes to infinity. The experimental results confirm that the information coming from virtual masses is properly transmitted on the entire retina. In the case of video signals, our experimental analysis on scanpaths leads to state of the art results which can clearly 
be interpreted when considering the reduction to the gravitational model ~\cite{zanca2019gravitational} for infinite
propagation velocity. The bottom up is that we can reach state of the art results by
a computational model which is truly local in space and time.
As a consequence, the proposed model is very well-suited for Single Instruction 
Multiple Data (SIMD) specialized implementations.
Moreover, the proposed theory on focus of attention 
sheds light  on the way it takes place in biological processes~\cite{marr1976understanding}.


\marginpar{\small FOA driven by gravitation that is characterized by the corresponding potential $\varphi^0$}
According to the gravitational model of ~\cite{zanca2019gravitational}, the trajectory of  the focus of attention
$t\in[0,T]\mapsto a(t)\in \bbR^2$, starting at $a(0)=a_0$ with velocity
$\dot a(0)=a_1$, is the solution of the following Cauchy problem:
\begin{equation}\label{FOA-trajectory}
\begin{cases}
\ddot a(t)+\varpi\dot a(t)+\nabla u^0(a(t),t)=0;\\
a(0)=a_0;\\
\dot a(0)=a_1,
\end{cases}
\end{equation}
where $\varpi>0$ and the scalar function $u^0\colon \bbR^2\times[0,T]
\to
$ is
defined as follows,
\begin{equation}\label{grav-potential}
u^0(x,t):={\frac{1}{2\pi}}\int_{\bbR^2}\log{\frac{1}{\lVert x-y\rVert}}\mu(y,t)\,
dy.
\end{equation}
Here $\lVert x-y\rVert$ is the Euclidean norm in $\bbR^2$ and $\mu\colon
\Omega\subset\bbR^2\times [0,T]\to[0,+\infty)$ is the mass distribution at a
certain temporal instant that is present on the retina and that is responsible 
of ``attracting'' the focus of attention. Such mass distribution involves details and motion, 
and, formally, it is determined by
\begin{equation} \label{eq:mass_density}
\mu(x,t) = \mu_1(x,t)\left(1-I(x,t)\right)+\mu_2(x,t).
\end{equation} 
\marginpar{\small The role of the inhibition of return function $I$ which drives the change 
of the virtual mass}
In particular, $\mu_1 = \alpha_1 \lVert\nabla b \rVert$, where $b\colon\bbR\times
[0,T]\to\bbR$ is the brightness, while $\mu_2 = \alpha_2 \lVert \partial_{t} b \rvert$,
while $\alpha_1$ and $\alpha_2$ are positive parameters.  The term
$I(x,t)$ implements the inhibition of
return mechanism, and it satisfies 
\begin{align}
	I_t+\beta I = \beta \exp(-\lVert x-a(t)\rVert^2/2\sigma^2),
\label{IoR-eq}
\end{align}
with $0<\beta<1$ ($I_t$ is the time derivative of $I$).
We can promptly see that this model for $I$ nicely implements the mechanism for 
avoiding to get stuck in the focus of attention. If $x$ approaches $a(t)$ then the system dynamics
of $I_{t}$, leads to $I_{t} \simeq 1$, with a velocity which 
depends on $\beta$. As $I_{t} \simeq 1$, from Eq.~(\ref{IoR-eq}) we 
see that the component of the virtual mass coming from spatial 
details disappears, that favoring the escape from the focus of attention.
Because of the need to know the distance from the focus of attention
from Eq.~(\ref{grav-potential}), we can see that the model of \cite{zanca2019gravitational}
is not local in space.

Notice that the
potential $u^0$ satisfies the Poisson equation on $\bbR^2$:
\begin{equation}\label{Poisson-equation}
-\nabla^2u =\mu,
\end{equation}
where $\nabla^2$ is the {\it Laplacian} in two dimensions. 
Such result, which is the
two-dimensional analogue of the Poisson equation for the classical
gravitational potential,  can be checked by direct calculation. 
%
%
Because the mass density $\mu$ is time dependent, and its temporal
dynamics is synced with the temporal variations of the video,
Eq.~\eqref{Poisson-equation} should in principle be solved for any $t$.
We will discuss how the values
of the potential in a spatial neighbour of $(x,t)$ are exploited to estimate the values of
$u$ at $(x,t+dt)$ by interpreting Eq.~\eqref{Poisson-equation} as
an elliptic limit of a parabolic or hyperbolic equation instead.

Evaluating the gravitational force acting on the FOA starting with the Poisson equation requires to compute the potential $u(x,t)$ due to the virtual masses
at each frame, thus ignoring any temporal relation. This key remark also underlines the strong limitation of the solution proposed 
in~\cite{zanca2019gravitational} (and also in~\cite{zanca2017variational}), where the gravitational force is
re-computed at each frame from scratch. 
The main idea behind the reformulation presented here is
that since we expect that small temporal changes in
the source $\mu$  cause small changes in the solution $u$, then
it is  natural to model the potential $u$ by dynamical
equations which prescribe, for each spatial point $x$, how the solution must
be updated depending on the spatial neighborhood of $x$ at time $t-dt$.  

\marginpar{\small Biologically plausible wave propagation of FOA}
We can introduce an explicit temporal
dynamics in Eq.~\eqref{Poisson-equation}
by introducing  the two following ``regularizations'' 
\begin{equation}
\begin{aligned}
&
H:
\begin{cases}
c^{-1}u_t=\nabla^2u+\mu& \hbox{in}\quad \bbR^2\times(0,+\infty);\\
u(x,0)=0,& \hbox{in}\quad \bbR^2\times\{t=0\},
\end{cases}\\
&
W:
\begin{cases}
c^{-2}u_{tt}=\nabla^2u+\mu& \hbox{in}\quad
\bbR^2\times(0,+\infty);\\
u(x,0)=0,\quad u_t(x,0)=0& \hbox{in}\quad \bbR^2\times\{t=0\},
\end{cases}
\end{aligned}
\label{eq:temp-reg}
\end{equation}
where $c>0$ and $u_t$ ($u_{tt}$) is the first (second) time derivative of $u$.
Problem $H$ 
is a Cauchy problem for the heat equation with source $\mu(x,t)$,
whereas problem $W$ is a Cauchy problem for a wave equation.
The term $c$ in $H$ represents the diffusivity constant, whereas
the constant $c$ in problem $W$ can be regarded as the wave propagation velocity.
The reason why we can consider problem $H$ and $W$ as {\em temporal
regularizations} of Eq.~\eqref{Poisson-equation} is due to a  
fundamental link~\cite{DBLP:journals/corr/abs-2006-11035} between the gradients $\nabla u_H$ and
$\nabla u_W$ of the solutions $u_H$ and $u_W$ to  problems
$H$ and $W$ in Eq.~\eqref{eq:temp-reg} with 
$u^0$, which is  the solution, described in Eq.~\eqref{grav-potential}.
Those gradients for  $H$ and $W$ both converge to $\nabla u^0$ as $c\to+\infty$.
%
The interpretation of this result is actually quite straightforward. For
problem $H$ it means that the solution of the heat equation in a substances
with high diffusivity $c$, instantly converges to its stationary
value which is given by Poisson equation~\eqref{Poisson-equation}.
For problem $w$, the above convergence statement turns out to be the two dimensional
analogue of the infinite-speed-of-light limit in electrodynamics and  in
particular it expresses the fact that the retarded potential
(see~\cite{jackson2007classical}), which in
three spatial dimensions are the solutions of problem $W$, converges to the
electrostatic potential as the speed of propagation of the wave goes to
infinity ($c\to+\infty$).
\footnote{It is worth mentioning that while this kind of regularization
is well-known in three dimensions, the same properly has not been formally
stated in two dimensions. 
A formal proof of the property in the case of 
two dimensions is given in~\cite{DBLP:journals/corr/abs-2006-11035}.}

Although both temporal regularization $H$ and $W$ achieve the goal of
transforming the Poisson equation into an initial value problem in time from
which all subsequent states can be evolved from, the different nature of
the two PDE determines, for finite $c$, qualitative differences in the
FOA trajectories computed using Eq~\eqref{FOA-trajectory}. In the remainder, we then consider the following generalized version of Eq.~\eqref{eq:temp-reg}
\begin{equation} \label{eq:dumped-wave}
{\begin{cases}
\gamma u_{tt}(x,t)+\lambda u_t(x,t) = 
c^2 \nabla^2 u(x,t) + \mu(x,t)& \hbox{in}\quad
\bbR^2\times(0,+\infty);\\
u(x,0)=0,\quad u_t(x,0)=0& \hbox{in}\quad \bbR^2\times\{t=0\},
\end{cases}}
\end{equation}
where $\lambda\geq 0 $ is the drag coefficient and $\gamma\geq 0 $. Such equation in one spatial dimension (and without the source term
$\mu$) is known as the telegraph equation (see~\cite{evans2010partial}).
More generally, it describes the propagation of a damped wave. The pure diffusion case $H$ corresponds to $\gamma = 0$, with a diffusion coefficient equals to $\alpha =c^2/\lambda$ and a source term of $\mu(x,t)/\lambda$. With $\gamma =1$ and $\lambda = 0$ we obtain the pure wave equation $W$ instead. The FOA model proposed here is based on equations ~\eqref{eq:dumped-wave} along with the
inhibition of return equation expressed by~\eqref{IoR-eq}.

Clearly, Eq.~\eqref{eq:dumped-wave} is local in both space and time, which is 
a fundamental ingredient of biological plausibility. In addition, they are very well-suited
for SIMD hardware implementations. At a first sight, Eq.~\eqref{IoR-eq} does not 
possess spatial locality. While this holds true in any computer-based retina, 
in nature, moving eyes rely on the principle that you  can simply 
pre-compute $\exp(-\lVert x-a(t)\rVert^2/2\sigma^2)$ by an appropriate foveal structure. 
Interestingly, the implementation of moving eyes are the subject of remarkable interest
in robotics  for different reasons (see e.g.~\cite{MRucci-2012}).
\marginpar{\small Question no. 8: What drives focus of attention?}
\begin{svgraybox}
Driving the focus of attention is definitely a crucial issue. 
We conjecture that this driving process 
must undergo a developmental process, where we begin with details and 
optical flow and we carry on with the fundamental feedback from the 
environment which is clearly defined by the specific purpose of the agent.
\end{svgraybox}

\section{The virtuous loop of focus of attention}
In haplorhine primates the process of learning to see is interwound with that of focussing attention. 
We can access to their strict conjugation by cognitive statements but, 
as it will become more clear in the next chapter, we can also give such a tight 
connection an in-depth mathematical formulation.\index{feature conjugation}

\marginpar{\small Where do you focus attention in the early stages of life?}
During early cognitive stages,  in newborns attention mechanisms are mostly driven 
by the presence of details and movements. 
However, this is not the whole story. 
Humans are endowed with an exceptional ability for detecting faces
and, already shortly after birth, they preferentially orient to faces.
Simple tests of preference have been carried out which
provide evidence on the fact that  infants as young as newborns 
prefer faces and face-like stimuli over distractors~\cite{Lee:2009}.
However, the issue is quite controversial and there is 
currently no agreement as to how specific or general 
are the mechanisms underlying newborns' face preferences~\cite{Farroni17245}. 
%
%
The neural substrates underlying this early preference is also quite an important subject of
investigation. In~\cite{Vallorticara2019}, the authors point out that 
measured EEG responses in 1- to 4-day-old infants and discovered 
reliable frequency-tagged responses when presenting faces.
Upright face-like stimuli elicited a significantly stronger frequency-tagged response 
than inverted face-like controls in a large set of electrodes. 
Overall, their results suggest that a sort of cortical route specialization  is already functional 
at birth for face perception.
%
%
This specialization makes sense. In other species of animals, the  
degree of visual skills that are exhibited at the birth are remarkably 
higher than in humans, which is clearly connected with very different 
environmental constraints. From the viewpoint of information theory,
the degree of visual skills that are exhibited at the birth are related to the
genetic heritage. There is an interesting tradeoff in nature between the 
transmitted visual skills and the transmitted mechanisms for their learning.
Focus of attention doesn't escape this general rule, but it seems to 
be strongly biased towards the learning component, so as  
early  behavior is quite elementary. This is the reason why we stressed
the analysis of a model based on the magnitude of the spatiotemporal gradient of the brightness. 
\begin{figure}[H]
	\centering
	\includegraphics{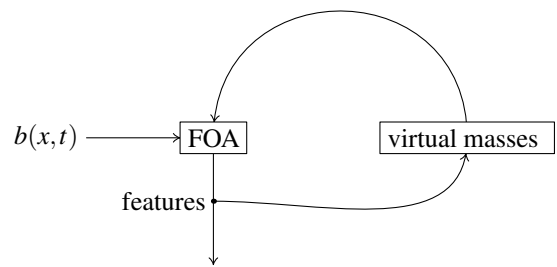}
\caption{The virtuous loop of focus of attention. After a phase in which FOA
is driven by visual details that are regarded as virtual masses, as time goes by, the 
development of visual features results in new virtual masses that also provide focus
of attention by means of the same attraction mechanisms.}
\label{NeuronModel-Fig}
\end{figure}
In children the mechanisms that drive  the focus of attention are strongly connected 
with the developmental stages. As already stated, at  early stages of life, they  only focus 
attention on details and movements with some preference for  visual patterns like faces. 
As time goes by, visual features conquer a semantic value and, consequently, the focus of attention
is gradually driven by specific intentions and corresponding plans. Of course, this is only possible 
after having acquired some preliminary capability of recognizing objects. Interestingly, as the
forward process that facilitates high level cognitive tasks from the focus of attention becomes effective
a corresponding backward process begins the improvement of the focus of attention. A
reinforcement loop is generated which is finalized to optimize the final purpose of the agent in
its own learning environment as it shown in 
Fig.~\ref{NeuronModel-Fig}.

\marginpar{\small The FOA loop and the {\em duality principle}}
The driving mechanisms of the FOA, as it emerges from the virtual masses connected 
with $\nabla b = [\nabla_{x} b, \partial_t b]$ are clearly primitive! In many species of animals - surely in 
humans - the driving process behind the FOA is definitely more complex. We use to look at ``something interesting''
and don't just focus on details. No all details are in fact interesting, and it looks like we early end up into circular 
issues when trying to establish what ``interesting'' means. When restricting our scope to 
vision we can decouple the process of eye movements with the intention which is purposely driving 
our attention to specific tasks. In order to exhibit such a skill we already need to possess 
the competence of visual perception we are interested in. Hence, it looks like we cannot neglect 
a developmental structure in the emergence of FOA and, correspondingly, in the acquisition of visual skills.
Hence, the early mechanisms of FOA could be actually based on the field generated by
sources corresponding to visual details, defined by $\nabla b$. Is there anything better?  
Can a visual agent refine the skills gained using visual details as time goes by? As will see in detail in the 
next chapter, we are primarily interested in tracking any visual structure with some meaning for the agent 
more than single pixels. The immediate consequence is that the virtual masses used so far should 
be updated to something meaningful that goes beyond the brightness of the pixels! 
A convolutional net, with its visual features, clearly offers the possible solution: instead of simply 
considering virtual masses based on $\nabla b$ we can consider virtual masses associated with the 
visual features themselves. This gives rise to a virtuous cycle: From initial acquisition of FOA based
on  $\nabla b$ we learn visual features, which in turn, become virtual masses to be used for 
a further improvement of FOA. This cycle reinforces the mutual acquisition of FOA and visual features,
which become somehow conjugate. At any stage of development, the agent drives attention on 
the basis of his current visual features, which are consistent with FOA movement. 
This interesting cyclic behavior can be interpreted in terms of the {\em duality principle}
which leads to regard motion and features as two sides of the same medal.
%
%
Interestingly, while motion invariance is a mechanism for feature learning based on the external visual source 
on the opposite, FOA is an internal mechanism for generating the agent's velocity from  
features. 

\marginpar{\small FOA is the first field which needs to be activated}
Finally, we need to say something more about the way the FOA loop is activated. 
In the simplest case, it can be stimulated simply by the brightness, along with its
spatiotemporal changes that give rise to the virtual masses which draw attention. 
Of course, the presence of richer features at the birth produce different virtual masses
capable of driving more sophisticated focussing mechanisms. 
No matter what the specific choice is, any visual agent which misses the activation of FOA
is basically a frog: The agent cannot incorporate environmental motion invariance because
of the {\em drifting problem}. The introduction of mechanisms for avoiding to get stuck
focussing in single points of the retina (e.g. see the inhibition of return) plays a fundamental
role in the introduction of saccadic movements. Overall, they play a crucial role in 
the computational mechanisms behind the development of motion inviariance. 


\chapter{Principles of motion invariance}
\label{MInv-sec}
\AtBeginShipoutNext{\AtBeginShipoutUpperLeft{%
  \put(1in + \hoffset + \oddsidemargin,-5pc){\makebox[0pt][l]{
\boxformat 
  \framebox{\vbox{
\hbox{Published by Springer, Cham---Cite this chapter:}
\smallskip
\hbox{\url{https://doi.org/10.1007/978-3-030-90987-1_3}}}
  }}}%
}}

\vspace{-4cm}
\begin{quote}
In science there is and will remain a Platonic element which could not be taken away without ruining it. 
Among the infinite diversity of singular phenomena science can only look for invariants.\\
~\\
Jacques Monod, 1971
\end{quote}

\section{Introduction}
The
strong assumption used throughout this book is that any
information-based interpretation of visual perception only relies on
motion. Unless we are only interested in pairing visual and linguistic skills
in pre-defined application domains,
{\em all you need is motion invariance}\index{motion invariance}, which we claim is at the origin of the development
of visual skills in all living animals. Basically, this claim is directly facing the 
supervised learning approach that has already 
successfully shown the capabilities of deep convolutional networks.
Their spectacular achievements clearly indicate the  fundamental role of
deep architectures in the learning of very complex functions by means of a number of
parameters that couldn't be even considered ten years ago. 
However, while the emergence of such a power of deep convolution nets 
is now widely recognized, the associated supervised learning protocol, which 
has dominated the computer vision applications in the last few years
is currently the subject of heated discussions. 
The extraordinary visual abilities of the eagle and other animals seem to suggest that the fundamental processes of vision obey visual interaction protocols that go beyond supervised learning. In this book we address primarily the question on whether we can replace the 
fundamental role of huge labelled databases with the simple ``life in a visual environment''
by means of an appropriate interpretation of the information coming from motion. 
\marginpar{\small {\em Motion is all what you need}: The two principles of motion invariance}
A major claim here is that motion is all what you need for extracting information from a visual source.
Motion is what offers us an object in all its poses. 
Classic translation, scale, and rotation invariances can clearly be gained by appropriate movements of a given object.
However, the experimentation of visual interaction due to motion goes well beyond 
and includes the object deformation, as well as its obstruction. 
Only small portions can be enough for object detection, even in 
presence of environmental noise. How can motion be exploited? 
In this chapter we establish two fundamental principles of visual perception that 
shed light on the acquisition of the identity and on the abstract notion of object as perceived by
humans. 
\marginpar{\small I and II principles of visual perception: 
object identity and affordance}
The {\em first principle of visual perception} involves consistency issues, namely  the preservation of material points  
during motion. Depending on the pose, some of those points are projected onto the retina, others are hidden.
Basically, the material points of an object are subject to  {\em motion invariance of the 
corresponding pixels on the retina}. A moving object clearly doesn't  change its identity and, therefore,
imposing motion invariance conveys crucial information on its recognition.
Interestingly, more than the recognition of an object category, this leads to the discovering of 
its identity. 
Motion information does not only confer object identity, but also its affordance, its function in real life.\index{affordance}
Affordance makes sense for a species of animal, where specific actions take place.
A chair, for example, has the affordance of seating a human being, but can have 
many other potential uses. The {\em second principle of visual perception} is about its 
affordance as transmitted by coupled objects - typically humans.  The principle states that
the affordance is invariant under the coupled object movement. Hence, a chair gains the 
seating affordance independently of the movement of the person who's seating (coupled object).
\marginpar{\small A field theory for vision}
As already seen in the previous chapter concerning FOA, there are a number of 
reasons for adopting a field theory of vision, where velocities, brightness, and 
features turn out to be associated with a given pixel. This makes it possible 
to establish motion invariance principles in all their facets, without getting
stuck in the chicken-egg  dilemma mentioned in Section~\ref{Unbilicalsec}.
This is very much in line with the very well established literature in the 
field of optical flow\index{optical flow}, where the emphasis is on making sense of the perceived
velocity on the retina. In many real-world cases, the velocity of material points
projected onto the retina differ from the perceived velocity, which is referred to
as the optical flow. Interestingly, we shall explore richer interpretations
of the optical flow by the introduction of velocities that are referred to specific
features. Overall, features and velocities are vision fields. 
This chapter is organized as follows. In the next section we introduce the idea 
of learning in spatiotemporal environments, while in Section~\ref{IdentityAffordanceSec}
we discuss the fundamental distinction between object identity and its 
affordance. In Section~\ref{FromMP3Pix-sec} we introduce the classic notion
of optical flow while discussing how a map can be established from the
visibile material points to the retina. In Sections~\ref{MPI-sec} and~\ref{CMI-sec}
we present the two fundamental principles of motion invariance connected
with object identity and affordance, respectively. 
Finally, in Section~\ref{CouplingVF-sec} we extend motion invariance as presented
by the two previous principles to the general case of features.

\section{Computational models in spatiotemporal environments}
\label{CompModSec}
%
%
For many years, scientists and engineers have proposed
solutions to extract visual features, that are mostly based on  clever heuristics
(see e.g.~\cite{sift2004}), but more recent results indicate that most remarkable
achievements have come from deep learning approaches that 
have significantly exploited the accumulation of huge visual databases labelled
by crowdsourcing~\cite{NIPS2012_4824,DBLP:journals/corr/SimonyanZ14a}.\index{deep learning}

However, in spite of successful results in specific applications, 
the massive adoption of supervised learning leads to face artificial problems that, 
from a pure computational point of view, 
are likely to be significantly more complex than natural visual tasks 
that are daily faced by animals. In humans, the emergence of cognition from visual
environments is interwound with language. 
However, when observing the spectacular skills of the eagle that catches the prey,
one promptly realizes that for an in-depth understanding
of vision, one should begin with a neat separation from language! 

\marginpar{\small What is the role of time?}
What are the major differences between human and machine learning in vision?\index{machine learning}
Can we really state that the ``gradual'' process of human learning is somewhat 
related to the ``gradual'' weight updating of  neural networks?
 A closer look at the mechanisms that drive learning in vision tasks suggests that nowadays models
of machine learning mostly disregard the fundamental role of ``time'', which 
shouldn't be confused with the iteration steps that mark the weight update.
First, notice that in nature learning to see  takes place in a context
where the classic partition into learning and test environment is arguable.
On the other hand, this can be traced back to early ideas on statistical machine learning
and pattern recognition, which are dominated by the principles of statistics.
The gathering of the training, validation, and test data leads to discover 
algorithms that are expected to find regularities in big data collections. 
This is good, yet it enables artificial processes whose underlying communication
protocol might not be adequate in many real-world problems. 
\marginpar{\small Batch mode learning: In this extreme case, the agent is expected to 
know information on its visual interaction over all its life before coming to life!}
In the extreme case of batch mode learning, the protocol assumes that the agent
possesses information on its life before coming to life. 
Apparently this doesn't surprise computer vision researchers, whereas it sounds odd for the layman,
whose viewpoint shouldn't be neglected, since we might be trapped into an artificial
world created when no alternative choice was on the horizon.
The adoption of mini-batches and even the extreme solution of on-line stochastic
gradient learning are still missing a truly incorporation of time. 
Basically, they pass through the whole training data many times, a 
process which is still far apart from natural visual processes, where  
the causal structure of time dictates the gradual exposition to video sources. 
There is a notion of life-long learning  that is not  captured in nowadays computational 
schemes, since we are ignoring the role of time which imposes causality. 

Interestingly, when we start exploring the alternatives to huge collection of labelled images,
we are immediately faced with a fundamental choice, which arises when considering their 
replacement  with video collections. What about their effectiveness?
Suppose you  want to prepare a video collection to be used for learning the market segment 
associated with cars (luxury vehicles, sport cars,  SUVs / off-road vehicles, ...).
It could be the case that a relatively small image database composed of a few thousands of labelled
examples is sufficient to learn the concept. 
On the other hand, in a related  video setting with the same space
constraint, this corresponds with a few minutes of video,
a time interval in which it is unreasonable to cover the variety of car features that can be
extracted from 10,000 images!  Basically, there will be a lot of nearly-repetitions of frames
which support scarse information with respect to the abrupt change 
from picture to picture.
This is what motivates a true paradigm shift in the formulation of learning theories for vision.
In nature, it is unlikely to expect the emergence of vision from the accumulation 
of video. 
\marginpar{\small Question no 2:  Animals gradually conquer visual skills in their own environments. Can 
computer do the same?}
Hence, couldn't machines do same? A new communication protocol can be defined
where the agent is simply expected to learn by processing 
the video as time goes by without its recording. 
Interestingly, this might open great opportunities to all research centers to compete in another
battlefield.
\begin{svgraybox}
The bottom line is that while we struggle for the acquisition of huge labeled databases, 
the actual incorporation of time might led to a paradigm shift in the 
process of feature extraction.
We promote the study of the agent life based on the ordinary notion of time, 
which emerges in all its facets.
The incorporation of motion invariance might be the key for overcoming the
artificial protocol of supervised learning. We claim that such an invariance 
is in fact the only one that we need.
\end{svgraybox}

\marginpar{\small Can animals see in a world of shuffled frames?}
This comparison between animals and computers leads to
figure out what human life could have been in a world of visual 
information with shuffled frames.
A related issue has been faced in~\cite{Wood2016} for the acquisition of 
visual skills in chicks. It is pointed out that 
``when newborn chicks  were raised with virtual objects that moved smoothly over time, 
the chicks developed accurate color recognition, shape recognition, 
and color-shape binding abilities.'' Interestingly, the authors notice that 
in contrast, ``when newborn chicks were raised with virtual objects that 
moved non-smoothly over time, the chicks' object recognition abilities were severely impaired.''
When exposed to a video
composed of independent frames  taken from a visual database, like ImageNet, that 
are presented at classic cinema frame rate
of 24 fps,
humans seem to experiment related difficulties in such a non-smooth visual presentation. 
 
Interestingly, it turns out that our  visual skills completely collapse in a task that is successfully faced
in computer vision! As a consequence, one might start formulating conjectures
on the inherent difficulty of artificial versus natural visual tasks. The remarkably 
different performance of humans and machines  has stimulated the curiosity
of many researchers in the field. 
Of course, you can start noticing that in a world of shuffled frames, a video requires order
of magnitude more information for its compressed storing than the 
corresponding temporally coherent visual stream. 
This is a serious warning that is typically neglected in computer vision, since it suggests that
any recognition process is likely to be more difficult when shuffling
frames. One needs to extract information
by only exploiting spatial regularities in the retina, while disregarding
the spatiotemporal structure that is offered by Nature. 
The removal of the thread that Nature used to sew the visual frames might prevent us
from the construction of a good theory of vision. 
Basically, we need to go beyond the current scientific peaceful interlude and  
abandon the safe model of restricting computer vision to the processing of images.
%
Working with video was discouraged at the dawn of computer vision because
of the heavy computational resources that it requires, but the time has come to
reconsider significantly this implicit choice.
Not only humans and animals cannot see in a world of shuffled frames, but we also
conjecture that they could not  learn to see in such an environment.
%
Shuffling visual frames is the implicit assumption of most of nowadays computer vision
approaches that, as stated in the previous section, corresponds with neglecting the role of 
time in the discovering of visual regularities. No matter what computational scheme 
we conceive, the presentation of frames where we have removed the temporal structure
exposes visual agents to a problem where a remarkable amount of information 
is delivered at any presentation of new examples. When going back to the previous
discussion on time, one clearly sees that the natural environmental flow must be somehow synchronized with the agent computational capability. 
The need for this synchronization is in fact one of the reasons for focussing attention
at specific positions in the retina, which confers the agent also the gradual 
capability of extracting information at pixel label. Moreover, as already pointed
out, we need to abandon the idea of recording a data base for  statistical assessment. 
There is nothing better than human evaluation in perceptual tasks, which could stimulate
new ways of measuring the scientific progress of computer vision.

The reason for formulating a theory of learning on video instead of on images
is not only rooted in the curiosity of grasping the computational mechanisms 
that take place in nature.
\marginpar{\small Question no. 3 on shuffling of video frames; does it increase the complexity of learning to see?}
A major claim in this book is that those computational mechanisms are also fundamental
in  most of computer vision tasks.

\begin{svgraybox}
It looks like that, while ignoring the crucial role of 
temporal coherence, 
the formulation of most of nowadays current computer vision tasks lead us to
tackle problems that are 
remarkably more difficult than those Nature has prepared for us!
\end{svgraybox}

\section{Object identity  and affordance}
\label{IdentityAffordanceSec}
Human visual skills are somewhat astonishing. We get in touch with the spectacular capabilities of
interpreting visual scene whenever we open the challenge of describing the underlying computational process.
From one side, humans can recognize the objects identity regardless of the different poses in which they come. 
This is quite surprising! Objects are available in different positions, at different scales and rotations, 
they can be deformable and obstructed. Only small portions can be enough for their detection, even in 
presence of environmental noise.
\begin{figure}[H]
	\centering
	\includegraphics[width=9cm]{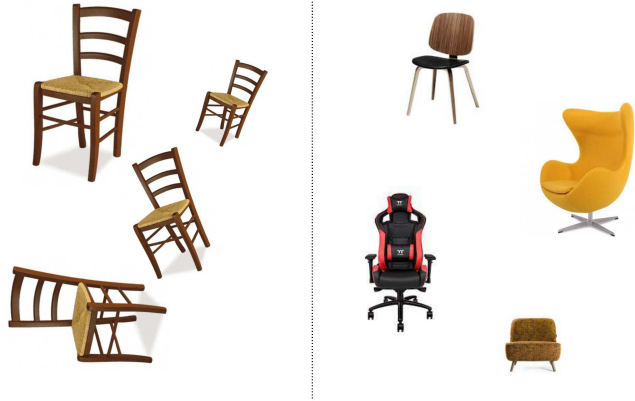}
\caption{On the left side we can see the same chair in different positions, scales, and rotation. On the right side, different types of chairs are shown as 
an example of gaining the abstract notion of chair.}
\label{Chairs-Fig}
\end{figure}
I recognize my plaid blanket that I put on the sofa from the others in my house. It can be folded more or less well, creased, stretched out, piled up somewhere, but it's still her. 
If I squeeze a  Perrier plastic bottle with my hands and close it with its cap so that 
it maintains the flattened shape, I continue to recognize it without difficulty.
If I accidentally break my favorite cup, I can still recognize it from many of its pieces, maybe not all of them; a bit like when I see it despite being obstructed by other objects.
Children recognize their Teddy Bear from different angles. As
they bite it, not only does the Bear deform as a result, but they experience a very peculiar sight of it. Yet,
they perceive it consistently. 
\marginpar{\small Object identity: does it emerge from the massive number of poses provided by its motion?}
However, you can't really hope to recognize your bear by a front picture, 
if the only thing that distinguishes it from another is that 
it is slightly torn behind the head. It is that torn part that gives it its identity. 
It looks like the object motion is always behind the different poses of the object. 
There is  nothing else than motion which concurs to present the different ways we perceive an object, 
with its own identity. Hence, one might wonder whether the object identity
 comes from the massive number of poses provided by its motion.
The corresponding learning process should develop 
such an invariance by neural-based maps.
Deep learning\index{deep learning} has already been proven to be very effective to replicate those
visual capabilities, provided that we rely on tons of supervisions. Interestingly, it could be case that
those explicit linguistic-based supervisions can simply be offered by motion. 
When considering the virtually infinite availability of visual information offered by motion 
the supervised learning approach may have hard time to 
reach the human capabilities to define the identity of objects.

%
%
Any learning process that relies on the motion of the given object can only aspire to 
discover its identity.\index{affordance}
 The motion invariance process is in fact centered around
the object itself and, as such, it does reveal  its own features in all possible expositions
that are gained during motion. However, as shown in Fig.~\ref{Chairs-Fig}, in addition to the capability of 
identifying objects (left-side), humans clearly gain abstraction (right-side).
\marginpar{\small Object affordance}
Humans, and likely most animals, also conquer a truly different understanding of visual scene that
goes beyond the conceptualization with single object identities.
In early Sixties, James J. Gibson coined the {\em affordance} in~\cite{Gibson1966}, even though 
a more refined analysis came later in~\cite{plasticq}. In his own words:
\begin{quote}{\em
The affordances of the environment are what it {\em offers} the animal, what it {\em provides} 
or furnishes, either for good or ill. The verb to {\em afford} is found in the dictionary, the noun affordance is not. I have made it up. I mean by it something that refers to both the environment and the animal in a way that no existing term does. It implies the complementarity of the animal and the environment.
}\end{quote}
Affordance makes sense for any species of animal, where specific actions take place.
A chair, for example, has the affordance of seating a human being, but can have 
many other potential uses. We can climb on it to reach a certain object or we 
can use it to defend ourselves from someone armed with a knife.
It was also early clear that no all objects are significantly characterized
by  their affordance. In~\cite{plasticq}, the distinction between {\em attached and detached objects}
clearly indicates that  some can neither be moved nor easily broken. While our home can be given
an affordance, it is attached to the earth and cannot be grasped. Interestingly, while a small cube is 
graspable, huge cubes are not. However, a huge cube can still be graspable by an appropriate ``handle.'' 
Basically,  one can think of object perception in terms of its properties and qualities. 
However, this is strictly connected to the object identity, whereas most common and useful 
ways of  gaining the notion of objects likely comes from their affordances. 
The object affordance is what humans -- and animals -- normally pay attention to. 
As  infants begin their environmental experiences they likely look at object meaning more
then at the surface, at the color or at its form. This interpretation is kept during our ordinary 
environmental interactions. 
\marginpar{\small Do we really need to classify objects?}
The notion of object  affordance  somewhat challenges the apparently indisputable philosophical 
muddle of assuming fixed classes of objects. This is a slippery issue! To what extent can we 
regard object recognition as a classification problem? Clearly, 
there is no need to classify objects in order to understand what they are for. 
Hence, we could consider the advantages of moving from nowadays dominating computer vision  
approach to a truly ecological viewpoint that suggests experimenting niches based set of well-defined, typically
limited object affordances.

\section{From material points to pixels}\label{FromMP3Pix-sec}
First, we begin discussing the classic notion of optical flow\index{optical flow}, which is in fact 
of central importance in computer vision. \marginpar{\small What is the optical flow?}It's a slippery
concept and 
we shall see that its in-depth analysis is very important for 
interesting developments on more general tracking issues.
%
%
\marginpar{\small Foveated animals always experiment motion!}
In computer vision one typically considers the case of moving objects
acquired by a fixed camera as well as the case of egocentric vision 
that entails analyzing images and videos captured by a wearable camera. 
However, because of eye movements,
foveated animals experiment the presence of optical flow even 
in the case in which there is neither the movement of the animal nor of the visual
environment. Hence, one can always consider to deal with the general case
in which the visual information is acquired in the reference of the retina, where we
are always in front of optical flow.

%
%
The object movements in the real world can be naturally framed in a three dimensional
space  (3-D) and, as such, the velocity of any single material point turns out to be a 
3-D vector.  Animals and computers can only perceive the movement
as a 2-D projection of material points onto the  correspondent pixels
of the retina. Clearly, finding such a 
correspondence is not an easy problem. First, some material points are not visible.
When they are visible, the map that transforms material points
to pixels is not necessarily unique, especially  for visually uniform  regions.
A trivial case in which ill-position arises is simply 
when the camera frames an entire white picture. There is no way to tell if
something is moving; it could either be a wall or a big moving white object.
While one can always exclude this case, the underlying ambiguity still
remain in ordinary visual environments.

\marginpar{\small The principle of brightness invariance}
An enlightening approach to the problem of estimating the optical flow 
of the pixels corresponding to material points of a moving object
is that of imposing the principle of {\em brightness invariance}. 
Let $b(x,t)$ be the brightness at $(x,t) \in \Omega \times (0,T)=:\Gamma$. 
In the big picture of this book, $T$ denotes the duration of video
segments delimited by saccadic movements.
The principle consists of stating
that the moving point associated with the trajectory $x(t)$ doesn't change its brightness, 
that is $b(x(t),t) = c$, being $c$ constant over time. Hence, this can be
restated by
\begin{equation}
\frac{d b(x(t),t)}{dt} = \sgrad b \cdot v +  b_t = 0,\quad \forall t\in(0,T).
\label{OFconstraint}
\end{equation}
This is a condition on the velocity $v=\dot{x}$ that  can admit infinite solutions since it is a scalar equation with two unknowns. 
%
An enlightening approach to the problem of estimating the optical flow 
of the pixels corresponding to material points of a moving object was
given by Horn and Shunck in a seminal paper published at the beginning of the
eighties~\cite{HornAI1981}. 
They adopted a regularization principle which turns out to have 
a global effect on the guess of the optical flow. 
Basically, they proposed to determine the optical flow $v$
as the minimization of
\begin{equation}
	E(v_{1},v_{2})
	=\int_{\Omega}\big(\sgrad v_1\big)^2+ \big(\sgrad v_2\big)^2,
\label{OpticalFlowReg}
\end{equation}
subject to the constraint~(\ref{OFconstraint}).
One can promptly realize that while this is a well-posed  formulation, it doesn't
take into account the change of brightness of moving material points 
when lighting conditions change significantly. 
If the object moves from a luminous to a dark area, then the condition~(\ref{OFconstraint})
doesn't really reflect reality and the corresponding estimation of the velocity 
becomes quite poor. 

\marginpar{\small Color tracking}
Notice that the brightness might not necessarily be the ideal signal to track. 
Since the brightness can be expressed in terms of the red $R$, green $G$,
and $B$ components as
$b=R+G+B$, one could think of tracking single color components
of the video signal by using the same invariance principle stated by
Eq.~(\ref{OFconstraint}). It could in fact be the case that one or more of the
components $R$, $G$, $B$ are more invariant during the motion of the corresponding
material point. In that case, in general, each color can be associated with 
a corresponding velocity $v_R,v_G,v_B$ that might be different. 
In so doing, instead of tracking the brightness, one can track the single colors.
On the opposite, The simultaneous track of  all the colors, with the same velocity
$v=v_R=v_G=v_B$ yields
\begin{align}
	v\cdot \sgrad
	\begin{pmatrix}
	 R\\  G\\ B
	 \end{pmatrix}
	 + \frac{\partial}{\partial t}
	 \begin{pmatrix}
	 R \\ G \\ B
	 \end{pmatrix}
	 = 0,
\label{ColorTrackingEq}
\end{align}
where $v\cdot\sgrad (R,G,B)
:=(v\cdot\sgrad R, v\cdot\sgrad G,v\cdot\sgrad B)
$ and $\sgrad (R,G,B)$
is generally nearly-singular. However, it's worth mentioning that
the simultaneous tracking of different channels might contribute
to a better positioning of the problem. One can think of the color component
as features that, unlike classic convolutional spatial features
are temporal features. We can in fact regard the color components
as the outputs of three convolutional filters in the temporal domain.

Interestingly, as it will discussed in the following, humans and likely many 
species of animals, very well deal with this problem when they are involved in 
tracking tasks. An in-depth analysis on this issue leads to 
a novel of velocity which is interwound with the notion of visual feature.
Before facing this major topics, it's convenient to better explore
the meaning of the optical flow and grasp the limitations behind the principle of brightness invariance.\\
~\\
\noindent {\sc Problems with the Principle of brightness invariance}\\
When the the brightness of the background changes an illusion phenomenon can arise
that do compromise the correct estimation of the actual kinematic velocity. 
Let us consider an artificial example to illustrate this issue. In particular, suppose that 
\begin{equation}
	b(x,t)= t x\cdot\hat e_1
\label{LinearBrGrowth}
\end{equation} 
where $\hat e_1=(1,0)’$, then a solution of the
brightness invariance condition is $v(x,t)=-(x\cdot\hat e_1/t,0)'$. Notice that here the choice
$v\cdot\hat e_2=0$ has been done quite arbitrarily since any value of the projection of the velocity
along $\hat e_2$ would satisfy the brightness invariance condition. Anyway, this example shows how a
fixed image with changing lighting over time can generate an illusion of motion.

\section{The principle of material point invariance}
\label{MPI-sec}
A cat that is chasing a mouse continues its chase effectively even if the mouse 
passes through areas of spatially or temporally variable brightness 
like the one considered in Eq.~(\ref{LinearBrGrowth}).
Likewise, a flickering light doesn't 
significantly help the mouse escape. How can this be possible?
The cat is not only tracking single material points, but he has clearly got
the capability of perceiving the overall picture of the mouse, as well as many of his
distinguishing features. 
Hence, tracking in nature involves directly objects and their features more than single material points. 

\marginpar{\small The cat chasing the mouse:  What is the velocity of the mouse estimated by the cat?}
As  the cat chases the mouse, it looks like he is given the task to estimate the mouse velocity.
This does require a deeper analysis.
Are we talking of the velocity of the barycenter of the mouse? 
How can the cat determine the barycenter? 
Again, the pixel-wise computation that has been invoked in Section~\ref{PixelwiseCompS}
comes to help. We can think of convolutional-like features on $(x,t)$ that somehow
characterize the way the mouse is visually interpreted by the cat, who's tracking those features. 
We can associate different velocities at $(x,t)$ depending on 
the feature we are referring to. As the feature conquers an interpretable semantics
the velocities associated with  each pixel offer a novel scenario for the cat, who
doesn't restrict to the flow of the velocity associated with the material points. 
\marginpar{\small The principle of material point consistency - {\tt MPI}}
In Sec.~\ref{IdentityAffordanceSec} we introduced the fundamental distinction
between object identity and affordance. We begin  addressing all issues
involving object identity, since it covers motion invariance of the object features.
The process of recognizing the object identity is based on the following
{\em Material Point Invariance principle} ({\tt MPI}) (I Principle), which blesses the pairing 
of any feature of a given object, including the brightness,  with
its own velocity.

\subparagraph{First Principle of Perceptual Vision:  Material point Invariance}
Let us consider the motion of an object $\mathscr{O}$ over $\Gamma$.
As already mentioned, we are considering burst of movements that can be
delimited by saccadic jumps of duration $T$ at most.  
The reason for this limited duration has already been discussed when 
we introduced the problem of the concept drift.
\marginpar{\small Invariance takes place in portions of video delimited by saccadic movements}
Now, let us denote by $P(t) \in \mathscr{O}$ one of its material points of
object $\mathscr{O}$ at $t <T$. Let  $x(t) = \pi(P(t))$ be the pixel 
which comes from  the projection of $P(t)$ onto the retina $\Omega$, 
where $\pi(\cdot)$ denotes the projection. 
Regardless of the knowledge of $\pi(\cdot)$, we can think of the 
invariance property which is induced by the trajectory $x(t)$. The brightness
is approximatively constant on such a trajectory, that is  $b(x(t),t)=c$.
Now, let $\varphi: \Gamma \to \bbR$ be any of the visual features of $\mathscr{O}$. 
The underlying assumption is that the computation of $\varphi$ on any $(x,t)$ is made possible
mostly by the knowledge of $b(\cdot,t)$, that is by the frame at time $t$.
In a sense, $\varphi(x,t)$ is supposed to be a feature that is computable by 
a ``nearly-forward process'' by involving a ``quick'' dynamical structure. 
A cognitive view of this assumption corresponds with the capability of 
describing ``features'' in a video nearly-independently of time,
on single frames, when focusing attention to a certain pixel.
In any case, single frames do support visual information without needing
to involve sequential processing. Apparently, this seems to be an unrealistic 
limitation but, on the opposite, it turns out to be one of the most important 
requirements of visual cognition.  However, as it will become clear in the 
following, the need of developing tracking feature capabilities gives rise
to an overall dynamical computational process that will be described in
Ch.~\ref{driving-principles}.

\begin{figure}
\centerline{
\includegraphics[width=12cm,height=4cm]{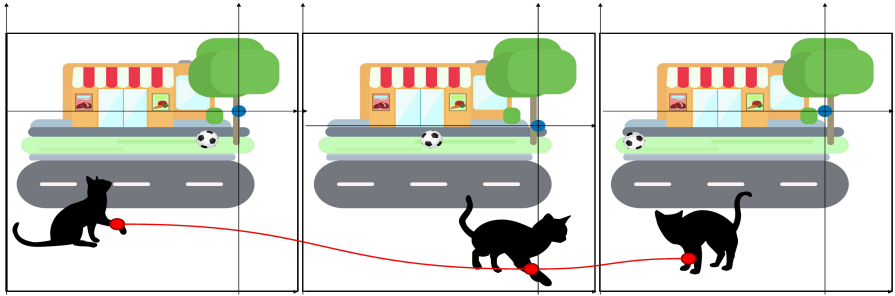} 
}
\caption{\small The Principe of Material Point Invariance: When considering any
pixel, the correspondent visual feature doesn't change during its motion, which will 
be considered either with respect to the computer retina or to the focus of attention.}
\label{VFG-fig}
\end{figure} 
 
We can parallel the notion of optical flow\index{optical flow} for the brightness by introducing the 
corresponding notion for $\varphi$. Hence,
the following consistency condition holds true:
\begin{equation}
\varphi(x_{\varphi}(t),t) = \varphi(x_{\varphi}(0),0) 
	= c_{\varphi},\quad \forall t\in[0,T]
\label{MPI-principle}
\end{equation}
where $x_{\varphi}(t)$ denotes a trajectory of the associated feature  $\varphi$
and $c_{\varphi} \in \bbR$; if $x_\varphi(t)=x$ we let $v_\varphi(x,t)=\dot x_\varphi(t)$. See Fig.~\ref{VFG-fig} for a sketch of the idea behind the consistency condition.
It is important to notice that this principle relies on the underlying assumption of
the joint existence of $\varphi$ and on a of the 
corresponding trajectory $x_{\varphi}$
that satisfies the condition stated by Eq.~(\ref{MPI-principle}), which
somewhat corresponds with thinking of $\varphi$ as a signal
which is propagating with velocity 
$v_{\varphi}$. In general we can compute
$v_{\varphi}$ by a function which depends on 
$\tgrad \varphi(\cdot,t)= (\sgrad \varphi(\cdot ,t), \varphi_t(\cdot,t))'$.
Like for the brightness, in general,  the invariance condition generates an
ill-posed problem. In particular,  when the moving object has a uniform color,
we noticed that brightness invariance holds for virtually an  infinite number of trajectories. 
Likewise, any of the features $\varphi$ is expected to be spatially smooth - nearly
constant in small portions of the retina. This restores the ill-position of the classic problem
of determining the optical flow that has been addressed in the previous section. 
\marginpar{\small Double face of ill-position in feature extraction}
This time ill-position has a double face. Just like in the classic case of estimating the optical flow,
$v_{\varphi}$ is not uniquely defined. On top of that, now the corresponding
feature $\varphi$ is not uniquely defined, too.
Basically, the {\tt MPI} only states that material point invariance on the feature 
trajectories $x_{\varphi}$ can be drawn jointly with the associated feature.
Since the I Principle only establishes consistency on trajectories, we need
to involve additional information for making the learning process well-posed.
A regularization process similar to the one invoked for the optical flow $v$---see 
Eq.~(\ref{OpticalFlowReg})---can also be imposed for $v_{\varphi}$.
However, unlike for the brightness, this time $\varphi$ is not given. On the opposite, 
it is in fact characterizing the object, and it represents a ``moving entity'' of which
we also want to measure the velocity.  
As discussed in the following, the functional structure of both $\varphi$ and $v_{\varphi}$ plays in fact a crucial role in the learning process and, correspondently, on its 
development. 
\marginpar{\small Neural-based structure for $\varphi$ and $v_{\varphi}$}
The structure of  $\varphi$ is affecting the 
associated velocity $v_{\varphi}$ and vice versa.
In Section~\ref{VRes-sec}, we shall see that, 
as one begin thinking of the learning process behind  
$v_{\varphi}$, it is natural to pair such a process with the feature discovery,
which leads to assume also a neural-based support for $v_{\varphi}$.
The learning agent is supposed to jointly acquire both the features and their corresponding velocities.

\marginpar{\small What if the eyes and/or the body are moving? What if also the objects are moving?
Take the FOA as a reference for movements!}
The formulation of the {\tt MPI} principle raises the question on the 
precise meaning of the trajectory $x_{\varphi}(t)$. So far, we have considered the 
case in which an object is moving in front of a fixed retina.
What if the object is fixed and
the eyes and/or the body are moving? Moreover, it could also be the case that we are 
tracking a moving object while moving ourselves. This holds for 
machines as well. Since foveated animals move their eyes continually, their perception
can naturally be understood by  taking the FOA as a reference. As such, if we move, a stationary object turns out
 to be in motion along with the background. 
 The bottom line is that {\em foveated animals are always in front of motion}, 
 which conveys a continuous information flow.
 Of course, as pointed out later on, the motion invariance 
 principle 
 is independent of the frame of reference.
The micro-saccadic movements enforce invariance on visually uniform 
patterns while the object tracking provides a richer information on the object,  since 
it is acquired in different areas of the background.
Apparently the body movement makes the problem considerably more involved,
since the optical flow which arises doesn't distinguish the objects anymore.
If we move uniformly on the right without moving our eyes, this clearly corresponds with a uniform optical flow
on the retina with constant velocity on all the pixels in the opposite direction.
It's interesting to see what happens if we propagate the feature consistency
stated by {\tt MPI} over the entire retina with respect to that constant velocity. 
Of course, in general, no feature meets such a criterion apart from the approximate satisfaction 
of the brightness stated by the invariance condition of Eq.~(\ref{OFconstraint}).
Basically, it looks like that if we focus on the certain pixel with velocity $v$, the
corresponding uniform movement doesn't support any 
information for discovering visual features! 
However, as will be better seen in the following, the pairing of the features with their
associated optical flow leads to establish an effective learning process also in this case, 
since the velocity fields of the features are not just reflecting the velocity associated with brightness invariance.

\index{motion invariance}\index{feature conjugation}
\marginpar{\small Conjugated features: We think of $(\varphi,v_{\varphi})$ as 
an indissoluble pair which must be jointly learned. We can detect the feature $\varphi$ since ``we can see'' it
moving at velocity $v_{\varphi}$.}
Equation~(\ref{MPI-principle}) considers a single trajectory defined on the whole temporal horizon $[0,T]$, over which a very strict condition is imposed. In practice, we are interested in considering multiple trajectories, one of each point of the retina, that are locally defined in time. This leads to differential conditions that describe the local relationships over time of the way $\varphi$ develops in the retina coordinates.
Formally, for any pair $(x,t) \in \Gamma$, let us overload the notation $x_{\varphi}(t)$ to represent the trajectory for which 
$x_{\varphi}(t) = x$ and so that $\dot x_\varphi(t)=v_\varphi(x,t)$. 
Given the optical flow $v_{\varphi}$, we say that $\varphi$ is
 a {\em conjugate feature} with respect to  $v_{\varphi}$ provided that
 $\forall (x,t) \in\Gamma$,
 we have
\begin{equation}
\left( \varphi \Join v_{\varphi} \right) (x,t) := 
	\frac{d \varphi(x_{\varphi}(t),t)}{dt} =
	\sgrad \varphi(x,t)  \cdot v_{\varphi}(x,t) +\varphi_t(x,t) =0,
\label{eq:T-inv}
\end{equation}
where $\Join\colon\bbR^{\Gamma} \times\bigl(\bbR^{\Gamma}\bigr)^2\to \bbR^{\Gamma}$
performs the mapping\footnote{
Here with the symbol $\bbR^X$ we denote the set of all maps $f\colon X\to\bbR$.} $(\varphi,v_\varphi)\mapsto \varphi\Join v_\varphi$.
Notice that if there is no optical flow in a given pixel $\bar{x}$, that is
if $v_{\varphi}(\bar{x},t)=0$ for all $t\in[0,T]$, then 
$\varphi_t(\bar x,t) = 0$. This means that the absence of the 
optical flow in $\bar{x}$ results into $\varphi(\bar{x},t)=c_{\varphi}$
for all $t\in[0,T]$, which is the obvious consistency condition that
one expects in this case. 
Likewise, a constant field $\varphi(x,t)$  in 
$C \subset \Gamma$ results into the conjunction
$\varphi \Join v_{\varphi}=0$ on $C$, independently of $v_{\varphi}$.
Of course, since  $b \Join v=0$  we can legitimately state
that $b$ is in fact a conjugate feature---the simplest one that is associated with the
brightness that is based on a single pixel.
As for the optical flow associated with the brightness, also in this case,
 the regularization term~(\ref{OpticalFlowReg}), 
where $v$ is replaced with $v_{\varphi}$, contributes to a well-posed formulation.
Basically $(\varphi,v_{\varphi})$ is an indissoluble pair that plays a fundamental
role in the learning of the visual features that characterizes the object. 
Sometimes in what follows we will drop the subscript $\varphi$ of
$v_\varphi$ and $x_\varphi$ when the notation is clear from the context.

\marginpar{\small Frame of reference} 
The I Principle has been stated by assuming the retina of the computer as the
frame of reference. Foveated animals obviously experiment the MPI invariance by 
taking the fovea of their eyes as frame of reference. We can follow the 
same choice for machines. The trajectory $a(t)$ of the FOA makes it possible 
to set a frame of reference with axes parallel to those of the computer retina
and center corresponding with the current position $a(t)$ of the FOA.
In this new frame of reference we always experiment motion, which suggests
that there is always information coming from the invariance of the I Principle,
including the case of still images and stationary visual agent.

\subparagraph{Feature Grouping}
As already noticed,
when we consider color images, the brightness invariance can be considered for
the separated components {\tt R,G,B}. Interestingly, for a material point of a certain color,
given by a mixture of the three components, we can establish the same brightness invariance
principle, since those components can move with the same velocity. Said in other words,
different features can share the same velocity.
We can generalize this case by considering a group of features which share the
same velocity. Hence, let 
$\varphi_{i} \Join v=0$ $\forall i=1,\ldots,m$ be, that  is all features are
conjugated with the same optical flow $v$, then we can promptly see that 
any feature $\varphi$ of
\marginpar{\small Functional space $\spn(\varphi_1,\dots,\varphi_m)$ represents 
a {\em feature group}, that is 
uniformly conjugated with the same optical flow} $\spn(\varphi_1,\dots,\varphi_m)$
is still conjugated with $v$. We can think of $\spn(\varphi_1,\dots,\varphi_m)$ as a  functional space 
conjugated with $v$.
As already pointed out, the brightness is a quasi-invariant feature,
but as we increase the degree of abstraction, we can aspire to increase the
approximation of the conjugation of feature $\varphi$ with its correspondent 
optical flow $v_{\varphi}$. Basically, large objects are perceived with a high degree
of abstraction and are easier to track than their small parts.
As already stated, the degree of ill-posedness of  
the determination of the classic optical flow is pretty high. A uniformly colored 
red rectangle, which is translating in front of us, makes it very difficult to guess the velocity of 
the internal pixels of the rectangle. We cannot really track them!
We can only make a guess that can be
driven by the smoothness of the optical flow~(\ref{OpticalFlowReg}). However, 
we can easily follow the whole rectangle, just as  the cat chases the mouse, that is 
regarded as a whole object. This way of conjugating features with the corresponding velocity
seems to offer a formal support for the intuitive statement that we are tracking a moving object.
A fundamental comment concerning the tracking 
of a feature vector $\phi=(\varphi_1,\dots,\varphi_m)'$ 
which is very much related to the discussion on
color tracking that is stated by Eq.~(\ref{ColorTrackingEq}). In this
case the invariance on $\phi$ yields
\marginpar{\small The regularization effect of {\em feature grouping}}
\begin{align}
\sgrad\phi \cdot v+\phi_t =0,
\label{varphi-tracking-eq}
\end{align}
where $\sgrad\phi\in\bbR^{m\times 2}$ is defined as $(\sgrad\phi)_{ij}={\phi_i}_{x^j}\equiv(\sgrad\varphi_i)_j$.
Notice that, if we consider the case in which the only scalar feature we are tracking is the brightness, then Eq.~\eqref{varphi-tracking-eq} boils down to a single equation with two unknowns (the velocity components). Differently, in the case of the feature group $\phi$, we have $m$ equations and still two unknowns.
The dimension $m$ of matrix $\sgrad\phi$ 
can enforce the increment of its rank, which leads to a better positioning of the
problem of estimating the optical flow $v$.
Because of the two-dimensional structure of the retina, which leads to 
$v \in \bbR^2$ and since 
$\sgrad\phi \in \bbR^{m\times2}$, with
$m \geq 2$ it turns out that any feature grouping regularizes
the velocity discovery. In order to understand the effect of feature 
grouping we can in fact simply notice that a random choice of the features 
yields $\rnk\sgrad\phi = 2$.
As a consequence, linear equation~(\ref{varphi-tracking-eq})
admits a unique solution in $v$. Moreover, from Rouch\'e-Capelli theorem
we can also promptly see that in order to achieve the tracking of the 
feature group, the features need to satisfy 
\[
	\rnk\sgrad\phi=\rnk (\sgrad\phi\mid -\phi_t).
\]
However, this regularization effect of feature grouping doesn't prevent 
ill-positioning, since $\varphi$ is far from being a random map. 
One the opposite, it is supposed to extract a uniform value in portions of the
retina that are characterized by the same feature. Hence, 
$\rnk\sgrad\phi=1$ 
is still possible whenever the features of the group are somewhat dependent.

\marginpar{\small Inheritance of conjugation}\index{feature conjugation}
Feature groups, that are characterized by their common velocity, can give rise
to more structure features belonging to the same group. This can promptly be
understood when we go beyond linear spaces and consider 
for a set of indices $\mathcal{F}$
\begin{equation}
\left\{\begin{aligned}
    &\alpha =\sum_{j \in {\cal F}} w_{j} \varphi_{j}\\
    &\varphi = \sigma(\alpha).
\end{aligned}\right.
\label{NeuronLikeFea}
\end{equation}
Like for linear spaces, if $\varphi \Join v = 0$ then
\[
    \varphi \Join v = 
    \sigma^{\prime}(\alpha) \left(\sgrad \alpha \cdot v + \partial_{t} \alpha
    \right)
    =\sigma^{\prime}(\alpha)
    \sum_{j \in {\cal F}}\bigl( w_{j} \sgrad \varphi_{j} \cdot v +
    w_{j} \partial_{t}\varphi_{j}\bigr).
\]
From the conjugation equation $\varphi \Join v =0$ we get 
\[
    \varphi \Join v  = \sigma^{\prime}(\alpha)
    \sum_{j \in {\cal F}} w_{j} \left( \nabla \varphi_{j} \cdot v +
    \partial_{t}\varphi_{j}\right)
    = 0.
\]
Hence, we conclude that if $\forall j \in {\cal F}$ we have $\varphi_{j} \Join v=0$ then
also the feature $\varphi$ defined by Eq.~\ref{NeuronLikeFea} is conjugated with $v$, that is
$\varphi \Join v = 0$. We can promptly see that vice versa doesn't hold true. 
Basically the inheritance of conjugation with $v$ holds in the direction towards more
abstract features. Of course, the feedforward-line recursive application of the derivation stated by Eq.~(\ref{NeuronLikeFea}) 
yields a feature $\varphi$ that is still conjugated with $v$. In this case we use the notation
$\vdash \varphi$ to indicate such a data flow computation.

\subparagraph{Computational Models for $(\varphi, v_{\varphi})$}
The notion of conjugate features does require the involvement of
the specific computational structure of $\varphi$. 
In general, we assume that  $\varphi$ and $v_{\varphi}$ turn out to be expressed by
\begin{align}
\begin{split}
	&\frac{\partial \varphi}{\partial t}(x,t) + c\varphi(x,t)= c \alpha_{\varphi}\big(b(\cdot,t),a(t),t\big)(x); \\
	&\frac{\partial v_{\varphi}}{\partial t}(x,t) + c v_{\varphi}(x,t)=  
	c \alpha_{v}\big(\tgrad \varphi(\cdot,t),a(t),t\big)(x),
\end{split}
\label{aggreg-eq}
\end{align}
where $\alpha_\varphi\colon \bbR^\Omega\times\Omega\times[0,T]\to\bbR^\Omega$,
$\alpha_v\colon\bbR^\Omega\times\Omega\times[0,T]\to(\bbR^\Omega)^2$, 
$t\mapsto a(t)$ is the trajectory of the focus of attention and $c>0$\index{focus of attention}.
Here, $c$ is supposed to drive very quick
dynamics, so as if the input  $b(\cdot,t)$ is kept constant
we end up quickly into the associated stationary condition
$\varphi(x,t)= \alpha_{\varphi}\big(b(\cdot,t),t\big)(x)$.
Basically, this corresponds with performing a ``nearly-forward computation''.
In human vision the system dynamics is expected to be compatible with
classic  frame per second frequency.
The same computational structure is assumed
for the conjugated velocity ${v}_{\varphi}(x,t)$.
The explicit dependence on $t$ in $\alpha_{\varphi}$ and $\alpha_{v}$
is due to the underlying assumptions that those functions undergo a 
learning process during the agent's life.
In Eq.~\eqref{aggreg-eq} it is assumed that the computation 
of the features  at a
point $x_0\in\Omega$ requires the knowledge of the brightness
on the whole retina. However, this assumption can be 
relaxed by requiring that we pass to the function 
$\alpha_\varphi$ the restriction
$b(\cdot,t)|_{\Omega_\varphi(x_0)}$
of the brightness to 
a subset $\Omega_\varphi(x_0)\subset\Omega$.
In doing so the computation of $\varphi(x_0,t)$ makes it possible
to take decisions on the actual portion of the retina $\Omega_\varphi(x_0)$ that can 
be virtually considered 
when computing the feature on pixel $x_0$. This is somewhat related to the 
convolutional neural networks computational scheme and 
it will be specifically 
described in Ch.~\ref{DNN-sec}. 

The computational structure of $v_{\varphi}$ is suggested by its conjugation with $\varphi$,
where we see the dependence on $\tgrad \varphi$.
This is related to the classic idea early illustrated by Lukas and Kanade
in~\cite{lucas1981iterative} for the optical flow, where its determination is based on 
stating the conjugation condition on a receptive field $\Omega_b(x)$ centered on 
the pixel $x$ where we focus attention. 

Of course, the conjugation of $\varphi$ and $v_{\varphi}$ does involve
the choice of both $\alpha_{\varphi}$ and $\alpha_{v}$. From Eqs.~(\ref{eq:T-inv}) 
and~(\ref{aggreg-eq}), we get the following consistency condition on $\varphi_t$ 
\begin{equation}
	c^{-1}\sgrad \varphi(x,t)  \cdot v_{\varphi}(x,t) =\varphi(x,t)  - \alpha_{\varphi}\big(b(\cdot,t),a(t),t\big)(x).
\label{partial-varphi-cons}
\end{equation}
As we can see, for large values of $c$,
this corresponds to a  quasi-forward computation. 
\\
~\\
{\em Regularization for Determining $(\varphi, v_{\varphi})$}\\
We have already discussed the ill-posed definition of features conjugated 
with their corresponding optical flow. Interestingly, we have shown that a feature group 
$\phi = (\varphi_i,\ldots,\varphi_m)'$, 
which shares the velocity inside the group, exhibit an inherent regularization 
that, however, doesn't prevent ill-positioning, especially when one
is interested in developing abstract features that are likely constant over
large regions in the retina.
Let us assume that we are given $n$ feature groups $\phi^i$, $i=1,\dots n$, each one 
composed of $m_i$ single features ($m_i$-dimensional feature vector)
$\phi^i\in\bbR^{m_i}$. Furthermore, let us  assume that each group $\phi^i$ shares the same velocity $v^i$ with all the features of the group. Let us also denote with 
$\allphi=(\phi^1,\dots, \phi^n)$ and with 
$\allv=(v^1,\dots,v^n)$.
\marginpar{\small Default ``null'' and spatial smoothness}
We can impose the generalization of the smoothness term used 
for the classic optical flow to the velocities of the features\index{optical flow}
by keeping the following term small:
\begin{equation}
	E= \frac{1}{2} \sum_{i=1}^{n} \int_{\Gamma} \bigl(\Vert\sgrad v^i(x,t)\Vert^2
	+\lambda_{\varphi} |\phi^i(x,t)|^2+ \lambda_{\sgrad} \Vert\sgrad \phi^i(x,t)\Vert^2\bigr)\, d\mu(x,t).
\label{GenOpticalFlow}
\end{equation}

Here,  $\Vert A\Vert^2=\sum_{i,j} A_{ij}^2$ and  $\lambda_{\varphi}$, $\lambda_{\nabla}$
are positive constants that
express the relative weight of the regularization terms and the measure $\mu$ is an appropriately 
weighted Lebesgue measure on $\bbR^2_x\times\bbR_t$: $d\mu(x,t)=\exp(-\theta t)g(x-a(t))\, dxdt$.
Here  we have $\theta>0$, so as $\exp(-\theta t)$ produces a decay as time goes by.
Function $g$ is non-negative and it has a radial structure and decreases 
as we get far away from the origin.
It takes on the maximum for $x=a(t)$, where
$t\mapsto a(t)$ denotes the coordinates of the focus of attention. 
Notice that $E$ is a functional of the pair $(\varphi,v)$, that is, once it is given,
we can compute $E(\allphi,\allv)$.
When comparing this index with the one used to 
regularize the classic optical flow (see Eq.~(\ref{OpticalFlowReg}))
we can see a remarkable difference: $E$ is also a function of time! Why is that necessary?
While the brightness is given, the features are learned as time goes by, which
is just another facet of the feature-velocity conjugation. 
It's worth mentioning that $E$ does only involves spatial smoothness
whereas it doesn't contain any time regularization term.
As it will become more clear in Chapter~\ref{driving-principles} 
such a regularization is in fact introduced into the weight of 
the neural networks that are used to implement Eq.~(\ref{aggreg-eq}).
We can also introduce a form of temporal regularization by 
upper-bounding $|v^{i}|$, since it directly affects the conjugated features
thus limiting $\partial_{t} \phi^{i}$.
There's also another difference with respect to  the 
classic optical flow (see Eq.~(\ref{OpticalFlowReg})): There's also a penalizing
term $(1/2) |\phi^i|^{2}$ which favors the development of  $\phi^i=0$.
Of course, there's no such requirement in classic optical flow, since $b$
is given. On the opposite, the discovery of visual features is expected to be 
driven by motion information, but their ``default value'' is expected to be null. 
We can promptly see that the introduction of the regularization term~(\ref{GenOpticalFlow})
doesn't suffice. 
The conjugation $\Join$ is in fact satisfied also by 
the trivial constant solution $\varphi=c_{\varphi}$.

\marginpar{\small What happens in the next frames?}
Important additional information 
comes from the need of exhibiting  the human visual skill of
reconstructing pictures from our symbolic representation. 
At a certain level of abstraction,
the features that  are gained by motion invariance,
possess a certain degree of semantics that is needed to interprete the
scene. However, visual agents are also expected to deal with 
actions and react accordingly. 
As such a uniform cognitive task that visual agents are  expected to 
carry out is that predicting what will happen next, which is translated 
into the capability of guessing the next incoming few frames in the scene. 
We can think of  a predictive computational scheme based 
on the  $\phi^i$ codes which works as follows:
\begin{align}
\label{GlobalResp-y}
	&\frac{\partial y}{\partial t}(x,t) + c y(x,t)= c \alpha_y\big(\allphi(\cdot,t),t\big)(x).
\end{align}
\marginpar{\small The prediction tasks consist of discovering an output $y$ such that
$y(\cdot,t) \simeq  b_t(\cdot,t)$.}where $\alpha_y\colon\bigl(\bbR^\Omega\bigr)^n\times[0,T]\to\bbR^\Omega$.
%
In particular, the prediction $y$ needs to satisfy the condition established 
by the index
\begin{equation}
	 R= \frac{1}{2} \int_\Gamma 
	\bigl(y(x,t)- b_t(x,t)\bigr)^2\, d\mu(x,t).
\label{VideoPredictionEq}
\end{equation}
Of course, as the visual agent gains the capability of predicting what
will come next, it means that the developed internal representation
based on features $\allphi$ cannot correspond with the mentioned
trivial solution. Interestingly, it looks like visual perception doesn't
come alone! The typical paired skill of visual prediction that animals exhibit
helps regularizing the problem of developing invariant features.
\marginpar{\small No system dynamics apart from that inherited by motion!}
More sophisticated predictions
can rely on a dynamical system to predict the features and the velocities.
However, this book makes the fundamental assumption of disregarding
dynamical processes for the representation of $(\allphi, \allv)$.
The computational models~\eqref{aggreg-eq}  and~\eqref{GlobalResp-y}
exhibit in fact a ``nearly-forward structure.'' This assumption is somehow related to the
hypothesis of picture perceivable actions that is introduced in detail 
in Section~\ref{CMI-sec} concerning the notion of affordance. 
It is claimed that system dynamics is needed for high-level human-like cognitive
processes (e.g. understanding actions like to tie the perfect bow or solving the Rubik's cube).
On the opposite, most animal and human actions do not need to involve
system dynamics but simply the dynamics connected with the optical flow of the
features. In any case, the prediction carried out by $y$
is assumed to be  rich enough, a process that vaguely reminds us of classic 
predictive coding~\cite{rao:natneuro99}. A recent survey on deep learning\index{deep learning} techniques
on the construction of the prediction scheme stated by Eq.~(\ref{VideoPredictionEq})
has been published in~\cite{DBLP:journals/access/ZhouDE20}.

%

A few comments are in order concerning the regularization term and the
functional which controls prediction.
\begin{itemize}
\item {\em Features $\allphi$ are expected to express visual 
properties with different degree of abstractness that is useful for 
perception as well as for action, since we need to gain prediction skills.} \\
We must bear in mind 
that those features are somehow characterizing symbolic properties of a 
given object. In particular, they are expected to embrace a  
non-negligible portion $\Omega_\varphi(x)$ of the retina from which 
one is expected to be able to provide descriptions. As we collect all
those ``descriptions'' we expect that they are sufficient to reconstruct the 
retina. Of course, since they are expressed by real numbers, they
address hidden semantics that we could find it hard to describe.
Yet, for any pixel, we collect formal descriptions that drive prediction.
For example, in order to help intuition, one can think of 
a red cat with bald spots and wet fur.
Notice that is  a truly global description of a whole object.
It could also be the case that the cat is given a more articulated symbolic description 
which refers to his head or to any other part of the body.
This kind of descriptions might be enough to identify a cat in a limited population and, consequently,
it could be enough for the generation of the corresponding picture.
The  $\allphi$ codes are basically of global nature and, 
therefore, the predictive reconstruction cannot exhibit trivial solutions.
Clearly, the $\varphi=c_{\varphi}$ which satisfies motion invariance is not acceptable
since it doesn't reconstruct the input. 
This motivates the involvement of prediction skills typical of action
that, again, seems to be interwound with perception. 

\item $E$ and $R$ {\em are functional, where the dependency on $\Omega_\varphi(x)$ 
is the sign of a corresponding spatiotemporal dependency induced by the FOA}. \\
As pointed out in the list of motivations for choosing foveated eyes 
in item~(\ref{VariableRR}) of  Sec.~\ref{Focus-section}, there are
good reasons for choosing variable resolution when considering the 
temporal structure of the computation and the focus of attention mechanism.
Moreover, still in the same list, the claim in~(\ref{DisambiguatingLT}) 
puts forward the limitations of weight sharing with the need of 
disambiguating decisions in different positions in the retina. 
Concrete computational mechanisms for carrying out this kind of computation
will be described in Section~\ref{VRes-sec} by discussing deep neural networks.

\item {\em Unlike what we have seen for Horn \& Schunck variational 
	formulation of the optical flow, here the weights depend on the
	 environmental time.} \\
	This is in fact at the basis of the variational structure of the learning
	processes considered in this book. 
\end{itemize}

%
%
\marginpar{\small Formulation of learning the pair $(\varphi,v)$ according to 
the Principle of Least Cognitive Action}
Even though the learning process will be described in Ch.~5, here we want to provide
a preliminary formulation of learning, which 
has its roots in related recent studies \cite{DBLP:journals/tcs/BettiG16,DBLP:journals/corr/abs-1808-09162}. Let  $S$ be the following functional 
\begin{equation}
	  S(\allphi, \allv)=  \lambda_{E}E(\allphi,\allv)
	 +\lambda_{R}R(\allphi,\allv) + \frac{1}{2} \sum_{i=1}^n\int_\Gamma
	  |\phi^i \Join v^i(x,t)|^{2}\, d\mu(x,t),
\label{CognitiveActionDef}
\end{equation}
that, throughout this book, is referred to as the  {\em cognitive action}. 
Here, $\lambda_{R}$ and $\lambda_{E}>0$ are the regularization parameters.
Learning to see means to discover the indissoluble pair $(\allphi^{\star},\allv^{\star})$
such that 
\begin{equation}
	(\allphi^{\star},\allv^{\star}) = \argmin_{(\allphi,\allv)} S(\allphi,\allv).
\label{LearningPos}
\end{equation}
Basically, the minimization is expected to return the pair $(\allphi, \allv)$,
whose terms should nominally be conjugated. The case in which we reduce
to consider only the brightness, that is when the only $\allphi$ is $b$, corresponds with 
the classic problem of optical flow estimation. Of course, in this case
the term $R$ is absent and the problem has a classic solution. 
Another special case is when there is no motion, so as 
the integrand $(\phi^i \Join v^i)^{2}$ is simply null $\forall i=1,\dots,n$. 
In this 
case the learning problem reduces to the unsupervised extraction of 
features $\allphi$.
In classic pattern recognition,
the world is dramatically characterized by the underlying principle 
that perceptual vision is in fact a  classification process. One defines
the classes in advance and the subsequent learning process
is affected by the choice of which classes are in fact characterizing
the given visual world. Hence, the introduced classification task
is somewhat biasing the interpretation of the given visual environment.
When the subtleties of language become 
an important issue to consider in visual descriptions, it becomes
early clear that one cannot reasonably define the classes in advance.
Ordinary visual descriptions consist of linguistic statements, which are well beyond
object categorization, with classes defined in advance. 
The compositionality of language
suggests that the visual learning processes centered around classification
are likely specializing towards a sort of artificial process. 
The fundamental requirement of prediction stated by~(\ref{VideoPredictionEq}) leads to 
reconstruct video, regardless of any prior categorization has been
eventually assumed on the objects. 

\marginpar{\small The universal property of codes $(\allphi^{\star}$ gained 
along with $\allv^{\star})$}
The well-posedness of learning according to~(\ref{LearningPos}) for the
 pair $(\allphi,\allv)$ can be established also without involving
the I Principle by imposing motion invariance. The prediction and the 
regularization terms yield in fact an appropriate definition of the minimum.
The codes $\varphi$ and the associated velocities $v_{\varphi}$ 
are extracted with the purpose of predicting $b_t$
and to gain uniform values as much as possible for both variables.
The introduction of motion invariance, however, produces the fundamental
effect of making the decisions consistent with motion.
It turns out that in the given visual environment, we need not express to express the solution
over the real field, but on the space of countable features. It is in 
fact the integer which identifies the feature that can be used for
representing the information that has been learned.
In a sense $(\allphi,\allv)$ are countable symbols over
the given visual environment. 
While this can always be assumed
in any neural network, we need to involve visual environments 
without the bias of classification for conquering visual interpretation. 
Again, the fundamental difference with respect to typical deep learning approaches to computer vision is that features are motion consistent, which in turn minimizes the amount of information needed to perform visual prediction.
\marginpar{\small Motion invariance and MDL}
The satisfaction of $\varphi \Join v_{\varphi} = 0$ corresponds with 
$d \varphi/dt = 0$ along moving trajectories.
Notice that this condition doesn't prescribe the constant value of $\allphi$ 
uniformly on the retina, but only on trajectories. When $v^i=0$ the conjugation 
yields constant in time $\phi^i$ (keep the decision on the pixel). While we expect  that
the $\varphi \neq  0$ be ``strongly met'' when the feature 
is active, we expect  $\varphi \simeq 0$ when it is non-active. 
Such a ``constrastive condition'' is not the outcome of invariance, since invariance only enforces 
$\varphi = c$. However, the additional regularization term penalizes $\varphi \neq 0$, 
which is the value that the feature takes on when it is non-active. Clearly, non-null values of the
features are sustained by need of performing frame prediction. 
We end up into the conclusion that any point of the retina, for any frame, either belongs to the partition where 
the feature is active or non-active, with constant value for 
$\varphi$ with both cases. Motion invariance suggests that the storage of the 
representation of $\varphi$ collapses to a finite representation. 
Notice that all this analysis holds for a functional space which admits abrupt 
transition from active to non-active values. However, the approximation capabilities of neural nets 
makes it possible to get very close to this ideal condition.

\begin{figure}
\centerline{
$\includegraphics[width=1.5cm,height=3cm]{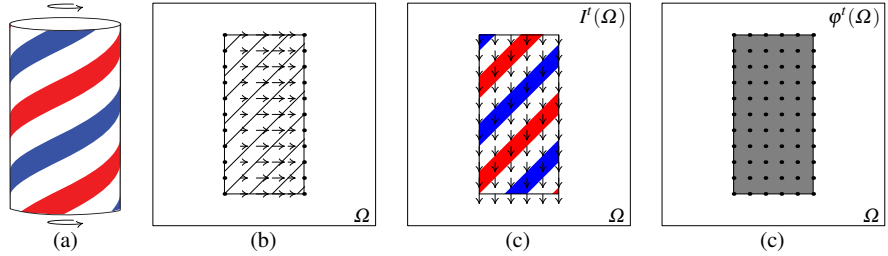}\atop\hbox{(a)}$\quad\hfil  
$\includegraphics[width=3cm,height=3cm]{./figs/8.mps}\atop\hbox{(b)}$\quad\hfil 
$\includegraphics[width=3cm,height=3cm]{./figs/9.mps}\atop\hbox{(c)}$\quad\hfil 
$\includegraphics[width=3cm,height=3cm]{./figs/10.mps}\atop\hbox{(c)}$\quad\hfil
}
\caption{\small Barber's pole example. 
(a) The 3-D object spinning counterclockwise.(b) The 2-D projection of the pole
and the projected velocity on the retina $\Omega$.
(c) The brightness of the image and its optical flow pointing upwards. (d) A feature map
that respond to the object and its conjugate (zero) optical flow.
}
\label{BarbPole-fig}
\end{figure}
\noindent {\em Barber's Pole and Degree of Invariance}\\
Now, let us consider another example of optical illusion can be produced by objects themselves.
This is very instructive in order to get an intuition on the meaning of velocity of features.
The following classic  {\em barber's pole} offers an intriguing example to understand the
importance of introducing appropriate optical flows for the visual features. 
As the pole rotates counterclockwise  it's hard to visually track the single pixels. 
Clearly, since they are the projections of
rotating material points, the velocity $v$ of a single pixel is a 
horizontal vector that can easily be 
computed when the rotational speed of the pole is given, but its visual perception 
does not arise, since the resulting optical flow is in fact an optical illusion.
Now, as we focus attention on the 
stripes of the pole, surprisingly enough, we can easily track them; 
they have an apparent upward motion. Now, suppose $\varphi_{r}$ is 
a feature that characterizes the red stripe, that is $\varphi_{r}(x,t) = 1$ iff $(x,t)$ is in inside a
stripe. As the barber's pole rotates,  the conjugated velocity $v_{\varphi_r}$
is  vertical. In this case, this velocity is the one which is also conjugated with the 
brightness under Horn and Schunck regularization stated by~(\ref{OpticalFlowReg}).
 An additional abstraction can be gained when looking
at the whole object: its conjugated velocity is zero, since we assume that the whole barber's pole
is fixed. Hence, while the optical flow
corresponding to single pixels can hardly be perceived, in this case, two features with 
an increasingly higher degree of abstraction are perfectly 
perceivable with their own conjugated velocities. 
This is very much related to the mentioned chasing skill of the cat!
In real-world scenes, the nice conjugation  of the vertical velocity with the stripes
in the barber's pole can hardly be achieved. Unlike the case of the brightness, however,
we can always think that a feature does represent a more distinguishing property that 
is associated with the object. The postulation of the existence of those features 
corresponds with stating their invariance. In general,
one can  believe in the indissoluble conjunction of features and velocities, 
stated by $\varphi \Join  v_{\varphi}=0$,
and, consequently, on their joint discovery. As we see the barber's pole stripes
we also track their vertical movement when the pole is rotating. The perception of the 
vertical movement comes with the perception of the corresponding stripe and vice versa.

\section{The principle of coupled motion invariance}
\label{CMI-sec}
%
%
As pointed out in Section~\ref{IdentityAffordanceSec}, visual perception encompasses of both object 
identification and affordance\index{affordance}. The  {\tt MPI} (I Principle) only involves identification.
In this section we begin addressing 
seminal Gibson's contributions~\cite{Gibson1966,plasticq}, which have inspired scientists for decades. 
His studies in the field of psychology have been contaminating the community of computer vision
(see e.g.~\cite{DBLP:conf/cvpr/GallFG11}) also in the era of deep learning~\cite{DBLP:journals/corr/abs-1807-06775}.
Object affordance is interwound with correspondent
actions, but studies on these topics are often carried out independently one of each other in computer vision. 
Interestingly, the interpretation of actions is typically regarded as a sequential process. By and large, actions 
are perceived as the outcome of a sequence of frames. 
Studies in deep learning for action recognition somehow aggregate features and optical flow.
Sequences are  taken into account mostly by combining Convolutional Neural Networks
and Long-Short-Term Memory recurrent neural networks~\cite{s19051005}, which are in charge
for sequential processing. 

\marginpar{\small Picture perceivable actions and the transmitted affordance}
Interestingly enough, humans can promptly perceive many common actions 
from single pictures, which has clearly an impact in the notion of affordance!
For example, we can easily recognize the act of jumping, as well as the act of pouring a liquid into a container
from a single picture, which can also be enough for understanding that somebody is eating or drinking. 
The act of taking a penalty kick or scoring a basket are also {\em picture perceivable actions}.
Hence, it looks like there are actions that are perceived from single
pictures without presenting video clips. 
It's important to realize that when shifting attention to actions, we need to focus on their 
meaning that is paired with sophisticated human experiences.
While we can recognize from a single picture the act of pouring a liquid into a container, 
which nicely yields  the correspondent
affordance, we cannot recognize the act of solving Rubik's cube (while we can say that the cube is beeing manipulated by someone). A single picture can convey
suspiciousness on the acts of killing, but it cannot suffice to drive the conclusion on the murder. Similarly, a photo in which I'm trying to make myself the bow
does not really correspond with the action labelled as ``how to tie the perfect bow'', since the photo could just 
trace my beginning act which might not result in a successful tie of the perfect bow! 
The acts of solving the Rubik's cube, killing, and tying the bow do require video clips for their interpretation, whereas 
other actions, like running, jumping, grazing the grass, and pouring a liquid into a container are picture perceivable.  
The natural flow of time conveys a sort of direction to actions like seating and standing up; they are similar, 
yet somewhat opposite in their meaning. The same holds for the act of killing and cutting the bread. 
We can hit with the knife or remove it from an injured person. Clearly, the simple act of removal can support an
important difference: the person who's acting might be the killer, who's just removing the knife for hitting again, or 
might be a policeman who has just been notified of the murder. Those actions definitely need 
a sequence-based analysis for their interpretation and they also support a different affordance, which has in fact a truly global nature. 
On the opposite, whenever actions are perceivable from single pictures the
corresponding affordance has a local nature.
Regardless of this difference, in general, more actions can refer to the 
same object (e.g. the chair and the knife), which means that the process of object 
recognition generally involves multiple affordances. The notion of chair clearly arises from both the acts of seating and standing up, but it also emerges when we 
use it for reaching an object.

\marginpar{\small Affordance is conveyed also when the action is not perceivable by a single picture}
The same holds true for many other objects. This discussion suggests that 
we transfer affordance to objects even with picture perceivable actions. Moreover, while some
actions, like cutting the bread and seating on a chair could still be confused when interpreted by 
a single picture,  it can be enough to convey affordance to the bread and the chair, respectively. 
Single pictures of humans carrying out actions on an object do convey affordance. While we can argue
on the degree of semantic information which is transmitted, in general, those pictures are definitely
important for understanding the function of the object. While solving the Rubik's cube 
does require the  frames of the entire video where the act is shown, 
also in this complex case a single picture on cube manipulation
does convey precious affordance.
The cube can be recognized as an object that can be manipulated, regardless of the fact that
the manipulation actually leads to the solution.
 When considering different actions on a certain object we
can immediately realize that they all convey affordance. For example, pictures of the acts of
cutting or breaking bread, both transfer an affordance that is useful for recognizing bread.
This discussion strongly pushes the importance of the {\em local notion of affordance}
and suggests that more abstract interpretations corresponding to portions of video
may not play the central role in the transmission of affordance. They are in fact conveying 
a more abstract and high level information that can be interpreted only by sophisticated
reasoning models. This goes beyond the basic principle of perceptual vision and 
action control and it is beyond the scope of this book.
\\ ~\\
\noindent {\em Second Principle of Perceptual Vision: Coupled Motion Invariance}  ({\tt CMI})\\
Like for object identification, the notion of affordance becomes more clear when  shifting
attention at pixel level, which is in fact what foveated animals do. First, we need to assume a sort of animal-centric view 
since object affordance is better understood when it is driven by an animal\footnote{As it will be better seen in the 
following, a more general view
on object affordance suggests that it can come also from another object which is moving, but the
following analysis clearly holds regardless of who's transferring the affordance}. 
This assumption is especially useful to favor the intuition behind the 
II principle of Perceptual Vision, but it will be removed and the analysis will be
extended to the case of visual features, thus paralleling the statement of the 
{\tt MPI} principle.
We shall begin discussing interactions between animals and objects that can be
identified by corresponding codes but, as stated in the following section, the ideas
herein presented  can be
given a more general foundations in terms of the interaction between objects in visual environments.
Without limitation of generality, suppose we are
dealing with human actions, and let us consider a pixel $x$ where both a person, with identity $p(x,t)$, 
and an object, with identity  $o(x,t)$, are detected.
As already noticed, in general, we can also assume that the object
defined by $p(x,t)$ is not necessarily referring to a person, but to tools typically used in human actions.
For instance, $p(x,t)$ could be the identifier of a 
hammer which is used to hang a picture. Interestingly, a pixel could contain the identity of the hammer, 
the picture, and the fingers of a person. In general, it could be the case that more than one 
object are defined at $(x,t)$, which depends on the extent to which we consider the contextual information.  
In that case, the action also gives rise to a corresponding number of optical flows at $(x,t)$ that are 
conjugated with the coupled objects. For the sake of simplicity, suppose
the optical flow defined by $v_{p}(x,t)$ in pixel $x$ at time $t$ comes from the person who's providing the
object affordance. 
\marginpar{\small II Principle of Perceptual Vision: Coupled Motion Invariance ({\tt CMI}):
The person identifier $p(x,t)$, paired with object identifier $o(x,t)$}
Following the conditions under which the I Principle of Visual Perception has been established,
let us consider a single movement burst delimited by saccadic movements. For any
$(x,t) \in \Gamma$ we can introduce the {\em coupling relation} $\between$ between objects as follows:
\[
	p\between o(x,t):=\gamma(p(x,t)) \land \gamma(o(x,t)).
\]
Here, $\gamma$ returns the Boolean decision behind $o$ and $p$; it is typically
$\gamma: [0,1] \to \{0,1\}$. A value is returned that is based on thresholding criteria.
Of course, depending on the measure of 
$\mathscr{C}_{p \between o}=\{\,(x,t)\in \Gamma: p\between o(x,t)=1\,\}$
we can experiment a different degree of object coupling, so as one can 
reasonably establish whether the field interaction between $p(x,t)$ and $o(x,t)$ is significant.
Given $\epsilon>0$ we say that the coupling $o \between p$ is $\epsilon$-significant
provided that\footnote{$\Leb^3$ here is the Lebesgue 
measure on $\bbR^3$}
$\Leb^3(\mathscr{C}_{p \between o})>\epsilon$ and write $o \between_{\epsilon} p$. 
\marginpar{\small {\em Degree of affordance} $\alpha_{op} \in \bbR$ of $o$ conveyed by $p$.}
Whenever this happens, it turns out to be convenient to introduce the 
{\em degree of affordance} $\alpha_{op} \in \bbR$ of $o$ conveyed by $p$.
Unlike the information on the object identification, the degree of affordance transmits the information on how the person defined by $p$ is using $o$.
We are now ready to establish the Coupling Motion Invariance principle 
as follows:
\marginpar{\small \\ Coupled Motion Invariance}
 \begin{equation}
  	{\tt CMI}:  \qquad
	 \alpha_{op} \Join (v_{o} - v_{p}) = 0.
\label{AffordanceMO}
 \end{equation}
This statement is very related to the I Principle. 
If the interacting object $p$ is not moving, that is $v_{p}=0$, then {\tt CMI} reduces
to $\alpha_{op} \Join v_{o}=0$, which is reminiscent of the constraint
$o \Join v_{o} = 0$. Hence, in this case $\alpha_{op}$ resembles the identifying
code $o$.
Now, there are good reasons
for thinking of the two principles separately and keep $\alpha_{op}$ and $p$
separated. The I Principle is about object identification, whereas
the II Principle is about affordance, and they are in fact conveying different, yet complementary,  
information! A visual agent needs to recognize chairs, but he also needs 
to identify the specific type of chair,
and he must also be capable of 
inducing that the specific object is in fact part of a more general category. Such a category  
may reflect a few distinctive visual properties of an object which also support
its function. This is in fact the reason for keeping $o$ and $\alpha_{op}$ separated. The difference emerges when $p$ interact with $o$ also 
by its velocity $v_{p}$. In this case, Eq.~(\ref{AffordanceMO}) still states the
motion invariance property of $o$, 
with respect to the relative velocity $v_{o}-v_{p}$.
It's worth mentioning
that this must be interpreted in the local sense and that {\tt CMI} and {\tt MPI}
are essentially stating the same property of motion invariance of $o$, but
the fundamental difference is the use of the relative velocity.
The presence of $p$ with velocity $v_{p}$ is in fact the
indication of the affordance conveyed by $p$ to $o$, which results in 
the different reference where the object velocity must be measured. 
When the fixed reference is
chosen we are electing the distinctive feature $o$ of the object, whereas
the presence of interacting objects leads to express invariance with 
respect to their reference. 

It's worth mentioning that the same person $p$ can propagate the  affordance
to objects with different identity $o_{1},o_{2}$. 
For example, we manipulate the Rubik's cube as well as a hammer, or a knife.
This induces  the relation $o_{1} \stackrel{p}{\sim} o_{2}$ 
which holds if
\begin{equation}
(p \between_{\epsilon} o_{1}) \quad\hbox{and }\quad  (p \between_{\epsilon} o_{2}).
\label{A-rel1}
\end{equation}
We notice in passing that this relationship doesn't assume a specific semantics in
the action carried out by $p$. While in the previous example we were considering
the act of manipulation on different objects, the above stated field interaction 
might also hold for touching, pushing, or shaking. Of course, this analysis could be
extended if one can classify actions, but this is not considered in this book\footnote{The
basic ideas are the same whenever in the coupling $o \between p$, the object identifier
$p$ is replaced with an action field.}.   
Two persons, defined by $p_{1},p_{2}$ can also transfer the same affordance to $o$,
which is represented by the dual relation 
with respect to~(\ref{A-rel1}) 	$p_{1} \stackrel{o}{\sim} p_{2}$
which means:
\begin{equation}
 (p_{1} \between_{\epsilon} o) \quad\hbox{and }\quad  (p_{2} \between_{\epsilon} o).
\label{A-rel2}
\end{equation}
The abstraction of Eq~(\ref{A-rel1}) leads to the notion of action.
For example, the same person can manipulate different
objects. This characterizes a certain action of $p$ which holds for 
a certain class of objects. We can in fact manipulate 
the Rubik's cube as well as the knife and many other objects.
The abstraction of Eq.~(\ref{A-rel2}) consists of considering the case
in which object affordance is conveyed independently of the person.
In this case  Eq~(\ref{A-rel1}) can be replaced with $o_1 \sim o_2$.
Overall, when a visual learning environment along with multiple 
interactions is considered we can create a partition on object identifiers $O$
which yields the quotient space ${\cal Q}=\mathscr{O} / \sim$ 
where the equivalent classes can be thought as the class of the object.

\marginpar{\small Inherent affordance}
Now, let us assume that in a visual environment there is collection
${\cal P}_{o}$ of person identifiers who are coupled with $o$. The joint 
coupling with all people in ${\cal P}_{o}$ leads to think of the corresponding 
affordance $\alpha_{o{\cal P}}$. However, as ${\cal P}_{o}$ becomes large enough
it does characterize the ``living environment'' and
it makes sense to think of the {\em inherent affordance notion} $\alpha_{o}$
for which the following condition holds true:
\begin{equation}
	\alpha_{o} \Join (v_{o}- v_{j}) =0 \qquad \forall j \in {\cal P}_{o}.
\label{FormalIA}
\end{equation}
For example, when considering the notion of knife, we can think of its
inherent affordance which is gained by the act of its manipulation by a virtually
unbounded number of people. Of course, Eq.~(\ref{FormalIA}) can only be 
given a formal meaning, since in real-world environments the notion
of inherent affordance can only be approximatively gained.
Unlike the single pairing of $o$ with $p$ where the affordance is modeled
by $\alpha_{op}$, here the inherent affordance $\alpha_{o}$ involves
a collection of constraints that is logically disjunctive over ${\cal P}_{o}$.
This can be expressed by the corresponding risk
\begin{equation}
	A = \sum_{j \in {\cal P}_{o}}\big(\alpha_{o} \Join (v_{o}-v_{j})\big)^{2}.
\end{equation}
\marginpar{\small Dealing with attached objects}The 
way we conquer a formal notion of affordance, stated by Eq.~\eqref{AffordanceMO} works for both {\em attached} and 
{\em detached objects}, since their movement is not necessary. The affordance comes from people who are using 
the object. However, the way people interact with attached and detached objects is definitely very different.  
We experiment an interaction with objects that can be directly  grasped  that is far away 
from the interaction with
big fixed objects. While  hammers can be grasped and moved so as their motion with respect  to the
environment can  be used for conquering their identity, a building can neither be grasped nor moved.
Yet, humans and attached objects can still interact so as they can get the affordance from
human movements. As we look out the window or open the shutters, the corresponding movement transfers
an affordance. Something similar happens as we walk close home.  Even though a chair is clearly
detached, it is typically fixed as we seat down or stand up. Of course, it can itself slightly move, but 
the affordance is basically gained from the person who's using it. 

\marginpar{\small Attached and detached objets involve a different optical flow; identification
is favored for detached objects.} 
 Because of the structure of the optical flow generated during object-human interaction, it's
 quite obvious that the identity of detached objects can better be gained. On the opposite, for both attached 
 and detached objects the affordance can be similarly gained. There's still a difference, since
 in detached objects the optical flow comes from both humans and objects.
 Hence, it looks like attached objects are likely more difficult to learn, since the
 development of correspondent identification features is more difficult than for 
 detached objects.

\section{Coupling of vision fields}
\label{CouplingVF-sec}
As discussed in the previous two sections, the I and the II Principles 
come from the need of developing two different, yet complementary, visual skills: 
object identification and affordance.
A certain discrepancy in expressing the two principles may not have gone unnoticed.
While the I Principle has been expressed on object features, the II Principle has involved 
object identifiers, that is higher level concepts. 
Obviously, this discrepancy could be reconciled by also stating the I Principle directly in terms
of object identifiers, so as in both cases we directly involve objects as atomic entities.
That would be fine, but the simple statement of the I Principle on the object feature
suggests that  both principles might be interpreted in terms 
of a {\em Vision Field Theory}, the purpose of which is that of studying the interactions 
between features and velocities. As already noticed, the II Principle doesn’t really need to make explicit
 the semantic of the actions that transfer affordance. 
There's one more reason to focus on features instead of objects, which is connected
with the chicken-egg dilemma discussed in Sec.~\ref{VideoEnvSec}: while we can define
the velocity of features by involving video information only, as it will be pointed out in
Sec.~\ref{obj-rec-sec}, this is not sufficient to fully capture the meaning of  attached objects.

\marginpar{\small Affordance is also conveyed by visual features with hidden semantics}
Hence a question arises on whether or not affordance can be gained also by ignoring the object identity
which transfers affordance. 
Notice that while it is gained by motion invariance of the same object, 
the essence of affordance  also comes from a motion invariance condition that involves 
the motion of who's using the object. However, in the statement of the II Principle there 
could be hidden semantics in the entities which convey affordance. As such, we  can 
think of visual features and of a more general view that involves their coupling.
\marginpar{\small $\psi_{ij}$ is the {\em feature affordance} that $\phi^{j}$ transfers to 
$\phi^{i}$: it comes from the coupling $\phi^i \between \phi^j$}
Let  us consider the collection of features $\{\,(\phi^{i}, v^{i}) :  i=1,\ldots,n\,\}$ which 
arises from a certain visual environment. 
In what follows we regard $\phi^{i}$ as scalar
features (i.e. in our previousely introduced notation $m_i\equiv 1$ for $i=1,\dots,n$). 
In general, these features cannot be easily associated
with object identifiers, but we can consider visual features as  interacting fields.
While the conjugation of self-coupling $\phi^{i} \Join v^{i}=0$ of the I Principle 
favors the feature identification, the coupling  $\phi^{i} \between \phi^{j}$, with 
$i \neq j$ can either come from the interaction of different objects or within the same object.
The first case corresponds with what we have already discussed concerning
object affordance. In the second case, we are still in front of a true field interaction 
between features; for example, we can move our hands while walking, which 
likely establishes a vision field where the features corresponding to the head are
somewhat  coupled with the velocities associated with the hands. 
Interestingly, while making these examples helps the intuition and it is definitely stimulating, 
the mathematical structure behind vision field coupling speaks for itself. 
Hence, we replace the given definition of affordance $\alpha_{op}$ by
introducing the field $\psi_{ij}\colon\Gamma\to\bbR$, which indicates the 
affordance that feature $\phi^{j}$ transfers to feature $\phi^{i}$.

\marginpar{\small Vision field-based restatement of the II Principle and symmetry  of $\psi_{ij}$}
Based on this premise on feature coupling we can provide a more general statement of the
II principle. When considering this general perspective, instead of assuming that
humans are transferring affordance to a fixed object, one must take into account
a velocity field for each interacting feature. 
Under these conditions, the II Principle can be stated as:
\begin{align}
\begin{split}
	{\tt CMI-bis}:\qquad
	 \psi_{ij} \Join (v^{i} - v^{j}) = 0\qquad  1\le i,j\le n.
\label{Pi-equation}
\end{split}
\end{align}
Like in the statement~(\ref{AffordanceMO}) of the II Principle which refers to objects,
$\psi_{ij}$  is conjugated with velocity $v^{i} - v^{j}$,
which is in fact the relative velocity of feature $\phi^i$ in the reference of
feature $\phi^j$, which tranfers the affordance. 
Notice that if affordance feature $\psi_{ij}$ doesn't receive any motion 
information from $\phi^{j}$, since $v^{j}=0$, then the consequent condition 
$\psi_{ij} \Join v^{i}=0$  is in fact a self-motion invariance corresponding
with the I Principle.
However, it can better be interpreted as a {\em regularization statement}
for $\psi_{ij}$ that leads to express consistency on the affordance interpretation of the
feature itself. In another words, it can be interpreted as a feature self-coupling, that is
as a reflective property of $\psi_{ij}$.

\marginpar{\small Inherent affordance $\psi_{i}$: it comes from the interactions with all
the environmental features} 
As already stated in the previous section, when the visual environment is given,
as time goes by, the object interactions begin obeying statistical regularities and
the interactions of feature $\phi^{i}$ with the other features become very well defined.
Hence, the notion of $\psi_{ij}$ can be evolved towards the {\em inherent affordance}
$\psi_i$ of feature $\varphi_{i}$. Based on Eq.~\eqref{FormalIA} we define
the inherent feature affordance as the function $\psi_{i}$ which satisfies 
\begin{equation}
	\psi_{i} \Join (v^{i}-v^{j}) =0,\qquad\forall 
	j=1,\dots,n,\qquad  1\le i\le n. 
\label{FormalIA-f}
\end{equation}
This reinforces the meaning of inherent affordance, which is in
fact a property associated with $\phi^{i}$ while living in a certain 
visual environment.
As a consequence, the identification feature $\phi^{i}$ pairs with the corresponding 
affordance feature $\psi_{i}$, so as the enriched vision field turns out to be defined by 
${\cal V}=(\phi,\psi,v)$. In a sense, $\psi$ can be thought of as the 
abstraction of $\phi$, as it arises from its environmental interactions. 
A few comments are in order concerning the visual field.\index{vision fields} 
\begin{itemize}
\item
The pairing of $\phi^i$ and $\psi_i$ relies on the same optical flow
which comes from $\phi^i$. This makes sense, since the inherent affordance
is a feature that is expected to gain abstraction coming from the interactions
with other features, whereas the actual optical flow can only come from 
identifiable entities that are naturally defined by $\phi^i$. 
\item 
The inherent affordance features carry significantly redundant information. 
This can be understood when considering especially high level features that 
closely resemble objects. While we may have many different chairs in a certain
environment, one would expect to have only a single concept of chair, which 
in fact comes out as the affordance of an object used
for seating. On the opposite, $\psi$ assigns many different affordance
variables that are somewhat stimulated by a specific identifiable feature.
This corresponds to thinking of these affordance features as entities that 
are generated by a corresponding identify feature. 
\item
The vision field ${\cal V}$ is the support for high-level decisions.
Of course, the recognition of specific objects does only involve the field $\phi^i$,
whereas the abstract object recognition is supported by features $\psi_i$.
\end{itemize}

\marginpar{\small Regularization from logic implication}
The learning of $\psi_{i}$ is based on a formulation that closely resembles
what has been done for $\phi^{i}$, for which we have already considered
the regularization issues. In the case of $\psi_{i}$ we can get rid of the
trivial constant solution by minimizing
\begin{equation}
	I_{\psi}=\sum_{i=1}^{n}\int_\Gamma (1-\psi_{i}(x,t)) \phi^{i}(x,t)\,
	d\mu(x,t),
\label{FromLogImplicfi}
\end{equation}
which comes from the p-norm translation of $\phi \stackrel{c}{\rightarrow} \psi$.
Here we are assuming that $\phi^i,\psi_i$ range in $[0,1]$, 
so as whenever $\phi^{i}$ gets close to $1$, it forces the same 
for $\psi_{i}$. This yields a well-posed formulation thus avoiding 
the trivial solution. 

The inherent affordance $\psi_i$ that we characterize with
Equation~\eqref{FormalIA-f}, as we already observed, is tightly related with the feature field $\phi^i$; indeed they share the 
same index $i$ and the velocity field $v^i$ characterizes the
development of $\psi_i$ in an essential way. 

In order to abstract the notion of affordance even further 
we can, for instance, proceed as follows: For each 
$k=1,\dots, n$ we can consider another set of fields 
$\chi_k\colon\Gamma\to\bbR$ each of which satisfy the following
condition
\begin{equation}\label{eq:chi-conj}
\chi_k\Join v^j=0,\quad j=1,\dots,n,\quad j\ne k.
\end{equation}
In this way the variables $\chi_k$ do not depend 
on a particular $v_i$ but they will need to 
take into account, during their development the 
multiple motion fields at the same time. 

\marginpar{\small Logic regularization}

In order to avoid the trivial constant
solution, like for $\psi$ we ask for the minimization of  the logic implication term
\begin{equation}\label{eq:logic-reg-chi}
	I_{\chi}=\sum_{\kappa=1}^{n}\int_\Gamma (1-\chi_{\kappa}(x,t)) 
	\psi_{\kappa}(x,t)\, d\mu(x,t).
\end{equation}
This comes from the p-norm translation of $\psi  \stackrel{c}{\rightarrow} \chi$.
While this regularization terms settles the value of $\chi_{\kappa}$
on the corresponding $\psi_{\kappa}$, notice that 
the motion invariance condition~\eqref{eq:chi-conj} doesn't assume 
any privilege with respect to the {\em firing feature} $\psi_{\kappa}$. 
Unlike for $\psi$, the motion invariance on features $\chi$ is directly driven
by coupled features only, which contributes to elevate abstraction and 
lose the link with its firing feature.

\marginpar{\small Reduction and abstraction}
Once the set of the $\chi_k$ is given, a method to select 
the most relevant affordances could be simply done by a 
linear combination. In other words a subselection of 
$\chi_1,\dots,\chi_n$ can be done by considering
for each $l=1,\dots, n_\chi<n$ the linear combinations
\begin{equation}\label{AbstractAffVar}
X_l:=\sum_{k=1}^n a_{lk} \chi_k,
\end{equation}
where $A:=(a_{lk})\in\bbR^{n_\chi\times n}$ is a matrix of
learnable parameters. Notice that since $X_l\in\spn(\chi_1,
\dots,\chi_n)$, as we remarked in Section~3.4 satisfies 
$X_l\Join v^j=0$ for all $j=1,\dots,n$.
We can also extend the neural-like computation stated by Eq.~\eqref{NeuronLikeFea} 
so as to gain feature abstraction and reduction. The corresponding feature is
denoted by $\vdash \chi$. It's worth mentioning that the learning of coefficients
$a_{lk}$ doesn't involve motion invariance principles. Interestingly, as stated in 
Section~\ref{obj-rec-sec}, they can be used for additional developmental steps 
like that of object recognition. For example, they can be learned under the
classic supervised framework along with correspondent regularization.
%
%
\begin{table}
\caption{\textbf{Vision Field Theory}. Summary of the I and the II Principles of Visual Perception. The self, weak and strong-interactions obey 
obey conjugation equations derived from those principles. The reduced fields, denoted by $\vdash$, produces features with 
increasing degree of abstractness.
}
\begin{tabular}{ccccc}
\toprule
&\textbf{Self} &\textbf{Weak} & 
\textbf{Strong}  &\textbf{Reduction} \\
\midrule
\smallskip
\textit{VF}  &
 $\phi$ & $\psi$ & $\chi$ & $\vdash \phi, \vdash \psi, \vdash \chi$  \\

\textit{I, II Pr.} & $\phi^{i} \Join v^{i}=0$ &  $\psi_{i} \Join (v^{i}-v^{j})=0$ & $\chi_{i} \Join v_{\kappa}=0$ & $\vdash u = \sum u$\\

\textit{Reg.} &
 \ $\min E$, Eq.~(\ref{GenOpticalFlow})),\ \ $\ \min R$, Eq.~(\ref{VideoPredictionEq})
 & \ $\varphi \stackrel{c}{\rightarrow} \psi$, Eq.~(\ref{FromLogImplicfi}) &  $\ \psi \stackrel{c}{\rightarrow} \chi$ &  $\min \sum_{\kappa} a_{\kappa}^{2}$  \\
\bottomrule
\end{tabular}
\label{VFT-tab}
\end{table}

\noindent {\em Why a vision field theory?}\\
\label{WhyVFT-sec}
Tab.~\ref{VFT-tab} gives a summary of the Vision Field Theory, which reminds
us the role of the {\em weak}, {\em strong}, and {\em reduction} fields 
along with their interactions and regularization principles.
We find it interesting to point out that the I and II Principles are independent of the
visual agent, and they naturally emerge around the conjugation operator $\Join$,
which expresses feature invariance with respect to motion. 
The richness of the visual information, the body of the agent, and the number
of visual features come in place whenever we consider the specific visual agent which is involved
in the learning process.
Of course, rich visual sources do require a lot of visual  and
reduction fields, which are also likely playing a prominent role in those cases.
The presence of many features requires a correspondent computational 
structure that, in this book, is supposed to be a neural architecture. 
In the next chapter, the role of such a neural architecture is in fact
discussed in details and the connections with biology is explored. 
Interestingly, it's only when we involve the 
neural architecture that we begin appreciating the crucial role of FOA,
which is even better disclosed as we think of the learning process.

\marginpar{\small Feynman's understanding on the birth of the $E-B$ 
electromagnetic field in a capacitor}
\begin{figure}
\centerline{
\includegraphics{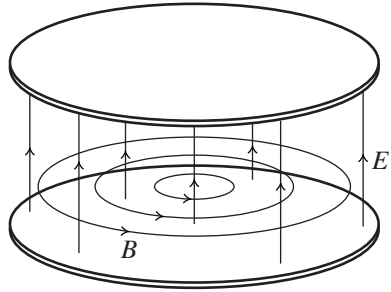} 
}
\caption{Richard Feynman's view on the electromagnetic field as a successive refinement process
to solve Maxwell's equations in a capacitor. The magnetic field arises with a circular 
structure as we feed the capacitor with AC. This, in turn, leads
to generating the induced $E$ fields.}
\label{Feynman-fig}
\end{figure}
The nature of the vision fields leads to intriguing connections with
electromagnetism that emerge when restricting to the information-based principles
dictated by vision fields. Let us consider the
very interesting interpretation on the emergence of the $E$-$B$
fields given by Richard Feynman in his lectures of 
physics~\cite{feynman2011feynman}.  
In order to explain how cavity resonators work, he offers a nice view on the 
$E$-$B$ interactions by beginning from the capacitor of  Fig.~\ref{Feynman-fig}.
He pointed out that there's in fact a nice picture behind the emergence of the fields 
which comes from the mutual successive refinement that one can clearly see
beginning from DC and then move to a signal with a certain frequency. 
At the beginning there is a spatially uniform electric field, whereas
the magnetic field is absent. As we move to a signal which changes over time
the displacement current $\epsilon \partial_{t} E$  generates a magnetic field
with circular symmetry. Since it changes itself with time, it subsequently generates an
electric field and, therefore, we end up into a cyclic mutual interaction
between $E$ and $B$.
Interestingly, this way of interpreting the interaction leads to the correct
solution, where the electric field, at a certain
frequency becomes null on the border of the capacitor. In particular, when the 
applied signal has a frequency defined by $\omega = 2\pi f$ then we have
\[
    E = E_{o} e^{j\omega t} J_{o}\left(\frac{\omega r}{c}\right),
\]
where $r$ is the radius of the plates, $c$ is the speed of light, and $J_{o}$ is
the Bessel function. This leads us to promptly understand how a cavity resonator can 
be constructed from the capacitor. In Feynman's words:  ``... we take a thin metal sheet and cut a strip
just wide enough to fit between the plates of the capacitor. Then we bend it into a cylinder 
with no electric field there ...''

This successive refinement scheme used for capturing the $E$-$B$ electromagnetic
interaction can be 
used also for vision fields\index{vision fields}. In our case the field interaction is governed by 
the conjugation  operator $\Join$. Like for electromagnetism, there's in fact
an external environmental interaction, the video signal, which fuel the fields.
The analogy is maintained when considering the case of DC in the
capacitor. This corresponds with a spatiotemporal constant video that
leads to constant features ($E$ field) and null velocity ($B$ field). 
As the video information comes, thanks to the presence of motion 
(the signal in the capacitor is not constant anymore),
the change of the features allows us to track their change by developing the
corresponding velocity\footnote{Think of the classic brightness invariance or to the
I Principle.}. Of course, that motion enables a corresponding
adaptation of the visual fields and, like for the capacitor, we 
end up into a cyclic process that, hopefully, converges. 
We conjecture that this concretely happens whenever the video source 
presents appropriate statistical regularities. The other side of the medal
is that of thinking of an upper bound on the corresponding amount of information
that is associated with the source.

\chapter{Foveated  neural networks}
\label{DNN-sec}
\AtBeginShipoutNext{\AtBeginShipoutUpperLeft{%
  \put(1in + \hoffset + \oddsidemargin,-5pc){\makebox[0pt][l]{
\boxformat 
  \framebox{\vbox{
\hbox{Published by Springer, Cham---Cite this chapter:}
\smallskip
\hbox{\url{https://doi.org/10.1007/978-3-030-90987-1_4}}}
  }}}%
}}

\vspace{-4cm}
\begin{quote}
The remarkable properties of some recent computer algorithms for neural networks seemed to promise a fresh approach to understanding the computational perspectives of the brain. Unfortunately most of these neural nets are unrealistic in important respects.\\
~\\
Francis Crick, 1989
\end{quote}

\marginpar{\small The slippery topic of biological plausibility}
\section{Introduction}
This
chapter is about visual bodies, regardless of biology. 
More then thirty years ago, Francis Crick contributed to activate a 
discussion on the biological plausibility of artificial neural
nets along with the corresponding learning algorithms.
As yet, this is still an active research direction carried out in a 
battlefield that is full of slippery and controversial issues. 
In the same paper of the above quotation~(\cite{Crick89}), the author
clearly states that 
``a successful piece of engineering is a machine which  does 
something useful. Understanding the brain, on the other hand, is 
a scientific problem.'' He noticed that this scientific problem 
is a sort of reverse engineering on the 
product of an alien technology, it's the task of 
trying to unscramble what someone else has made. 
The above statements can hardly be questioned.
Biological processes cannot neglect the importance of effectively
sustaining a wide range of simultaneous tasks connected with the 
essence of life. However, the impressive development of deep learning 
in the field of computer vision clearly indicates that there  are in fact
artificial intelligence\index{artificial intelligence} processes that make it possible to achieve concrete
performance, especially relying on a the truly different computational 
structure. The massive supervised learning protocol, along with the  Backpropagation algorithm are at the basis of spectacular
results that couldn't be figured out at the dawn of the connectionist
wave mostly carried out by the Parallel Distributed 
Processing (PDP)\index{deep learning} research group~\cite{Hinton86a,Hinton86b}.
Interestingly, the introduction of the connectionist models and the 
development of deep learning~\cite{lecun2015deeplearning} are not only
having a great technological impact, but they are also opening the doors
to the conception of computational processes that hold regardless of 
biology. 
\marginpar{\small Foveated-based  neural networks}
Considering the information-based principles and the vision field arguments 
on motion invariance discussed so far, in this chapter we 
claim that the computational mechanisms  behind motion invariance  can naturally\index{motion invariance}
be implemented by deep architectures that, throughout this book are
referred to as {\em Foveated-based  Neural Networks} (FNN).
They can in fact naturally carry out the conjugation of $\alpha_{\varphi}$
and  $\alpha_{v}$ (see Eq.~\ref{aggreg-eq}). This is expected to give rise to a machine 
capable of extracting codes that are motion invariant and that 
turn out to be very useful for visual perceptual tasks.
While deep learning has mostly been sustained by the supervised learning
protocol in computer vision, here deep architectures 
are only exposed to visual information provided by motion, with no labelled data. 
In this perspective FNN share with humans the same learning protocol
and, hopefully, can also shed light on brain mechanisms of vision. 
The information-based principles that are presented hold regardless of biology. 
This chapter is organized as follows: In the next section we address the crucial
role of hierarchical architectures along with the assumption of using 
receptive fields for neural computation. In Sec.~\ref{WW-mainstream-s}
we discuss the different role of features charged of detecting ``what'' and ``where'' is in the
visual scene and establish some biologic connection. In Sec.~\ref{VRes-sec}
we discuss variable resolution neural nets along with the relationship
with classic convolutional nets and we address 
the overall neural architecture.


\section{Why receptive fields and hierarchical architectures?}
 Beginning from early studies on the visual structure of the cortex~(\cite{hubel62}), 
 neuroscientists have
gradually gained evidence on the fact that it presents a hierarchical 
structure and that neurons process the video information on the basis of inputs 
restricted to receptive fields. 
Interestingly, the recent spectacular results of convolutional neural networks 
suggests that hierarchical structures based on neural computation with 
receptive fields play a fundamental role also 
in artificial neural networks~(\cite{lecun2015deeplearning}). 
As stated in Question 5 ({\bf Q5}, Sec.~\ref{10Q-sec}) we may wonder why are there 
visual mainstreams organized according to a hierarchical architecture 
with receptive fields and, if there is any reason why this solution has been
developed in biology.
First of all, we can promptly realize that, even though neurons are restricted
to compute over receptive fields, deep structures rely on large virtual contexts 
for their decision.  As we increase the depth of the neural network, 
the consequent pyramidal dependence that is established by the receptive
fields increases the virtual input window used for the decision, so higher
abstraction is progressively gained as we move towards the output. 
Hence, while one gives up on exploiting all the information available at a certain
layer,  the restriction to receptive field does not prevent from 
considering large windows for the decision. The marriage of receptive fields
with deep nets turns out to be an important ingredient for a parsimonious 
and efficient implementation of  both biological and artificial networks.
In convolutional neural networks, the assumption of using receptive fields
comes with the related hypothesis of {\em weight sharing} on units that are 
supposed to extract the same feature, regardless of where the neurons are
centered in the retina. In so doing we enforce the extraction of the same 
features across the retina. 
The same visual clues
are clearly positioned everywhere in the retina and the equality constraints on the weights
turn out to be a precise statement for implementing a sort of {\em equivariance
under translation}. 

%
%
Clearly, this constraint has effect neither on invariance under scale nor under rotation. 
Any other form of invariance that is connected with deformations is clearly
missed and is supposed to be learned. The current technology of convolutional
neural networks in computer vision typically gains these invariances thanks to 
the power of supervised learning by ``brute force''. Notice that since most
of the tasks involve object recognition in a certain environment, the associated limited 
amount of visual information allows us to go beyond the principle of extracting 
visual features at pixel level. Visual features can be shared over small windows
in the retina by the process of pooling, thus limiting the dimension of the network.
Basically, the number of features to be involved has to be simply related to the
task at hand, and we can go beyond the association of the features with the
pixels. However, the conquering of human-like visual skills is not compatible with
this kind of simplifications since, as stated in the previous section, humans can
perform pixel semantic labeling. There is a corresponding trend in computer 
vision where convolutional nets are designed to keep the connection with each
pixel in the retina at any layer so as to carry out segmentation and semantic pixel
label. Interestingly, this is where we need to face a grand challenge. So far, 
very good results  have been made possible by relying on massive labelling of collections of images.
While image labeling for object classification is a  boring task, human pixel labeling 
(segmentation) is even worst! 
Instead of massive supervised labelling, one could realize that motion and focus of attention\index{focus of attention}
can be massively exploited to learn the visual features mostly in an unsupervised
way. A recent study in this direction is given in~\cite{Betti2018-CN-arxiv}, 
where the authors provide evidence of the fact that
 receptive fields do 
favor the acquisition of motion invariance which, as already stated, is the 
fundamental invariance of vision. 
The study of motion invariance leads to dispute the effectiveness and the biological 
plausibility of convolutional networks. First, while weight sharing 
allows us to gain translational equivariance on single neurons, 
the vice versa does not hold true. 
We can think of receptive field based neurons organized in a hierarchical architecture
that carry out translation equivariance without sharing their weights. 
This is strongly motivated also by the arguable biological plausibility of the 
mechanism of weight sharing~\cite{OTT2020235}. Such a lack of plausibility 
is  more serious than the supposed lack of a truly local computational  
scheme in Backpropagation, which mostly comes from the lack of
delay in the forward model of the neurons~\cite{DBLP:journals/corr/abs-1912-04635}.

\begin{svgraybox}
Hierarchical architectures is the natural
solution for matching the notion of receptive field and 
develop abstract representations. This  also seem to 
facilitate the implementation of motion invariance, 
a property  that is at the basis of the biological structure of the visual 
cortex. The architectural incorporation of this fundamental invariance 
property, as well as the match with the need for implementing the 
focus of attention mechanisms, likely need neural architectures that
are more sophisticated than current convolutional neural networks.
In particular, neurons which provide motion invariance likely benefit
from dropping the weight sharing constraint.
\end{svgraybox}

\section{Why two  different mainstreams?}
\label{WW-mainstream-s}
In Section~\ref{Unbilicalsec} we have emphasized the importance of bearing in
mind the neat functional distinction between vision for action and vision for perception.
A number of studies in neuroscience lead to conclude that 
the visual cortex of humans and other primates is composed of two main information pathways
that are referred to as the  ventral stream and dorsal 
stream~\cite{GoodaleMilner92,GOODALE1998181}.
We typically refer to  the ventral ``what'' and the dorsal ``where/how'' 
visual pathways.  The ventral stream is devoted to perceptual analysis of the visual input, such as object recognition, whereas the dorsal stream is concerned with providing spatial localization and
motion ability  in the interaction with the environment. 
The ventral stream  has strong connections to the medial temporal lobe (which stores long-term memories), the limbic system (which controls emotions), and the dorsal stream.
The dorsal stream stretches from the primary visual cortex (V1) in the occipital lobe forward into the parietal lobe. It is interconnected with the parallel ventral stream which runs downward from V1 into the temporal lobe.
We might wonder: Why are there two different mainstreams? What are the reasons for such 
a different neural evolution? (see {\bf Q6}, Sec.~\ref{10Q-sec}).
This  neurobiological distinction arises for effectively facing visual tasks that are
very different. The exhibition of perceptual analysis and object recognition clearly requires 
computational mechanisms that are different with respect to those required for estimating the 
scale and the spatial position. Object recognition requires the ability of developing strong 
invariant properties that mostly characterize the objects themselves. By and large
scientists agree on that objects must be recognized
independently of their position in the retina, scale, and orientation. While we subscribe this point of 
view, a more careful analysis of our perceptual capabilities indicates that these desirable 
features are likely more adequate to understand the computational mechanisms 
behind the  perception  of rigid objects.
The elastic deformation and the projection into the retina gives in fact rise to remarkably more
complex patterns that can hardly be interpreted in the framework of geometrical invariances.
We reinforce the claim that {\em motion invariance is in fact the only invariance which does matter}. 
Related studies in this direction can be found~\cite{bertasius2021spacetime}.
As the nose of a teddy bear approaches children's eyes it becomes larger and larger. Hence, 
scale invariance is just a byproduct of motion invariance. The same holds true for rotation 
invariance. Interestingly, as the children deforms the teddy bear a new visual pattern 
is created that, in any case, is the outcome of the motion of ``single object particles''. 
The neural enforcement of motion invariance is  conceived by 
implementing the ``what'' neurons.
Of course, neurons with built-in motion invariance are not adeguate to make spatial estimations
or detection of scale/rotation. Unlike the ``what'' neurons, in this case motion does matter
and the neural response must be affected by the movement. 

\begin{svgraybox}
These analyses are  consistent with neuroanatomical evidence and suggest that 
``what'' and ``where'' neurons are important in machines
too.
The anatomical difference of the two mainstreams is in fact the outcome of a truly
different functional role. While one can ignore such a difference and rely on the 
rich representational power of big deep networks, the underlined difference 
stimulates the curiosity of discovering canonical neural structures to naturally 
incorporate motion invariance, with the final purpose of discovering different features
for perception and action. The emergence of the indissoluble pair $(\varphi, v_\varphi)$ helps
understanding the emergence of neurons specialized on the different functions
of perception and action.    
\end{svgraybox}


\section{Foveated nets and variable resolution}
\label{VRes-sec}\index{foveated neural networks}

\marginpar{\small Given a video, for conjugate features to exist we need an appropriate 
computational model, which corresponds with expressing the structure of 
the pair $(\varphi,v)$. 
}
The implementation of motion invariance
based  on conjugate features does require
the involvement of the specific structure of $\varphi$ and $v_{\varphi}$. 
Let us assume that they come from 
neural networks that carry out computations on pixel $x$ at time $t$
of the given video signal.
Of course, the neural networks are expected to take a decision at $t$ that
depends on the entire frame and on the specific pixel of coordinate $x$ where we 
want to compute the features. 
As already pointed out, following Eq.~\eqref{aggreg-eq},
we can consider deep neural networks based
on receptive fields, which means that we can always regard the 
output  of  neurons as  functions of  the corresponding virtual window. 
Let us focus on the computation of features $\varphi$. In that case suppose  
$\Omega_\varphi(x)$ is the  virtual window corresponding with 
$\varphi$ that is centered on $x$. 
Two deep neural networks can be used to express functions $\alpha_{\varphi}$
and $\alpha_{v}$. This leads to adopt the following computational model (compare what follows 
with Eq.~\eqref{aggreg-eq}):
\begin{equation}
\begin{aligned}
&\varphi(x,t) = \alpha_{\varphi}(b(\cdot,t),a(t),t)(x)
	= \eta_{\varphi}\big(w_{\varphi}(\cdot,t,a(t)),b(\cdot,t)\big)(x) \\
&v_{\varphi}(x,t) = \alpha_{v}(\tgrad \varphi(\cdot,t,a(t),t)(x) = 
	\eta_{v}\big(w_{v_{\varphi}}(\cdot,t,a(t)), \nabla \varphi(\cdot,t)\big)(x),
\label{conv-eq}
\end{aligned}
\end{equation}
Of course, the conjugation $\varphi \Join v_{\varphi}$ does involve\index{feature conjugation}
the choice of both $\eta_{\varphi}\colon
\bbR^N\times\bbR^\Omega\to \bbR^\Omega$ and $\eta_{v}
\colon
\bbR^M\times\bbR^\Omega\to \bbR^\Omega$, along with their weights
$w_{\varphi}\colon\Gamma\times\Omega\to\bbR^N$ and $w_{v_{\varphi}}
\colon\Gamma\times\Omega\to\bbR^M$ that are expected to depend on the
pixel on the retina and on the point of focalization of attention around which the foveated structure, as we will see
in what follows, is centered.
This is a remarkable difference with respect to convolutional
networks, where one assumes the weight sharing of the weights.
For CNN, the weight sharing is a fundamental constraint to make the feature extraction
independent of the pixel.
However, the assumption of acquiring visual frames under the driving
mechanisms of focus of attention changes the cards on the table.
\marginpar{\small Variable resolution and weights depending on the position in the retina}
As pointed out in the list of motivations for choosing foveated eyes 
in item~(\ref{VariableRR}) of  Sec.~\ref{Focus-section}, there are
good reasons for choosing variable resolution when considering the 
temporal structure of the computation and the focus of attention mechanisms.
Moreover, still in the same list, the claim in~(\ref{DisambiguatingLT}) 
puts forward the limitations of weight sharing when considering the need of 
disambiguating decisions in different positions in the retina. 
Interestingly, as we assume higher resolution  close to the
focus of attention we somewhat face the issue reported in item~(\ref{DisambiguatingLT}). 
In the following, we discuss the impact of variable retina resolution, which is strongly 
connected with FOA, in the architecture of neural network $\eta_{\varphi}$.
Of course, the same arguments hold for neural networks $\eta_v$.

\begin{figure}
\centering
\includegraphics{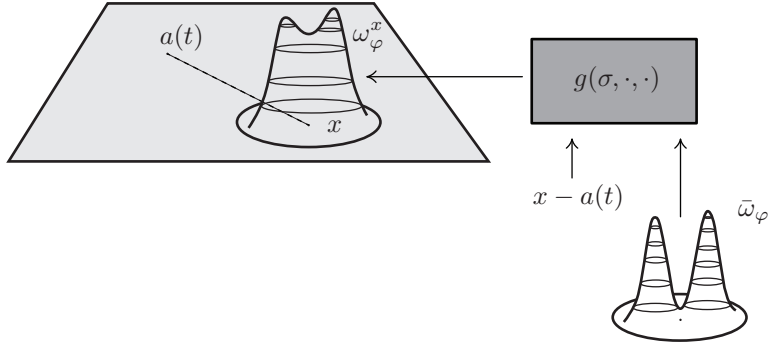}
\caption{The computation of the 
filters $\omega^x_\varphi$ is done 
through the function $g$ once the distance 
from the focus of attention is established.}
\end{figure}

Now let us regard the weights 
$\omega_{\varphi}^{x}$ of the
single neurons of $\eta_{\varphi}$.
For every fixed point on the retina $x\in\Omega$, and for every temporal instant $t\in[0,T]$
let $\omega_\varphi^x(\cdot,t)\colon\bbR^2\to\bbR$ be a compactly supported filter such that 
\begin{itemize}
\item $0\in\supp(\omega_\varphi^x(\cdot,t))\subset\bbR^2$;
\item $\diam\supp(\omega_\varphi^x(\cdot,t))<\diam\Omega $,
\end{itemize}
where $\diam A:=\sup\{|x-y|: x,y\in A\}$ and $\supp$ is the support 
of the filter.
In the continuum setting, those filters are used to compute
the activation $\upsilon$ of any neuron of $\eta_{\varphi}$ as  
\begin{equation}\label{eq:activations_FNN}
\left\{
\begin{aligned}
&\upsilon_\varphi(x,t)=(\omega_\varphi^x(\cdot,t)\star \hat y(\cdot,t))(x)
:=\int_{\bbR^2}\omega_\varphi^x(x-\xi,t)\hat y(\xi,t)\, d\xi,\cr
&\varphi(x,t)=\gamma(\upsilon_\varphi(x,t))
\end{aligned}\right.
\end{equation}
where $y\colon\Omega\times[0,T]\to\Real$ is the output of the
previous layer\footnote{Here for simplicity we are considering the scalar
case, although the extension to vectorial features is straightforward},
$\hat y(\cdot,t)\colon\Real^2\to\Real$ is the extension of
$y$ to $\Real^2$ with $\hat y(x,t)=0$ for all $t\in[0,T]$ if
$x\notin\Omega$ (zero padding).
As usual, 
we assume that the output $\varphi$ is computed from the activation $\upsilon_\varphi$
by the neural output function $\gamma$.  

Consider now the following way to compute the 
receptive field-based filters $\omega_\varphi^x$:
\begin{equation}\label{eq:filtersfromtemplate}
\omega_\varphi^x(z,t)=\int_{\bbR^2} g(\sigma,x-a(t),z-\xi) \hat{\bar\omega}_\varphi(\xi,t)\, d\xi,
\end{equation}
where $t\mapsto a(t)$ is a given trajectory (in our case it will be the trajectory
of the focus of attention), $\bar\omega_\varphi\colon K\times[0,T]\to\bbR$ 
is a single set of weights 
for each temporal instant defined over the compact set $K\subset\bbR^2$ containing the origin
and $g\colon\overline\bbR_+\times\bbR^2\times\bbR^2\to\bbR$
is an appropriate filter which depends on the parameter $\sigma$ and takes into 
account the position of the point $x\in\Omega$ with respect to the focus of attention. 
Clearly if
$g(\sigma,d,\cdot)$ is compactly supported, if
$0\in\supp(g(\sigma,d,\cdot))$ and if $\diam K+
\diam\supp(g(\sigma,d,\cdot))<\diam\Omega$ for any $\sigma\in\overline\bbR_+$ and 
for every $d\in\bbR^2$, then the two above conditions are satisfied.

\smash{\raise-7pc\rlap{\kern 13pc \includegraphics{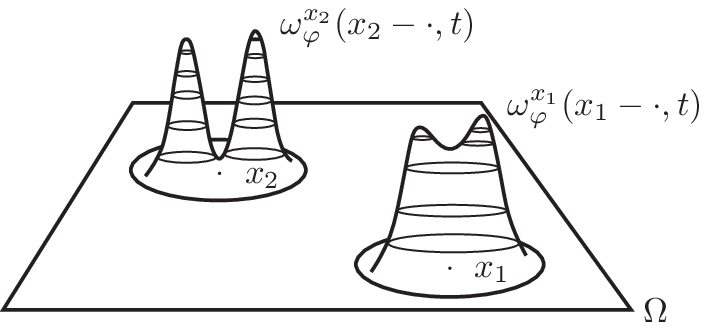}}}

\parshape 8 0pc 17pc 0pc 16pc 0pc 15pc 0pc 14.5pc 0pc 14pc 0pc 13pc 0pc 12pc 0pc \hsize  
Now, let's analyze the role of the focussing function $g$. We consider two 
extreme cases that give insights on the role of $g$ under the assumption that
$g(\sigma,d,x)=G(|d|,x)$ where $G\colon\overline\bbR_+\times\bbR^2\to\bbR$
and such that $\lim_{\eps\to0}G(\eps,x)=\delta$ (near the focus of 
attention the filter $g$ is extremely sharp). We also assume that $\lim_{z\to\diam\Omega}G(z,\cdot)=C$ i.e. far away from the focus of 
attention the filter $g$ become flat and constant\footnote{
This assumption is in contradiction with 
the fact that $g$ should have compact support; however it 
saves us from taking into account boundary effects that 
would raise technical issues that in our opinion would 
go against the purpose of giving clear insight on the role 
of $g$.
}
\index{focus of attention}

\begin{enumerate}
\item $|x-a(t)|\approx 0$: In this case under the above assumptions $g(\sigma,d,x)=\delta(x)$;
then it immediately
follows from Eq.~\eqref{eq:filtersfromtemplate} that $\omega^x_\varphi\equiv\hat{\bar\omega}_\varphi$.
\item  $|x-a(t)|\approx \diam\Omega$: Here we simply have, due to 
the fact that $\lim_{a\to\diam\Omega}G(a,\cdot)=C$,
\[\omega^x_\varphi(z,t)=C\int_K\bar \omega_\varphi(\xi,t)\,d\xi
=:C\mathcal{L}^2(K) \langle\bar\omega(\cdot,t)\rangle_K,\quad
\forall z\in\bbR^2,\]
where $\mathcal{L}^2(K)$ is the Lebesgue measure of $K$. Plugging this
expression back into Eq.~\eqref{eq:activations_FNN} we get
\[
\upsilon_\varphi(x,t)=C\mathcal{L}^2(K) \langle\bar\omega(\cdot,t)\rangle_K\int_{\bbR^2}\hat y(\xi,t)\,d\xi
=:C\mathcal{L}^2(K) \langle\bar\omega(\cdot,t)\rangle_K
 \langle y(\cdot,t)\rangle_\Omega,
\]
which means that when we are away from the focus of attention the 
value of the activations are spacially constant which is what we 
expect from a low-resolution signal.
\end{enumerate}

It turns out that FNN are based on filters that are somewhat in between those
associated with these two extreme cases.
Roughly speaking, this forces the learning process to be very effective where the
fovea is focussing attention, whereas as we move far away, 
the learning process only acts on possible modifications of the average of the weights
of the filter. This somewhat reflects the downsampling that arises on those pixels
where only the spatial average of the signal on the receptive field is actually perceived.
It's worth mentioning that the removal of 
the weight sharing principle leads to promote
a learning process which is stimulated more to adapt the parameters 
on the basis of the portions of the retina where we focus attention. In any pixel which is 
far away from the focus, we only take into account the average
of the weights, so as to better face the ambiguity issue~(\ref{DisambiguatingLT}).
There's a very good reason for assuming that the weights of the filter 
$\bar{\omega}_{\varphi}$ must be averaged when we are far away from the FOA.
\marginpar{\small Gaussian focussing filter}
%
%
A possible choice for $g$ is that of using the family of 
truncated Gaussians\footnote{Remember that $g$ has to have compact support.}
\begin{equation}
	g(\sigma,d,x):= \frac{1}{\sqrt{2 \pi}(\sigma+|d|)}
	\exp{\bigg(-\frac{x^2}{2(\sigma+|d|)^2}\bigg)}
	1_{\{|x|<\rho\}}(x).
\end{equation}
Here $\rho$ represent the radius at which we cut the tails
of the Gaussian to zero.
%
%
If we regard  $\sigma$  as a learnable parameter 
which depends on $t$ then the learning process could begin with very high
level of $\sigma$ and decrease to values close to zero.
Hence, at the beginning of learning we have $\sigma \rightarrow \infty$
which yields null weights. As time goes by, the value of $\sigma$ is 
gradually reduced until $g \rightarrow 0$ which, as already stated,
yields around
the focus of attention the maximum visual acuity stated by  $\omega^x_\varphi\equiv\hat{\bar\omega}_\varphi$.

\parshape 8 0pt \hsize 0pt 14pc 0pt 13pc 0pt 11pc 0pt 11pc 0pt 12pc 0pt 13pc 0pt 13pc
In the above discussion we always regarded the parameter
$\sigma$ that defines the family of maps $(d,x)\mapsto g(\sigma,d,x)$ as a real number; in general however it could be extremely interesting in order to augment the 
expressivity of the filters $\omega^x_\varphi$ to consider
the case in which $(d,x)\mapsto g(\sigma,d,x)$ is 
indeed a neural network and $\sigma$ are its weights (as it is shown in the side
figure).

\smash{\raise-4.5pc\rlap{\kern 11pc \includegraphics{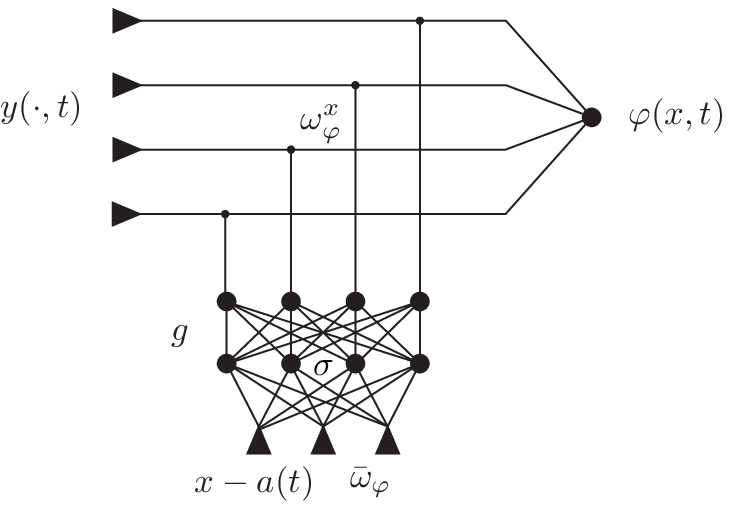}}}

\subparagraph{Circular crown FNN}
\hangindent=-13pc
\hangafter=-2
We can approximate the above computational model by defining receptive field base filters
whose shape changes depending on where we focus attention.
Let $R:=\diam\Omega$. Suppose also that we are given a collection
of $m$ filters $\omega_1,\dots,\omega_m$ each of which, for each fixed temporal
instant is a map $\omega_i(\cdot,t)\colon\Real^2\to\Real$ with compact
support which contains the origin.

Then fixed $t\in[0,T]$, and given the focus of attention
trajectory $t\mapsto a(t)$ we define for all $x\in\Omega$
\[
\omega^x(\xi,t):=\begin{cases}
\omega_1(\xi,t) & |x-a(t)|\le R/m;\\
\omega_2(\xi,t) & R/m<|x-a(t)|\le 2R/m;\\
\quad\vdots&\qquad\vdots\\
\omega_m(\xi,t) & (m-1)/m<|x-a(t)|\le R;\\
\end{cases}
\]
Notice indeed that there cannot be a $x\in\Omega$ that for some
$t\in[0,T]$ yields $|x-a(t)|> R$ by definition of diameter of $\Omega$.

Additional constraint on the form of the filters $\omega_1,\dots,\omega_m$
can be imposed to better exploit the focalization of attention
mechanism; for instance it is reasonable to think that the filter that
act closer to the point where we focus attention should be narrower and
could have rapid spatial variations in order to allow for crisper
transfer of information while the filters whose index is closer to $m$
should probably be broader and quite flat. Notice that this kind of
constraints could be implemented by imposing
constancy of the weights over their receptive fields when we move
farther from the focus of attention.

Another important remark is in order; when we choose 
$\omega_1\equiv\omega_2\equiv\dots\equiv\omega_m$
we recover the classical CNN architecture. This fact suggests 
that at the beginning of learning the initialization of 
the weights could be done uniformly
$\omega_1(\cdot,0)=\omega_2(\cdot,0)=\dots=\omega_m(\cdot,0)
=\omega^0$ at least to suggest to the network an initial 
equivariance under translation.

\subparagraph{Propagation of the visual signal through the network}
The propagation of visual information that has  been assumed
in Eq.~\eqref{eq:activations_FNN} is the typical forward 
\emph{instantaneous} propagation of NN. This basically means 
that in Eq.~\eqref{eq:activations_FNN} we are assuming 
that the speed of propagation of the signal from the 
input to the output of the neural network is much 
grater than the typical dynamic of the video signal.

In general however, in order to avoid some conceptual 
paradoxes (see \cite{betti2020backprop}),
it is much more natural to consider a propagation 
dynamic from input to output. One of the simplest 
way to model this propagation is to substitute
 Eq.~\eqref{eq:activations_FNN} with 
\begin{equation}\label{eq:activations_FNN_dynam}
\left\{
\begin{aligned}
&\upsilon_\varphi(x,t)=(\omega_\varphi^x(\cdot,t)\star \hat y(\cdot,t))(x);\cr
&\varphi_t(x,t)=
c F(\varphi(x,t),\gamma(\upsilon_\varphi(x,t))),
\end{aligned}\right.
\end{equation}
where $F\colon \bbR\times\bbR\to\bbR$ specifies the way in which the propagation takes place. 
For instance a possible choice here would be 
$F(a,b)=a-b$; in this case we would have
\[
\varphi_t(x,t)=
c[\varphi(x,t)-\sigma(\upsilon_\varphi(x,t))].
\]
This form makes also clear the role of the propagation
velocity $c$: When $c$ is big then the above equation 
formally reduces to the static propagation described in 
Eq.~\eqref{eq:activations_FNN}.

It is also important to compare this newly introduced
dynamics of the features with the computational model
that we introduced in Chapter~3, specifically in Eq.~\eqref{aggreg-eq}. This two dynamics are 
in fact rather different: The dynamics
proposed in Eq.~\eqref{aggreg-eq} was an added dynamic 
that was restricted at the level of the feature $\varphi$
and it computation still involved a classical forward 
computation in order to determine the value of $\alpha_\varphi$, while in  Eq.~\eqref{eq:activations_FNN_dynam} we are introducing a 
temporal delay in the computation of the features at a certain
level of our architecture in terms of the value of the 
features at previous layers.

\chapter{Information-based laws of feature learning}
\label{driving-principles}
\AtBeginShipoutNext{\AtBeginShipoutUpperLeft{%
  \put(1in + \hoffset + \oddsidemargin,-5pc){\makebox[0pt][l]{
\boxformat 
  \framebox{\vbox{
\hbox{Published by Springer, Cham---Cite this chapter:}
\smallskip
\hbox{\url{https://doi.org/10.1007/978-3-030-90987-1_5}}}
  }}}%
}}

\vspace{-4cm}
\begin{quote}
When I was in high school, my physics teacher - whose name was Mr. Bader - called me down one day after physics class and said, ``you look bored; I want to tell you something interesting.'' 
Then he told me something which I found absolutely fascinating, and have, 
since then, always found fascinating. Every time the subject comes up, I work on it.\\
~\\
Richard Feynman, physics lectures about the principle of least action
\end{quote}


%

\section{The simplest case of feature conjugation}
The brightness is the simplest feature which only involves a single pixel.
Hence, we are in front of the simplest case of feature conjugation\index{feature conjugation}
$b \Join v=0$, that has been discussed in Section~\ref{FromMP3Pix-sec}.
\marginpar{\small A soft-satisfaction of the $b \Join v=0$ conjugation}
The well-posedness of the problem that is gained by the constrained
minimization of the functional defined by Eq.~\ref{OpticalFlowReg}
is typically given a different formulation, which only involves a 
soft satisfaction of $b \Join v=0$. This can be formulated
by stating that, given $b$, one must minimize
\begin{equation}
	E(t) = \int_{\Omega} (b \Join v(x,t))^2 +
	 \lambda \big((\sgrad v_1(x,t))^2 + (\sgrad v_2(x,t))^2\big)\, dx,
\end{equation}
where  $\lambda>0$ is the regularization parameter.
The soft-enforcement of the conjugation constraint in this case 
makes sense. As already pointed out, this conjunction arises from 
the brightness invariance principle, which is only an approximate
description of what happens in real video. 
The stationary point of $E$, which is also a minimum, is obtained by
using the Euler-Lagrange equations
\begin{align}
\begin{split}
	\nabla^{2} v_1 =  \frac{1}{\lambda} (b \Join v) \ \partial_{x_1} b\\
	 \nabla^{2} v_2 =  \frac{1}{\lambda}  (b \Join v) \ \partial_{x_2} b,
\end{split}
\label{HS-EL-sol}
\end{align}
that are paired with appropriare boundary conditions on $v$.
%
%

\marginpar{\small The aperture problem and the barber's pole}
The regularization term~(\ref{OpticalFlowReg})
helps facing the ill-posedness of brightness invariance. 
Basically, one must distinguish the motion of material points and
their project onto the retina and the corresponding optical flow
that is interpreted as it is in an inherently ambiguous context.


While the regularization helps discovering one precise solution, 
it might not be really satisfactory in terms of the visual interpretation 
of the environment. Something similar happens in the brain.
Many motion sensitive neurons in the visual system close to 
the retina  respond locally while processing on small receptive fields. 
In higher level of the  hierarchical architecture of the visual system, 
the information from several of the neurons that respond locally
is integrated to allow us to perceive more global information.
The inherent  ambiguity in motion computation in neurons based 
on a small receptive field is called the {\em aperture problem}.
The aperture problem arises when looking at a moving image through a small hole -- the aperture. Different directions of motion can appear identical when viewed through an aperture\footnote{A very nice example is 
illustrated at \url{http://elvers.us/perception/aperture/}.}. 
In order to conquer a deeper interpretation we need to extract higher
level features along with their own conjugated velocity.



\section{Neural Network representation of the velocity field}
\label{OFbyNN-sec}\index{optical flow}
Now consider the case in which the velocity field
is estimated through a neural network $\eta_v
\colon \mathbb W_x\times C^\infty(\Omega;
\Real^{3c})\to(\mathbb V_x)^2$
where $c$ is the number of channels of the brightness.
Here $\mathbb W_x$ is the functional space that determines the
spatial regularity of the filters, while $\mathbb V_x$ is the
space in which we select the velocity field at each temporal
instant as a function of space; for the moment however we
do not want to discuss what could be appropriate choices of those spaces.
This map uses a set of filters $\omega\colon \Omega\to\Real^N$
to associate to the spatio-temporal gradient of the brightness $\tgrad b
(\cdot, t)$ evaluated at a certain temporal instant $t$ 
to the corresponding velocity field $v(\cdot,t)=
\eta_v(\omega,\tgrad b(\cdot,t))$.

Notice that at this level the ``neural network'' $\eta_v$
is a potentially very complicated map between functional spaces and thus
it is something that would be rather delicate to work with.
For this reason we abandon the very useful picture of the
theory on a continuous retina and for the remainder of this section we will
focus on a discrete description of the retina made up of discrete entities
that we will denote as pixels. However we will still retain continuity
in the temporal domain since this will allow us to describe the learning
process in terms of a continuous temporal dynamics that in our
opinion is very helpful.

So we model the retina as the set
$\Omega^\diamond:=\{(i,j)\in\mathbb N^2:1\le i\le w, 1\le j\le h\}$.
Then the brightness map can be modeled as a map $t\mapsto b(t)\in
\Real^{w\times h}$.\footnote{Here, for simplicity and for not having to deal with too many indices, we are considering images which have only one channel, like greyscale images, however if we let
$t\mapsto b(t)\in
\Real^{w\times h\times c}$ where $c$ is the number of channels of the
image at time $t$ (e.g. $c=3$ for RGB images) we recover the general case.}
We will also denote with
$t\mapsto b'(t)\in\Real^{w\times h}$ the discretization
of the temporal partial derivative $b_t(x,t)$ of the brightness.
The discrete spatial gradient operator $\Dgrad\colon \Real^{w\times h}
\to\Real^{w\times h\times 2}$ defined by its action on any
$x\in\Real^{w\times h}$
\[(\Dgrad x)_{i,j,1}=\begin{cases}
x_{i+1,j}-x_{i,j} &\hbox{if $1\le i<w$};\\
0&\hbox{otherwise,}
\end{cases},\quad
(\Dgrad x)_{i,j,2}=\begin{cases}
x_{i,j+1}-x_{i,j} &\hbox{if $1\le j<h$};\\
0&\hbox{otherwise.}
\end{cases}\]
Therefore, now that we are working in the discrete, the a network that
estimates the velocity field $f$ should be in general a function that given some
set of $N$ parameters, a spatial discrete gradient of an image, and
its temporal partial derivative computes a two dimensional vector (the optical
flow) for each pixel of the retina. More formally given the function
$f\colon[0,T]\times\Real^N\times\Real^{w\times h\times2}\times\Real^{w\times h}
\to \Real^{w\times h\times2}$ we are after the weight maps
$t\mapsto \omega(t)\in\Real^N$ such that $f(t,\omega(t),\Dgrad b(t),b'(t))$
should as times goes by be a good estimation of the optical flow.
Here the extra temporal dependence of the neural network is added since we also want to 
take into account in the computation the
value of $a(t)$, i.e. of the
position of the focus of attention\index{focus of attention}.

In a continuous  setting like the one in which we pose ourselves it is natural to
consider an online learning\index{online learning} problem for the parameters $\omega$.
A very natural way in which this problem can be formalized is as a
non homogeneous (online) flow on a suitably regularized Horn and Shunk
functional. To pose this problem define the map
$(x,s)\in[0,T]\times\Real^N\mapsto \HS(x,s)\in\Real_+$ where
\begin{equation}\label{eq:HS-potential-def}
\begin{aligned}
\HS(x,s)\!:=&
\frac{1}{2}\sum_{\substack{
1\le i\le w\\
1\le j\le h\\
}}
\left(
((\Dgrad b(s))_{i,j,1},
(\Dgrad b(s))_{i,j,2},b'(s))\cdot\begin{pmatrix}f_{i,j,1}(s,x,\Dgrad b(s), b'(s))\\
f_{i,j,2}(s,x,\Dgrad b(s), b'(s))\\
\\1\end{pmatrix}\right)^2\\
&\, +
\frac{\lambda}{2}\sum_{\substack{
1\le i\le w\\
1\le j\le h\\
}}
\Bigl(\bigl(\Dgrad f_{i,j,1}(s,x,\Dgrad b(s),
b'(s))\bigr)^2
+\bigl(\Dgrad f_{i,j,2}(s,x,\Dgrad b(s),
b'(s))\bigr)^2\Bigr),
\end{aligned}
\end{equation}
is the explicit version of the Horn and Shunk functional regarded as a function
of time (that as we will will take the place of the parameter $s$) and the weights of
the network (that will occupy the position taken by $x$).
We will also denote with $\nabla \HS(x,s)\equiv \HS_x(x,s)$ the gradient
with respect to the first argument of $\HS$ (i.e. the parameters of the
network) and with $\HS_t(x,s)$ the partial derivative with respect to the second
argument (i.e. partial temporal derivative).

We want to define the learning dynamic as the outcome of a variational
principle (known as the WIE principle~\cite{liero2013new}) on the functional
\begin{equation}\label{eq:WIE-fun-HS}
F_\eps(\omega):=\int_0^T
e^{-t/\eps}\left(
\frac{\nu\eps}{2}|\omega'(t)|^2+\HS(\omega(t),t)
\right)\, dt
\end{equation}
defined over the set $\bbX=\{\omega\in H^1([0,T];\bbR^N):
\omega(0)=\omega^0\}$. The wanted learning trajectory then it is 
obtained (see~\cite{liero2013new} and~\cite{phdthesis}) 
as follows:
\begin{enumerate}
\item For each fixed $\eps$ find a minimizer $\omega_\eps$
of~\eqref{eq:WIE-fun-HS};
\item Take the limit $\eps\to0$ and consider the limit trajectory
(in a suitable topology).
\end{enumerate}
If we assume that $\omega$ is smooth\footnote{This can actually be proven assuming regularity on $\HS(\cdot,t)$; in particular if $\HS(\cdot,t)\in\C^\infty(\R^N)$, then also $\omega\in C^\infty([0,T])$.} using Eq.~\eqref{ELE-appendix} and
Eq.~\eqref{eq:stat-cond-bound} we can write the 
stationarity condition for $F_\eps$ as
\begin{equation}
\begin{cases}
-\eps\nu\ddot\omega_\eps(t)+\nu\dot\omega(t)+\nabla\HS(\omega(t),t)=0&t\in(0,T)\\
\omega_\eps(0)=\omega^0,\quad\eps\nu\omega_\eps(T)=0.
\end{cases}
\end{equation}
Then taking the formal limit as $\eps\to 0$ yields the following
dynamics:
\begin{equation}\label{eq:HS-net-non-hom-flow}
\begin{cases}
\omega'(t)=-\frac{1}{\nu}\nabla\HS(\omega(t),t)&\\
\omega(0)=\omega^0.
\end{cases}
\end{equation}
Notice that in order to compute $\HS$ we should at some point compute
the gradient of the network with respect to its parameters; this can be
done algorithmically using the backprop.

Now notice that Eq.~\eqref{eq:HS-net-non-hom-flow}
represents a non-homogeneous flow on the ``potential''
$\HS$, the non-homogeneity comes from the fact that since we are dealing
with a video stream the gradients of the brightness changes all the time.
The goal of the learning is to reach a set of constant weights that
properly estimate the optical flow given any couple of frames in the video.
We argue however that the learning process itself could benefit from a
developmental stage in which the task is simplified by means of a
spatial filtering of the input.

In order to do this we consider the following ``smoothing'' process
of the input achieved for instance by a gaussian-like  filtering; given a
frame $x\in\Real^{w\times h}$ and a parameter $\sigma\ge0$ we consider
\[
x\mapsto \Phi(x,\sigma)\in\Real^{w\times h}.
\]
We assume that this filtering should give back the frame for $\sigma=0$,
i.e. $\Phi(x,0)=x$ and that for $\sigma>0$ should satisfy
$\Vert \Dgrad\Phi(x,\sigma)\Vert\le\Vert \Dgrad x\Vert$.

Now the idea is that the degree of filtering that is expressed by $\sigma$
could be learned alongside with the weights of the network. Of course in order
to avoid trivial solutions we should also require that this smoothing
should become as small as possible.

With this filtering of the input our $\HS$ potential defined
 in Eq.~\eqref{eq:HS-potential-def}
 should take into account an additional dependence. This can be readily
 done by considering $\HSs\colon[0,T]\times\Real^N\times\Real\to
 \Real_+$ defined by\footnote{Here we use Einstein convention and
 we do not explicitly write the two components of $f$ and $\Dgrad\Phi$
 for compactness.}
\begin{equation}
\begin{aligned}
\HSs(s,x,\sigma):=&
\frac{1}{2}
\left(
\bigl((\Dgrad \Phi(b(s),\sigma))_{i,j},\Phi'(b(s),\sigma)\bigr)\cdot\begin{pmatrix}
f_{i,j}(s,x,\Dgrad \Phi(b(s),\sigma), \Phi'(b(s),\sigma))\\
\\1\end{pmatrix}\right)^2\\
&\quad +
\frac{\lambda}{2}
|\Dgrad f_{i,j}(s,x,\Dgrad \Phi(b(s),\sigma),\Phi'(b(s),\sigma))|^2,
\end{aligned}
\end{equation}
where $\Phi'(b(s),\sigma)$ is the analogue of $b'(s)$ after the $\Phi$
transformation is applied.

Correspondingly the functional in Eq.~\eqref{eq:WIE-fun-HS} becomes
\[\begin{aligned}
\overline F_\eps(x,\sigma):=
\int_0^T
e^{-t/\eps}\Bigl(&
\frac{\nu_\omega\eps}{2}|\omega'(t)|^2
+\frac{\nu_\sigma\eps}{2}(\sigma'(t))^2\\
&\quad+\HSs(t,\omega(t),\sigma(t))
+\frac{k}{2}(\sigma(t))^2
\Bigr)\, dt.
\end{aligned}\]
Taking the same exact steps that led to Eq.~\eqref{eq:HS-net-non-hom-flow} with the additional variable $\sigma$ we get
\begin{equation}
\begin{cases}
\omega'(t)=-\frac{1}{\nu_\omega}\nabla\HSs(t,\omega(t),\sigma(t))&\\
\sigma'(t)=-k\sigma(t)-\frac{1}{\nu_\sigma}\HSs_\sigma(t,\omega(t),\sigma(t))&\\
\omega(0)=\omega^0,\quad \sigma(0)=\sigma^0.
\end{cases}
\end{equation}
Here $\sigma^0>0$ is a parameter chosen in such a way that $\Phi(b(0),\sigma^0)$
is a constant (or constant up to a certain tolerance) frame.

\section{A dynamic model for conjugate features and velocities}
In the previous section we described an online method to extract the
optical flow from the brightness.

We now turn to the definition of an online method for the
learning of the features and of their conjugates velocities
as it is described in Chapter~3. In order to do this we need to
lay down some useful notation. As we did in the previous section
we will still work on the discrete retina $\Omega^\diamond$.

The first ingredient that we need to put in the theory, as it
is widely discussed in Chapter~3 is a set of features that are
instrumental to the identification of objects. Here we consider
a set of maps $P^1,\dots,P^m$ such that
$P^k\colon\Real^{w\times h\times m_{k-1}}\to\Real^{w\times h\times m_k}$
with $m_0=1$ (or $m_0=3$ if we consider RGB images).
Examples of these transformation could be convolution or, better yet,
the foveated layer transformation described in the previous\index{foveated neural networks}
chapter. In any case the assumption is that each of these maps can be
regarded as parametric maps $\Pi^k\colon
\Real^{N_k}\times\Real^{w\times h\times m_{k-1}}\to\Real^{w\times h\times m_k}$
so that $\Pi^k(\omega^k,\cdot)\equiv P^k(\cdot)$ for some $\omega^k\in
\Real^{N_k}$, $k=1,\dots,n$. Once the dynamics of the set of weights
$t\mapsto \omega^k(t)$ has been 
fixed he resulting maps obtained from the composition of the $\Pi^k$
are exactly the feature maps on which we based our discussion in the
previous chapters:
\[
\varphi^i(t):=\biggl(\circop_{k=1}^i \Pi^k(\omega^k(t), \cdot)\biggl)(b(t))
\]
where $\circ$ means function composition and  
where as usual here with $\varphi^i$ we are thinking
about the discretized version of the feature maps $\varphi^i$
defined on the cylinder $\Gamma$. Consistently as we have done in
the previous section however we still retain the continuous temporal
dependence.

However since these are the building blocks of our theory, and we need to
prescribe a dynamics for the weights, it is also convenient to explicitly
consider the dependence of the features on the weights.
To do this we instead define the set of maps $A^i\colon\prod_{k=1}^n\Real^{N_k}
\times [0,T]\to \Real^{w\times h\times m_i}$ defined as
\[
A^i(\omega,t):= \biggl(\circop_{k=1}^i \Pi^k(\omega^k, \cdot)\biggl)(b(t))
\]
where in this case $\omega^k$ is thought as the $k$-th component of
$\omega$ in the space $\prod_{k=1}^n\Real^{N_k}$ (i.e.
$\omega=(\omega^1,\omega^2,\dots,\omega^n)$).
It is important to notice that at this stage we can think of the features
$\varphi^i$ either as scalar features as it has been done till now
(in that case $m_k\equiv1$ for all $k=1,\dots,n$) or as vector which
codes a single feature (like it happens for RGB signals).

The same exact construction can be done for the affordance features\index{affordance}
$\psi^i$ and $\chi^i$, and yields the map  $B$ for the
$\psi$ features and the $C$ map for the $\chi$ features.

Now the learning process is therefore described by the
dynamics of the weights $t\mapsto\omega_A(t)$, $t\mapsto\omega_B(t)$
and $t\mapsto\omega_C(t)$. As a note on notation we will
denote with  $\nabla A^i\colon \prod_{k=1}^n\Real^{N_k}
\times [0,T]\to \Real^{w\times h\times m_i}\times \prod_{k=1}^n\Real^{N_k}$
the gradient of $A^i$ with respect to its first argument (the parameters
of the network $\omega$) and with $A^i_t\colon
\prod_{k=1}^n\Real^{N_k}
\times [0,T]\to \Real^{w\times h\times m_i}$ the partial derivative with respect
to its second argument (i.e. the partial derivative with respect ti time).
On the other
hand, consistently with what we did in the previous section we
assume that the spatial gradient operator $\Dgrad$ acts separately on each
of the $m_i$ components of the feature so that
for instance, for each admissible $\omega$, $t$ and $i$ we have
$\Dgrad A^i(\omega,t)\in\Real^{w\times h\times m_i\times 2}$.
Similar definitions will be used for the maps $B^i$ and $C^i$.

Moreover for each feature $\varphi$ we need to model, again with a neural
network, the corresponding optical flow. As we have previously discussed
the input of this network will be the gradients of the brightness,
i.e. $\Dgrad b$ and $b_t$. We will denote the parameters of
the network that estimate the velocity as $\vartheta^k\in\Real^{M_k}$,
$v^i\colon\prod_{k=1}^n \Real^{M_k}\times[0,T]\to\Real^{w\times h\times 2}$.
As it happened for the  features here we compactly express the
dependence of $v^i$ on the whole set of parameters of the network
that estimate the velocities; again the learning process
should describe the trajectory $t\mapsto \vartheta(t)$.

The last ingredient that we need to lay down the learning problem is
the ``'decoding' part of the architecture. As we described in Chapter~3
a natural regularization for the features is the reconstruction
(or better yet the time shifted reconstruction); such
map should take as input the features $\varphi^i(t)$ and decode them into
the shifted frame $b(t+\tau)$\footnote{Here we are using the shifted brightness $b(t+\tau)$ instead
of the temporal derivative as indicated in 
Chapter~3 because, while clearly related
for suitable choices of $\tau$, this solution
seems to be more straightforward to implement.}.
So we define
the map $\Delta\colon \Real^d\times \prod_{k=1}^n\Real^{w\times h\times m_k}
\to\Real^{w\times h}$, where $d$ is the number of the parameters of the
decoder. In this case the dynamic of learnable parameters
$t\mapsto \omega_\Delta(t)$ should be guided by constraint
\[
\Delta(\omega_\Delta(t),A^1(\omega_A(t),t),\dots,A^n(\omega_A(t),t))=
b(t+\tau) \quad t\in[0,T].
\]
In what follows we will simply write $\Delta(\omega_\Delta(t),A(\omega_A,t))$
in place of the explicit term
$\Delta(\omega_\Delta(t),A^1(\omega_A,t),\dots,A^n(\omega_A,t))$.
The incorporation of such constraint is assumed to be done in a soft
way using for example a quadratic loss $q\colon \Real^{w\times h}\times
\Real^{w\times h}\to\overline\Real_+$,
with $q(a,b):=(a_{i,j} -b_{i,j})(a_{i,j} -b_{i,j})/2$.

The other constraint that we dealt with in the previous section,
i.e the brightness invariance, here is replaced by
$\bowtie$ relation between the features $\varphi^i$ and their
corresponding velocities $v^i$. Then the conjunction relation
between the $\varphi^i$ and the $v^i$ in these new variable can be
rewritten as
\[\frac{d}{dt}A^i(\omega_A(t),t)+ v^i(\vartheta(t),t)\cdot
\Dgrad A^i(\omega_A(t),t).
\]
Please notice here that the total derivative of $A^i$ replaces the
partial derivative of $\varphi^i(x,t)$ since $A$ explicitly depends on time
both through the input but also through the temporal dynamics of the
weights. This condition can be enforced by considering the
functional $\mathcal{T}(\omega_A,\vartheta)$ where
\footnote{With the usual conventions for
the Einstein summation, and where $|\cdot|$ indicate that we are summing
also on the spatial components of $A$.}
\[(x_1,z)\mapsto \mathcal{T}^A(x_1,z):=
\int_0^T
\Bigl\vert\frac{d}{dt}A^i(x_1(t),t)+ v^i(z(t),t)\cdot \Dgrad A^i(x_1(t),t)\Bigr
\vert^2\,
dt.
\]
A similar term can be used to impose the constraint on the $C$ field,
a possible choice, as it is displayed in Eq.~\eqref{eq:chi-conj} would be in that
case $\mathcal{T}^C(\omega_C,\vartheta)$, where 
\[(x_2,z)\mapsto \mathcal{T}^C(x_2,z):=
\int_0^T
\Bigl\vert\frac{d}{dt}C^j(x_2(t),t)+ v^i(z(t),t)\cdot \Dgrad C^j(x_2(t),t)\Bigr
\vert^2\,
dt
\]
where it is understood that the sum is performed over $i\ne j$.
Finally the $B$ field induce a term $\mathcal{T}^C(\omega_C,\vartheta)$
with 
\[
(x_3,z)\mapsto \mathcal{T}^B(x_3,z):=
\int_0^T
\Bigl\vert\frac{d}{dt}B^j(x_3(t),t)+ (v^i(z(t),t)
-v^j(z(t),t))\cdot \Dgrad B^j(x_3(t),t)\Bigr
\vert^2\,
dt,
\]
where again the summation is performed over $i\ne j$.

The discussion that has been carried on in Chapter~3 shows that beside
this defining constraint, in order to have a non-trivial and well defined
theory we also need regularization terms. First of all we need to make
sure that the dynamics of the unknowns $\omega_A$, $\omega_B$, $\omega_C$
and $\vartheta$ is well behaved; to this end we introduce the
term $\mathcal{S}(\omega_A,\omega_B,\omega_C,\omega_\Delta,\vartheta)$ where
\[
(x_1,x_2,x_3,y,z)\mapsto \mathcal{S}(x_1,x_2,x_3,y,z):=
\frac{1}{2}\int_0^T 
\vert x_k''\vert^2 + \vert x_k'\vert^2+ \vert y'\vert^2+\vert z'\vert^2.
\]
Notice that here we impose an higher regularity on the
variables $\omega_A$, $\omega_B$ and $\omega_C$ essentially because
their derivatives already appears in the terms that enforce the
conjugation between $A$, $B$ and $C$ and their velocities.

The other natural regularization that we need to impose is the spatial
regularization of $A$, $B$ and $C$ and to the velocity fields.
To this end we consider
the term $\mathcal{W}(\omega_A,\omega_B,\omega_C,\vartheta)$ where
$(x_1,x_2,x_3,z)\mapsto \mathcal{W}(x_1,x_2,x_3,z)$ is chosen to be given by
\[\begin{aligned}
\mathcal{W}(x_1,x_2,x_3,z):=
\frac{1}{2}\int_0^T\Bigl(&
|\Dgrad A^i(x_1(t),t)|^2 +|\Dgrad B^i(x_2(t),t)|^2\\
&+|\Dgrad C^j(x_3(t),t)|^2+|\Dgrad v^i(z(t),t)|^2\Bigr)\, dt.
\end{aligned}\]
We are also adding a term $\mathcal V(\omega_A,\omega_B,\omega_C)$ that controls the growth of the features by
simply weighting the
value of $A$, $B$ and $C$:
\[
(x_1,x_2,x_3)\mapsto \mathcal V(x_1,x_2,x_3):=
\frac{1}{2}\int_0^T\Bigl(|A^i(x_1(t),t)|^2 +|B^i(x_1(t),t)|^2
+|C^j(x_1(t),t)|^2
\Bigr)\, dt
\]
Finally we need the auto-encoding-like regularization and the
terms that propagates this kind of regularization from the $A$ to
the $B$ and $C$ field as well as
a cohercive term on the parameters $\vartheta$.
Here it is convenient for us to group these
terms into the functional $\mathcal I(\omega_A,\omega_B,\omega_C,
\omega_\Delta,\vartheta)$.
Here the functional $(x_1,x_2,x_3,x_4,z)\mapsto
\mathcal{I}(x_1,x_2,x_3,x_4,z)$ is defined as follows
\[
\begin{aligned}
\mathcal{I}(x_1,x_2,x_3,x_4,z):=
\int_0^T\biggl(& \frac{|z(t)|^2}{2} + \frac{|x_4(t)|^2}{2}+
q\bigl(\Delta(\omega_\Delta(t),
A(x_1(t),t)), b(t+\tau)\bigr)\\
&\quad+\Theta(A(x_1(t),t),
B(x_2(t),t))\\
&\qquad +\Upsilon(B(x_2(t),t),
C(x_3(t),t))\biggr)\, dt,
\end{aligned}
\]
where the terms $\Theta\colon\prod_{k=1}^m \Real^{w\times h\times n_k}
\times \prod_{k=1}^n \Real^{w\times h\times m_k}\to\Real$
and $\Upsilon\colon\prod_{k=1}^n \Real^{w\times h\times m_k}
\times \prod_{k=1}^p \Real^{w\times h\times r_k}\to\Real$
enforces the
regularization presented in Eq.~\eqref{FromLogImplicfi} and Eq.~\eqref{eq:logic-reg-chi} respectively.

With all this definitions lay down we can now state that the trajectory
of the weights $t\mapsto (\omega_A(t),\omega_B(t),\omega_C(t),\omega_\Delta(t),
\vartheta(t))$ compatible with a learning process from a visual stream
should be construct so that it optimizes the index
\begin{equation}\label{eq:index_first_form}
\begin{aligned}
\mathcal{S}(\omega_A, \omega_B,\omega_C,\omega_\Delta, \vartheta)
&+\mathcal{W}(\omega_A,\omega_B,\omega_C)+
\mathcal V(\omega_A,\omega_B,\omega_C)\\
&+\mathcal I(\omega_A,\omega_B,\omega_C,
\omega_\Delta,\vartheta)
+\sum_{\alpha\in\{A,B,C\}}
\mathcal{T}^\alpha(\omega_\alpha,\vartheta).
\end{aligned}\end{equation}
The previous statement however is vague enough since we have not
specified the functional space of trajectories over which we desire to
define the optimization problem. As soon as we try to define an appropriate
space, we quickly realize that non-causality issues arise from the boundary
conditions.

In order to overcome this problem in the next section  we
will discuss how to apply a variation of the WIE principle that that we used
in Section~5.2 for the
extraction of the optical flow to this problem. This approach 
as we will see will yield
a gradient flow-like dynamics (thus a causal problem). In order to 
help the intuition on this approach we will also give 
a corresponding interpretation in the discrete time.

However, before going on let us rearrange the terms that appears in
Eq.~\eqref{eq:index_first_form} so that subsequent analysis will be more
straightforward. The basic idea is that we want to group together,
inside the integral on time,
terms that depends on the variables $\omega_A$, $\omega_B$, $\omega_C$,
$\vartheta$ and $\omega_\Delta$, the terms that depends only on the
derivatives of those variables (which is essentially already
the $\mathcal S$ term) and the terms that  contains both the variables
and their derivatives (the terms that comes from the $\Join$ operation).
Let $t\mapsto u(t):=(\omega_A(t),\omega_B(t),\omega_C(t),\omega_\Delta(t),
\vartheta(t))\in\Real^D$, 
then the index in Eq.~\eqref{eq:index_first_form} is of the
form
\begin{equation}\label{eq:index_compact_form}
F(u):=\int_0^T \Bigl(T(u'(t), u''(t))+ \frac{1}{2}|Q(u(t)) u'(t)+b(u(t))|^2
+ V(u(t),t)\Bigr)\, dt,
\end{equation}
where $V$ collects all the terms of $\mathcal V$, $\mathcal W$ and $\mathcal I$,
$Q$ is a (non square) block matrix and together with $b$
defines the invariance term. Finally
$T$ is the rewriting of the term $\mathcal S$. It is also useful to
rewrite the $\bowtie$ term as follows:
\[
\frac{1}{2}|Q(u(t)) u'(t)+b(u(t))|^2=\frac{1}{2}u'(t)\cdot Q(u(t))Q'(u(t))
u'(t) +b(u(t))\cdot Q(u(t))u'(t) +\frac{1}{2} |b(u(t))|^2,
\]
which in turn can be rearranged as follows
\[
\frac{1}{2}|Q(u(t)) u'(t)+b(u(t))|^2=
\frac{1}{2}u'(t)\cdot M(u(t))
u'(t) +m(u(t))\cdot u'(t) +\kappa(u(t)),
\]
where this time $\forall x\in\Real^D$, $M(x)$ is a square symmetric,
positive semi-definite 
matrix, $m(x)\in\Real^D$ and $\kappa(x)\in\Real$. This last form is the one
that we will use in the next section to write the online learning rules.

\section{Online learning}
\label{Online-grad-sec}\index{online learning}
The first step that we need to take in order to obtain causal
rules is to appropriately rescale the functional $F$ defined in
Eq.~\eqref{eq:index_compact_form} with the $\eps$ factor as we did in
Section~5.2 for the extraction of the optical flow; in this case we
consider the following family of functionals:
\[\begin{aligned}
F_\eps(u):=\int_0^T e^{-t/\eps}\Bigl(\frac{\eps\nu}{2}|u'(t)|^2+
\frac{\eps\gamma}{2}u'(t)\cdot M(u(t))
u'(t) &+\eps \gamma m(u(t))\cdot u'(t)\\
&+\eps\gamma\kappa(u(t))
+ V(u(t),t)\Bigr)\, dt
\end{aligned}
\]
where we have chosen $T(a,b)=|a|^2/2$ and we have weighted both the
kinetic term and the constraint between the features and the velocities by $\eps$; we have also introduced weights $\nu>0$, $\gamma>0$.
Also in this case the variational problem can be set in the functional
space $\bbX$ defined as in~5.2 with the only difference that now 
the co-domain of the elements of $\bbX$ is $\bbR^D$ instead of $\bbR^N$.
In order to get local and causal laws the plan is the same that we discussed for the extraction of the optical flow from the brightness:
We write for each fixed $\eps$ the Euler Lagrange equation for $F_\eps$ and then we let $\eps\to0$.
Using Eq.~\eqref{ELE-appendix} and \eqref{eq:stat-cond-bound} we obtain the following condition for the minimizer $u_\eps$:
\begin{equation}
\begin{cases}
\begin{split}
&-\frac{d}{dt}\left(e^{-t\eps}\eps(\nu\Id+\gamma M(u_\eps(t))u'(t)
+e^{-t/\eps}\eps\gamma m(u_\eps(t)))\right)\\
&\quad+\eps e^{-t/\eps}\left(
\frac{1}{2}u_\eps'(t)\cdot M'(u_\eps(t))u_\eps'(t)+
\gamma m'(u_\eps(t))\cdot u_\eps'(t)+\kappa'(u_\eps(t))\right)\\
&\qquad+e^{-t/\eps}\nabla V(u_\eps(t),t)=0
\end{split}&t\in(0,T)\\
u_\eps(0)=u^0,\quad \eps\bigl[\bigl(\nu\Id+\gamma M(u_\eps(T))\bigr)u_\eps'(T)+\gamma m(u_\eps(T))\bigr]=0
\end{cases}
\end{equation}
As $\eps\to0$ we get the only surviving terms of the Euler
equation, i.e. the potential term and those which come from the differentiation of the exponential term, are
\[
\begin{cases}
\bigl(\nu \Id+\gamma M(u(t))\bigr) u'(t)+ \gamma m(u(t)) +\nabla V(u(t),t)=0,&
t\in(0,T);\\
u(0)=u^0.&\end{cases}
\]
Notice that since $M(x)$ is a positive semi-definite matrix for all
$x\in\Real^D$, $\nu\Id+\gamma M(x)$ is positive definite and hence invertible.
This means that our online update rules can be cast into the form
\begin{equation}\label{eq:gradient-flow}
\begin{cases}
u'(t)=-\bigl(\nu \Id+\gamma M(u(t))\bigr)^{-1}\bigl(\gamma m(u(t)) +\nabla V(u(t),t)\bigr),&
t\in(0,T);\\
u(0)=u^0.&\end{cases}
\end{equation}
A rather direct interpretation of the above equations can be obtained
by considering a discrete time setting and using the idea of Minimizing
Movements (see~\cite{ambrosio2006ennio}
and~\cite{gobbino1999minimizing})
with a special similarity term that contains the $\bowtie$
operation.
Let $(t_k)$ be a sequence of temporal
instants with $t_{k+1}-t_k=\tau>0$ and consider the following scheme
for computing the set of parameters of the model at the next step
\begin{equation}\label{eq:MMI}
\begin{aligned}
u^{k+1}=\argmin_{u} \Bigl(&V_k(u^k)+\nabla V_k(u^k)\cdot(u-u^k)\\
&\quad+\frac{1}{2\tau^2}(u-u^k)\cdot(\nu \Id +\gamma M(u^k))(u-u^k)\\
&\qquad +\frac{\gamma}{\tau}m(u^k)\cdot(u-u^k)+\gamma\kappa(u^k)\Bigr).
\end{aligned}\end{equation}
What we are actually doing here in order to compute the next step
is to optimize a second order approximation of $V$ (here we denote with
$V_k(x):=V(x,t_k)$, for all $x\in\Real^D$) and a discrete approximation
of the $\bowtie$ term. Equation~\eqref{eq:MMI} implies, just by imposing 
the stationarity condition of the term inside the $\argmin$,
\[
u^{k+1}=u^k-\tau^2(\nu \Id +\gamma M(u^k))^{-1}
\Bigr(\frac{\gamma}{\tau} m(u^k)+\nabla V_k(u^k)\Bigl),
\]
which can indeed be regarded as a discrete approximation
of~\eqref{eq:gradient-flow}.

In Section~5.2 we showed how a learnable smoothing signal can be 
introduced to relieve the load of information coming from the 
environment. Indeed the very same idea can be used here.

\section{Online learning: an Optimal-Control-Theory prospective}
\label{OC-sec}\index{online learning}
In Section~5.3 and Section~5.4 we framed the theory 
described in Chapter~3 and Chapter~4 under the assumption
that the model used to compute the feature fields and 
their corresponding velocities is a ``static'' FNN 
whose computational structure is described by 
Eq.~\eqref{eq:activations_FNN}. At the end of Section~4.3
we commented on the fact that we could introduce a 
more realistic model in which a propagation of visual 
information along the network is not istantaneous. 
In what follows we will only take into account the development
of the features $\allphi$ and their corresponding
features and in accordance with the notation 
introduced in Section 5.3 and in Chapter~3 
we will denote with 
$t\mapsto \phi^i(t)\in\bbR^{w\times h\times m_i}$
the discretized feature trajectories
(as usual we will denote with $\allphi$ the 
map $t\mapsto (\phi^1(t),\dots, \phi^n(t))$)
and with 
$t\mapsto v^i(t)\in\bbR^{w\times h\times 2}$
the corresponding \emph{discretized} velocity field.
Then the computational model described in 
Eq.~\eqref{eq:activations_FNN_dynam} could be more 
generally rewritten as
\begin{equation}\label{eq:dyn-model-phi}
\dot\allphi(t)=f(\allphi(t), u(t),a(t),t)
\end{equation}
where, consistently with the 
notation introduced in Section~5.3,
$u(t)$ is the set of parameters of the network,
$a(t)$ is the focus of attention\index{focus of attention}, which is a crucial
information to have in order to select the 
parameters in $u$ for the FNN
computation. The explicit dependence on $t$
takes into account the dependence on the visual 
signal (i.e. the brightness $b$). The map $f$
is related to the function $F$ of Eq.~\eqref{eq:activations_FNN_dynam} but of course 
acts on the newly arranged and discretized variables.

Correspondingly for the velocities we could write a 
similar model:
\begin{equation}\label{eq:dyn-model-v}
\dot\allv(t)=g(\allv(t), u(t),a(t),t).
\end{equation}
Here the function $g$ is the counterpart of the 
function $f$ in Eq.~\eqref{eq:dyn-model-phi} 
and it
aggregates the information at a certain level of the 
network that estimate the conjugate velocities 
to compute the field at a higher level.

Equations~\eqref{eq:dyn-model-phi} and~\eqref{eq:dyn-model-v} basically tells you
how to compute the features and the corresponding
velocities once the parameters $u$ are given, and 
the learning process is indeed the process of 
optimal selection of these parameters. 
The meaning of ``optimal'' as we saw in 
Section~5.3 can be specified in terms of the 
value of \emph{running cost} $r$ at time $t$
which depends of course on the value of 
$\allphi$, $\allv$ and of $u$ and that therefore
can be thought as a function 
$r\colon\bbR^{w\times h\times \sum_i m_i}
\times \bbR^{w\times h\times 2\times n}\times\bbR^D
\times [0,T]\to\bbR$ which specify the cost
$r(\allphi(t),\allv(t),u(t),t)$, through which we would like 
to formulate the learning problem as the minimum problem:
\[\inf_{u\in\mathbb{X}}
\int_0^T r(\allphi(t),\allv(t),u(t),t)
\,dt.\]
Such cost is the rewriting in this new variables of
 the potential term $V$ plus the invariance term on 
 the features. It is worth noticing that, as we
 already remarked in Chapter~3, the invariance term 
 $\dot \phi^i+v^i\cdot\Dgrad\phi^i=0$ in virtue of 
 Eq.~\eqref{eq:dyn-model-phi} could be enforced 
 by a penalty term
 \[
 \sum_{i=1}^n(f^i(\allphi,u(t),t)+v^i(t)\cdot \Dgrad\phi^i(t))^2.
 \]
Stated in this terms the learning problem can indeed be regarded as an optimal control problem 
(see~\cite{evans2010partial}) in which 
\begin{itemize}
\item $\allphi$ and $\allv$ represent 
what is usually called the \emph{state}
of the system;
\item the learnable parameters $u$ are 
the control parameters that should be chosen
compatibly with the state dynamic and 
with the cost $r$.
\end{itemize}
It is clear then that this point of view opens 
new and broad prospective to the study of the 
online learning process. Amongst them a 
particularly promising direction is that of addressing
the optimal-control-like problem described above 
using the methods of \emph{dynamic programming}
that lead to the study of Hamilton-Jacobi equations.

Finally we believe that is important to notice
that, in fact, also the focus of attention 
could in principle be considered a sort of 
``control parameter'' and could be incorporated in 
the vectorial variable $u$. In this case, as it is
described in Section~2.5 the equations that determines
the trajectory $t\mapsto a(t)$ will be coupled with 
those that perform feature extraction.

\section{Why  is baby vision  blurred?}
\label{BlurringSection}
There are surprising results from developmental psychology 
on what  newborns see. Basically, their visual acuity grows gradually 
in early months of life. Interestingly,  Charles Darwin had already noticed
this very interesting phenomenon. In his own words:
\begin{quote}
	{\small It was surprising how slowly he acquired the power of following with his eyes 
	an object if swinging at all rapidly; for he could not do this well when seven 
	and a half months old.}
\end{quote} 
At the end of the seventies, this early remark was given a technically sound
basis (see e.g.~\cite{DobsonVR-1978}).   In that paper, three techniques, --- optokinetic 
nystagmus (OKN), preferential looking (PL), and the visually evoked potential (VEP)--- were used to 
assess visual acuity in infants between birth and six months of age.
More recently,~(\cite{BraddickVR-2011}) provides 
an in-depth discussion on the state of the art in the field. It is clearly stated that 
for  newborns to gain adult visual acuity, depending on the specific visual test, 
several months are required. Then we are back to {\bf Q9} of Sec.~\ref{10Q-sec} on 
whether such an important cognitive process is  a biological issue or  if  it comes
from higher information-based laws of vision. 
Overall, the blurring process  pops up the discussion on the protection of the learning agent from
information overloading, but we want to deepen this topic 
at the light of the Vision Field Theory of this book. 

\marginpar{\small Online learning on video is mostly unexplored!}
While the enforcement of Vision Field Theory constraints\index{vision fields}
onto foveated neural networks\index{foveated neural networks} results into a clean variational problem, its
numerical solution might be hard. The solution herein presented in Sec.~\ref{Online-grad-sec}
very well resembles classic machine learning\index{machine learning} approaches based on gradient descent. 
However, it's worth mentioning that we are in front of an on-line gradient computation
that shouldn't be confused with stochastic gradient, for which there is a lot of 
experimental evidence on its behavior as well as a number of significant theoretical 
statements on its convergence. Unfortunately,
when working on-line the specter of forgetting behavior is always lurking!
The learning agent might adapt his weights significantly when looking at a still image
for days, thus forgetting what he learned earlier. The formulation given in Sec.~\ref{OC-sec}
considers the actual minimization of the functional but, as yet, its effective solution is 
an open problem. Hence, the ``solution'' that biology has discovered for children deserves
investigation.

\marginpar{\small Blurring the video for optical flow and beyond}
In Sec.~\ref{OFbyNN-sec} we have introduced the basic idea behind video blurring in
the case of prediction of the optical flow. While this is a dramatically simplified version
of the overall theory, the problem already presents the seed for the formulation of 
an appropriate solution in the general case. The input undergo a smoothing process
the purpose of which is to ``simplify life'' at the beginning of the agent's life.
In general, if we remove the information from the source by massive low pass filtering
the problem of learning has a trivial solution. The essence of the idea is that
of beginning from such a solution and then learn by tracking the constraints by 
properly controlling the smoothing parameter ($\sigma$ in  Sec.~\ref{VRes-sec})
that is expected to converge to zero at the end of learning. In general one 
can rely on additional variables to prevent from information overloading with the 
purpose of stabilizing the dynamical system, which tracks the injection of information. 
The development of any computation
model that adheres to this view is based on modifying the connections along with an
appropriate input filtering so as the learning agent always operates at an 
equilibrium point~(\cite{Betti2021-VB}).

\marginpar{\small Blurring and foveated nets}
When looking at the foveated networks, particularly to filter $g$, one can promptly 
recognize a natural access for smoothing the signal, since we can play with the
control variable $\sigma$ of Eq.~(\ref{eq:filtersfromtemplate}). The role of blurring
is likely connected with the degree of quality of the expected visual skills. 
The higher the quality the higher such a process is likely required. The effect of 
smoothing clearly simplifies stability issues of the learning equations that 
are more serious when the vision field graph become more and more complex.
While as already pointed out, chicks in the first days of life find it more difficult 
to see under quick movements, children do require much more time for 
conquering adult's visual skills. At the birth, chicks and children rely on the 
genetic transmission of information for their visual skills. When considering 
space constraints, children likely use a lot of that space for coding the learning mechanism 
with respect to chicks.

\marginpar{\small Feynman's discovery of Bessel's equation in capacitors and blurring processes}
In Sec.~\ref{WhyVFT-sec} we investigated the nature of vision fields and their relationship
with classic field theory with specific reference to electromagnetism. 
While the mutual refinement of the $E-B$ fields holds at any frequency, as we set 
a threshold for neglecting signals below its choice, we clearly see that 
the number of refinements (number of terms of the Bessel's function expansion) 
is strongly dependent on the chosen frequency. Notice that, because of the superposition
principle,  the signal can also be
carried out by supplying an input with rich spectral structure to the capacitor. 
Such a signal contains information that is related to such a composite structure and
it has an interesting analogy with the video signal. Now, we notice that one could
elaborate the described Feynman's analysis by assuming that the signal at the capacitor
begins from DC and gradually moves to its final spectral structure. The  
development of Bessel's term from the $E-B$ interaction is dual to the development 
of the vision field interaction, which suggests that the gradual exposition to the
information might simplify the learning process.\index{vision fields}


%
\begin{svgraybox}
	When promoting the role of time, the arising pre-algorithmic framework suggests
	extending the learning process to an appropriate modification of the input, that is
	finalized to achieve the  expected ``visual acuity'' at the end of the process of 
	learning. This likely stabilizes the learning process especially in complex
	foveated animals. Hence, as visual tasks involve the presence of highly structured
	objects - like for humans - the necessity of FOA leads to more complex 
	learning models that likely benefit from video blurring.
\end{svgraybox}




\chapter{Non-visual environmental interactions}
\label{APA-sec}
\AtBeginShipoutNext{\AtBeginShipoutUpperLeft{%
  \put(1in + \hoffset + \oddsidemargin,-5pc){\makebox[0pt][l]{
\boxformat 
  \framebox{\vbox{
\hbox{Published by Springer, Cham---Cite this chapter:}
\smallskip
\hbox{\url{https://doi.org/10.1007/978-3-030-90987-1_6}}}
  }}}%
}}

\vspace{-4cm}
\begin{quote}
Under normal conditions the research scientist is not an innovator but a solver of puzzles, and the puzzles upon which he concentrates are just those which he believes can be both stated and solved within the existing scientific tradition.\\
~\\
Thomas Kuhn, ``The Structure of Scientific Revolution'', 1962
\end{quote}


\section{Object recognition and related visual skills}
\label{obj-rec-sec}
This books has covered information extraction from visual sources as 
a consequence of the vision field\index{vision fields} theory. Animals and machines living
in a certain environment, however, are expected to carry out specific 
tasks that might involve actions and/or perceptions. In both cases
visual agents receive precious information which drives the learning 
process accordingly. The supervised learning protocol that has been
dominating computer vision for classification offers massive labelling
of the training set, so as the non-visual environmental interaction
results in the systematic labelling of each single example.
All animals, including humans, experiment truly different environmental 
interactions, which typically convey much less information with respect
to the supervised learning protocol. The predator-prey ordinary interactions
are not restricted to visual signals and, likewise, their movement
produces precious feedback for appropriate navigation plans. 
When shifting attention to humans, things become more sophisticated
and involve specific perceptual issues that seem to characterize
our more advanced abstraction capabilities.

\medskip
\noindent {\em Object recognition}\\
How can we recognize objects without massive supervision? How can children
associate the name to an object after a few explicit supervisions based on its
pointing?
%
%
The surprising limitations in frogs' visual skills suggest to better analyze
the fundamental role of eye movements and focus of attention\index{focus of attention}, which is mostly 
missing in frogs. Likewise, there is nothing similar to object supervision
based on pointing to a specific pixel in the retina, which clearly requires 
FOA. As discussed in this book, FOA 
is interwound with the development of visual features with different degree
of abstraction until the interpretation of objects is gained.
As such, it necessarily plays a crucial role for object
recognition, which does not typically require the full discovery of all
objects at a given frame. On the opposite, humans experiment a cognitive
process that is driven by the FOA, which leads to explore 
all the visual scene which is relevant from an information viewpoint. 
Of course, just like classic convolutional networks, the computational 
model driven by the focus of attention, referred to as foveated  networks\index{foveated neural networks}
in this book, can compute features and object categories on any pixel of the
retina. 

\marginpar{\small Objects' distinguishing properties and their recognition}
In order to discuss object recognition, we begin facing the following 
apparently trivial question: What is an object? Of course, it is connected with 
its visual interpretation. When trying to answer this question we early realize
that we need to unfold the semantics of objects. While they 
can be described in terms of visual features, objects are in fact characterized
by their physical structure, which is the distinguishing property that
hasn't been considered so far for visual features. 
The physical structure manifests itself through the connectedness of the composing atoms, 
which leads to regard objects as
single entities. The definition of the features
emerging from the two principles of visual perception doesn't reflect such a 
property that objects possess. While interaction features $\psi$ and $\chi$
gain properties connected with the object affordance, it's quite obvious that
they can only represent parts of an objects, since they are learned without
transmitting the above mentioned object physical connectedness property. 
Objects are like puzzles composed of different pieces (features); such 
a structured composition generates visual patterns whose composition needs
to be disclosed to the visual learning agent.


As already mentioned in Section~\ref{IdentityAffordanceSec},
the attachment to the earth, which leads to the distinction between
attached and detached objects results in a remarkable difference in their
affordance\index{affordance} and in the requirement of environmental interactions for 
gaining recognition.
\marginpar{\small Detached objects}
Detached objects can move with respect to the background, which provides 
information for capturing directly  the above mentioned property. 
The movement of only one object  with respect to the background can be detected 
easily by subtracting the velocity of the FOA. 
The estimation of the kinematic velocity makes it possible to localize 
the pixels where the object covers the retina. Some complex movements, which 
result in nearly null velocity of points of the object, don't compromise 
the acquisition of the object. Basically, in the special case of a 
single object moving with respect to the background there is an explicit
information coming from motion which characterizes the object. 
The optical flow\index{optical flow} in fact selects
the pixels of the moving object, so as to get semantic labeling for free. 

\marginpar{\small Attached objects}
The recognition of attached object presents a higher degree of complexity. We also need
to bear in mind that while some objects are attached in strict sense (e.g. think of houses of 
a village), even though others are not, they are visually similar to attached objects in their
visual environments since they rarely move. Unlike detached objects there is no optical flow
which transmits evidence on their identity. When looking at a pine tree from tens of meters
one perceive it thanks to its prominent trunk and to the green needles. 
There is no motion
information which can aggregate trunk and pine needles, so as we can perceive it as unique
connected entity. The trunk could in fact be aggregated with the
background without providing any additional cognitive evidence. Overall, complex objects which 
manifest themselves by many different parts seem to be more difficult to be recognized without 
involving explicit information on their structure. As already mentioned, the FOA is the crucial
mechanism which drives the recognition especially in this case. 
 \begin{figure}[H]
	\centering
	\includegraphics{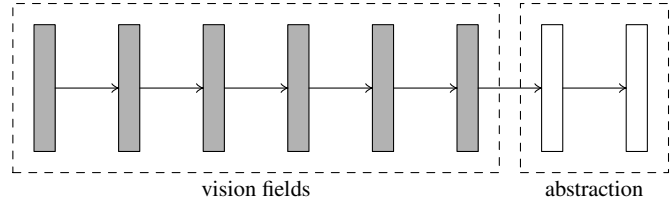}
\caption{\small The weights of the abstract module are learned in the second developmental stage
after having frozen the weights of the neurons that implement the vision fields.}
\label{MMI-Fig}
\end{figure}
A visual agent can ask 
questions on which  object(s) is (are) located in the pixel where he focuses attention,
which results in a very precious supervision. Of course, also the supervisor can  take the
initiative of providing the supervision, but this works only provided that the agent
has already reached a certain degree of learning of motion invariant\index{motion invariance} features. 
A fundamental results that emerges from the theory on vision fields is that,
as we provide the supervision on the FOA, the learning  process involves the weights $a$
used to properly select the visual features that concur to the recognition. Basically we have
\begin{equation}
    o_{i} = \sigma \left(\sum_{j} a_{ij} \chi_{j}\right),
\label{Obj-Rec-neuron}
\end{equation}
where we learn the weights $a_{ij}$ with the specific supervision on the FOA. 
This equation clearly unfolds the fundamental difference in the training process
that involves the extraction of high-level cognitive concepts, like structured
attached objects, with respect to the visual features $\chi_j$.

In the case in which $\chi_{j}$ is a feature group that is expected to be conjugated with the
velocity field $v$, also $o_{i}$ is conjugated with the same field, which can clearly explain
the segmentation capability. The localization of the pixels of the object
that are typically based on thresholding criteria on $a_{i}$ are in fact corresponding with
a related computation on the associated velocity field $v$ that allows us to 
identify the pixels covered in the retina by the object. Interestingly, the theory guarantees the consistency
with what can be extracted by $o_{i}$. 

Of course, we can also construct computational units for detecting objects $o_{i}$ which 
are based on feedforward-based computations described by $\vdash o_{i}$ and, therefore 
we can consider $o_{i}$ as a vision field that comes from different feature groups. 
This makes sense especially when considering highly structured objects. 
The corresponding conjugation\index{feature conjugation} with different velocity fields suggests that the overall
process of object recognition, as perceived by humans, needs to fully exploit the 
traditional supervision that is experimented with children. We simply provide supervision
on the presence of objects by pointing them straight or, alternatively, we answer children's questions 
on objects name. As discussed later on, the communication protocol is far richer than a
simple interaction on the specific point of FOA, but facing this case is very useful since
it also helps addressing more general linguistic interactions.

\marginpar{\small FOA-based supervision transferring: A single supervision on the FOA of attached objects results in a collection of supervisions.}

Hence, let us analyze the fundamental role of FOA that arises when we want to recognize
objects. The single supervision of an object in the FOA results in fact in a 
collection of related supervisions that are gained as the FOA moves in the retina. 
Let us assume that the frame of reference is located in the FOA. Basically, we can memorize the single supervision which can  
be transferred, later on, to all pixels related to the movement of focus. 
As such, we virtually collect frames referring to the same pixel where we focused attention seen from different frame references generated by the trajectory of the FOA.
This mechanism for supervision transferring holds until the initially supervised
pixel remains in the retina. It can even move itself, but it's important that its
location in the retina in order to activate the object recognition on its
position from the current FOA.


This analysis strongly supports the need of FOA for achieving top level object
recognition performance. Whenever FOA is missing, like in frogs, we are missing
the possibility of acquiring a lot of supervision for free. This holds also 
for convolutional neural networks, whose remarkable results seem to be 
due to the exploitation of the supervised learning protocol that is made possible 
in machines. Outside that battlefield, when the environmental interactions are
restricted to those typically experimented in nature, computational models 
of learning based on convolutional nets might be quite limited.

\section{What is the interplay with language?}
The interplay of vision and language is definitely one of the most challenging
issues for an in-depth understanding of human vision. While the vision field theory
presented in this book can be used as a basis for understanding from a functional 
viewpoint the visual skills in most animal species, humans exhibit a sophisticated 
behavior in the interplay between vision and language that is only superficially 
covered here. On the other hand, along with the associated successes, 
the indisputable adoption of the supervised learning protocol in most challenging
object recognition problems caused the losing of motivations for an 
in-depth understanding of the way linguistic information is synchronized 
with visual clues. In particular, the way humans learn the  
name of objects is far away from the current formal supervised 
protocol. This can likely be better grasped when we begin considering that 
top level visual skills can be found in many animals (e.g. birds and primates), 
which clearly indicates that their acquisition is independent of language. 

Hence, as we clarify the interplay of vision and language we will 
likely address also the first question on how to  overcome the need for ``intensive artificial supervision.'' 
Since first linguistic skills arise in children when their visual acuity is already 
very well developed, there is a good chance that simple early
associations between objects and their names can easily be 
obtained by ``a few supervisions'' because of the very rich 
internal representation that has already been gained of those objects. 
It is in fact only a truly independent hidden representation of 
objects which makes their subsequent association with 
a label possible! This seems to be independent of biology, whereas it looks 
like a fundamental information-based principle, which somehow
drives the development of ``what'' neurons. 

The interplay of language and vision has been recently very well 
addressed in a survey by~\cite{Lupyan_2012}).
 It is claimed that performance on tasks that have been presumed to be non-verbal is 
 rapidly modulated by language, thus rejecting  the distinction between verbal and non-verbal representations. While we subscribe the importance of sophisticated 
 interactions, we also reinforce the claim that identifying single objects
 is mostly a visual issue. However, when we move towards the acquisition of
 abstract notions of objects the interaction with language becomes more and more important.

Once again, the discussion carried out so far promotes the idea that 
for a visual agent to efficiently conquer the capabilities of recognizing
objects from a few supervisions, it must undergo some developmental steps
aimed at developing invariant representations of objects, so as
the actual linguistic supervision takes place only after the development of
those representations. But, when should we enable a visual agent to 
begin with the linguistic interaction? While one might address this question
when attacking the specific computational model under investigation, 
a more natural and interesting way to face this problem is to re-formulate the
question as:
\begin{quote}
	{\bf Q10:} 
	{\em
	How can we develop ``linguistic focusing mechanisms'' 
	that can drive the process of object recognition?}
\end{quote}
This  is done in a spectacular way in nature! 
Like vision, language development requires a lot of time. 
Interestingly, it looks like it requires more time than vision. 
The discussion in Section~\ref{BlurringSection} indicates that the
gradual growth of the visual acuity is a possible clue to begin with
language synchronization. The discussed filtering process offers a protection from
visual information overloading that likely holds for language as well. 
As the visual acuity gradually increases, one immediately realizes
that the mentioned visual-language synchronization has a 
spatiotemporal structure. At a certain time, we need to inform
the agent on what we see at a certain position in the retina. 
As already mentioned, an explicit implementation of such an association can be 
favored by an active learning process: the agent can ask itself
what is located at $(x,t)$. However, what if you 
cannot rely on such a precious active interactions?
For example, a linguistic description of the visual environment
is generally very sophisticated and mentions objects located in different
positions of the retina, without providing specific spatiotemporal information. 

\marginpar{\small Supervised learning from weak supervision}
Basically, this is a  sort of {\em weak supervision} that is 
more difficult to grasp. 
However, once again, developmental learning schemes can 
significantly help. At early stage of learning the agent's tasks can be 
facilitated by providing spatiotemporal information. For example, naming
the object located where the agent is currently focussing attention 
conveys information by a sort of human-like communication protocol.
As time goes by, the agent gains gradually the capability of recognizing
a few objects. What really matters is the confidence that is gained in 
such a task. When such a developmental stage is reached, linguistic 
descriptions and any sort of natural language based visual communication
can be conveniently used to reinforce the agent recognition confidence.
Basically, these weak supervisions turn out to be very useful since
they can profitably be attached where the agent  came up with a 
prediction that matches the supervision. 
The acquisition of the capability of recognizing objects described in the
previous section can in fact fire a subsequent reinforcement. If 
the agent detects the presence of a certain word on a given frame and he
can detect an object with that name, he can add such an information as 
an additional supervision, along with the correspondent compatible 
frames in which the object is still present in the retina.

 \begin{figure}[H]
	\centering
	\includegraphics{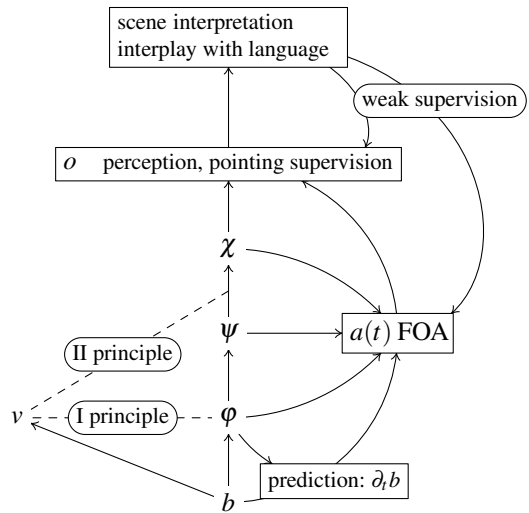}
\caption{\small Developmental learning to see. First vision fields are developed according
what is stated in the I and II Principle by following the arrows. The object acquisition
involves the additional pointing info on the FOA, while a virtuous loop with scene 
interpretation also reinforces object recognition.}
\label{DI-fig}
\end{figure}

\marginpar{\small Can it learn object names without explicit ``pointing supervision''?}
In the above discussion the weak supervision, which takes place without an explicit
``pointing information'', is used for reinforcing concepts that have been acquired
by supervision on the FOA. Can weak supervision also fire a learning process without 
leaning on pointing information? This is far more complex. The only possibility for 
the agent is that of discovering abstract reduced features $\chi$ that haven't been
labelled yet and attach them the candidate supervision. In principle, this is 
possible cognitive path that, however, is remarkably more difficult to follow with
respect to learning the name of an object in the FOA.

\marginpar{\small Developmental learning to see}
Overall, the investigation path followed in this book is disseminated of a number of insights
which seem to provide evidence that stage-based learning to see in animals is mostly an
information-based issue which holds regardless of biology. Fig~\ref{DI-fig} reports 
a summary of the drawn analysis which provide some evidence on the claimed role of 
time and on the need to undergo developmental stages. The stage transition is especially
evident as we involve the learning agent into the task of object recognition which 
requires additional information with respect to the video source. In particular
the weight of learning are independent of what has been obtained in the previous stages
on the construction of visually features invariant under motion. 

\begin{svgraybox}
	Computer vision and natural language processing have been mostly evolving independently
	one each other. While this makes sense, the time has come to explore the interplay
	between vision and language with the main purpose of going beyond the protocol of
	supervised learning for attaching labels to objects. 
	Interestingly challenges arises in scene interpretation when we begin considering 
	the developmental stages of vision that suggest gaining strong object invariance 
	before the attachment of linguistic labels.
\end{svgraybox}

\section{The ``en plein air'' perspective}
Posing the right questions is the first fundamental step to gain knowledge 
and solve problems. Hopefully, the questions raised in this book might give
the reader insights and contribute to face fundamental problems in computer vision. 
However, one might wonder what could be  the most concrete action for promoting
studies on the posed questions. So far, computer vision has  strongly benefited from the massive diffusion of
benchmarks which, by and large, are regarded as fundamental tools for performance evaluation.
However, it is clear that they are very well-suited to support the statistical machine learning\index{machine learning}
approach based on huge collections of labelled images. However, this book opens the doors 
to explore a different framework for performance evaluation. The emphasis on video instead of
images does not lead us to think of huge collection of video, but to a truly different approach
in which no collection at all is accumulated! Just like humans, machines are expected 
to live in their own visual environment. However, what is the scientific framework for 
evaluating the performance and understand when a theory carries out important new results?
Benchmark  bears some resemblance to the influential testing movement in psychology which has its roots in 
the turn-of-the-century work of Alfred Binet on IQ tests~(\cite{Binet1916}).
Both cases consist in attempts to provide a rigorous way of assessing the performance or the aptitude of a 
(biological or artificial) system, by agreeing on a set of standardized tests which, 
from that moment onward, become the ultimate criterion for validity. 
On the other hand, it is clear that the skills of any visual agent can be quickly evaluated 
and promptly judged by humans, simply by observing its behavior. 
How much does it take to realize that we are in front of person with visual deficits? 
Do we really need to accumulate tons of supervised images for assessing the quality of
a visual agent? The clever idea behind ImageNet~\cite{imagenet_cvpr09} is based on crowdsourcing. 
Couldn't we also use crowdsourcing as  a {\em crowdsourcing performance evaluation scheme}?
People who evaluate the performance could be properly registered so as to limit 
spam ( see e.g.~\cite{DBLP:conf/iciap/GoriLMMP15}). 

Scientists in computer vision could start following a sort of 
term {\em en plein air}, the term which is used  to  mimic the French Impressionist painters of the 19th-century and, more 
generally, the act of painting outdoors. This term suggests that visual agents should be evaluated 
by allowing people to see them in action, virtually opening the doors of research labs. 

While the idea of shifting computer vision challenges into the wild deserve attention 
one cannot neglect the difficulties that arise from the lack of a true lab-like environment
for supporting the experiments. The impressive progress in computer graphics, however, 
is offering a very attractive alternative that can dramatically facilitate the developments 
of approaches to computer vision that are based on the on-line treatment of the 
video (see e.g.~\cite{DBLP:journals/corr/abs-2007-08224}).

\begin{svgraybox}
	Needless to say, computer vision has been fueled by the availability of huge labelled 
	image collections, which clearly shows the fundamental role played by 
	pioneering projects in this direction (see e.g.~\cite{imagenet_cvpr09}).
	The ten questions posed in this paper will likely be better addressed only when 
	scientists will put more emphasis on the en plein air environment. In the 
	meantime, the major claim of this book is that the experimental setting 
	needs to move to virtual visual environments. Their photorealistic level along
	with explosion of the generative capabilities make these environment just
	perfect for a truly new performance evaluation of computer vision. The 
	advocated crowdsourcing approach might really change the way we measure
	the progress of the discipline.
\end{svgraybox}

\appendix 

\chapter{Calculus of Variations}

Calculus of variation is in its essence the  study of extremals of
functions $f\colon \bbX\to \overline \bbR$, where $\overline \bbR=
\bbR\cup\{-\infty,+\infty\}$. The case in which $\bbX$ is a Euclidean space
corresponds of course to the study of stationary points of a real valued
function on $\bbR^n$. 
For the purposes of this work we are mainly interested in
the case in which $\bbX$ is an infinite dimensional functional space.
In particular we will focus on the case in which $\bbX$ is an affine space
with vector space $V$ so that in particular for all $x\in \bbX$ and all $v\in
V$ we have that $x+v\in\bbX$. In this case then it is particularly
straightforward to generalize the usual concept of directional derivative
in the direction $v$ at point $x_0$ as follows:
\begin{equation}\label{gateau}
\delta F(x_0,v):= \lim_{s\to 0} \frac{F(x_0+s v)-F(x_0)}{s}.
\end{equation}
In general this quantity is called {\it G\^ateaux differential}, or more
traditionally {\it first variation}. The term G\^ateaux differential comes
from the notion of G\^ateaux differentiability in Banach spaces
(see~\cite{giaquinta2004calculus}). Notice also that if we define $\psi(s):=F(x_0+s v)$,
then $\delta F(x_0,v)=\psi'(0)$.

This quantity is particularly important for the study of extremals of a
functional since, as it happens for real valued functions, the vanishing of
this quantity for all $v\in V$ it is a necessary
condition to be satisfied by any local extremum of $F$.

\section{Integral Functional and Euler equations}
We will now restrict ourselves to functionals of the form
\begin{equation}
F(x):=\int_0^T L(t,x(t),\dot x(t))\, dt,
\end{equation}
where $x\in\bbX$ and $L(t,z,p)$ is a continuous real valued function of the
variables $(t,z,p)$.
Suppose furthermore that for example\footnote{A milder assumption
would be $\bbX\subset H^2((0,T); \bbR^n)$.} $\bbX\subset C^2([0,T]; \bbR^n)$.

Now we can use the following well known result about integration that
essentially says that
under appropriate regularity assumptions {\it the derivative of the integral is
the integral of the derivative}. More precisely if $f\colon [0,T]\times
[-\tau,\tau]\to\bbR$, then if we let
$\psi(s):=\int_0^T f(t,s)\, dt$
we have that
\begin{enumerate}
\item If $f$ is continuous in $ [0,T]\times [-\tau,\tau]$ then
$\psi$ is continuous in $[-\tau,\tau]$;
\item If $f_s$ is continuous in $ [0,T]\times [-\tau,\tau]$ then
$\psi'$ exists and it is given by
\[\psi'(s):=\int_0^T f_s(x,s)\, dx.\]
\end{enumerate}

If we take, as we remarked above $\psi(s):=F(x+s v)$, then
\[\psi'(s)=\int_0^T\frac{d}{ds} L(t, x+sv,\dot x+s\dot v)\]
Then, using the chain rule on $L$ we have that the first
variation~\eqref{gateau} looks like
\[
\delta F(x,v)=\int_0^T \bigl(L_z(t,x(t),\dot x(t))\cdot v(t)+
L_p(t,x(t),\dot x(t))\cdot \dot v(t)\]
Now consider the relation $\delta F(x,v)=0$ for all $v\in C^\infty$; here
we are considering $v\in C^\infty$ instead of $C^2$ functions since this
has the advantage that we can consider the same class of variations for all
differential equations of all order;  moreover it is consistent with the
usual conventions in the theory of distributions. This condition is
equivalent to
$$
\int_0^T \bigl(L_z(t,x(t),\dot x(t))\cdot v(t)+
L_p(t,x(t),\dot x(t))\cdot \dot v(t)\, dt=0
\quad\forall v\in C^\infty((0,T);\bbR^n)$$
This condition it is usually called the weak Euler equation for $x$.
Notice that in order for this condition to be well defined the function
$x(t)$ does not need more regularity than that declared in the definition of $\bbX$.

Now suppose that we take $v$ vanishing at the boundary, then by
integration by parts we get
\begin{equation}
\int_0^T \bigl(L_z(t,x(t),\dot x(t))-
{d\over dt}L_p(t,x(t),\dot x(t))\bigr)
\cdot v(t)\,dt=0.\label{ELE-byparts}\end{equation}
This, of course, can be done if we have enough regularity on both $L$ and
$x$. This integral relation can be turned into a differential equation by
using the fundamental lemma of calculus of variations:

\begin{lemma} Let $f\colon[0,T]\to \bbR^n$ be  continuous function. If
\begin{equation}\int_0^T f(t)\cdot v(t)\, dt=0\quad \forall v\in C^\infty_c((0,T);\bbR^n),
\label{FLCV}\end{equation} 
then $f\equiv 0$ in $[0,T]$.
\end{lemma}

We therefore have
\begin{equation}L_z(t,x(t),\dot x(t),\ddot x(t))-
{d\over dt}L_p(t,x(t),\dot x(t),\ddot x(t))=0.
\label{ELE-appendix}\end{equation}
Till now we have not taken into account the cases in which the variations
does not vanishes at the boundary; this for example happens when
if we formulate a variational
problem without specifying in $\bbX$ the value of the solution at the
boundaries. 

Notice however that since the fundamental lemma of the calculus of variation
the vanishing condition~\eqref{FLCV} need just to be verified for compactly
supported functions, if we can prove that the solution of the variational
problem is regular enough then the differential Eq.~\eqref{ELE-appendix}
will hold regardless of any boundary conditions. This being said let us  now
see what happens to the stationarity condition $\delta F(x,v)=0$
when we do not assume the vanishing of the variation at the boundaries.

We have already discussed the fact that Eq.~\eqref{ELE-appendix} 
still holds;
therefore we are left with the contributions only from the boundary terms of
the integration by parts; namely
\[
\left[L_p\cdot v\right]_0^T=0.
\]
In order for this term to be zero the only possibility, other than
the vanishing of $v$ is to have
\begin{equation}\label{eq:stat-cond-bound}
L_p=0,
\end{equation}
at the boundary where we do not know that the variation is vanishing.
This kind of boundary conditions usually are referred to as Neumann boundary
conditions since they generally depends on the derivatives of the solution.

\printindex

\bibliography{References,nn}
\bibliographystyle{plain}

\end{document}